\documentclass[letterpaper, 10 pt, conference]{ieeeconf}  % Comment this line out
\IEEEoverridecommandlockouts                              % This command is only
\usepackage{amsmath} % assumes amsmath package installed
\usepackage{amssymb}  % assumes amsmath package installed

\usepackage{amsmath,amsfonts,bm}

\def\1{\bm{1}}
\newcommand{\train}{\mathcal{D}}

\def\rvu{{\mathbf{i}}}

\def\rvu{{\mathbf{u}}}

\def\rvw{{\mathbf{w}}}
\def\rvx{{\mathbf{x}}}
\def\rvy{{\mathbf{y}}}

\def\sN{{\mathbb{N}}}

\def\sR{{\mathbb{R}}}

\def\sZ{{\mathbb{Z}}}

\newcommand{\E}{\mathbb{E}}

\newcommand{\KL}{D_{\mathrm{KL}}}

\usepackage[style=ieee]{biblatex}
\usepackage{hyperref}
\usepackage{graphicx}
\usepackage{xcolor}         % colors

\newcommand{\e}{\mathbf{e}}

\newcommand{\x}{\mathbf{x}}
\newcommand{\y}{\mathbf{y}}
\newcommand{\z}{\mathbf{z}}
\newcommand{\w}{\mathbf{w}}

\newcommand{\reals}{\mathbb{R}}

\newcommand{\hyf}{\hat{\y}_{f}}

\newcommand{\bP}{\mathbf{P}}
\newcommand{\bE}{\mathbf{E}}
\newcommand{\F}{\mathbf{F}}
\newcommand{\bnu}{\bm{\nu}}
\newcommand{\etab}{\bm{\eta}}

\newtheorem{Assumption}{Assumption}[section]
\newtheorem{lemma}{Lemma}[section]
\newtheorem{Problem}{Problem}[section]
\newtheorem{Remark}{Remark}[section]
\newtheorem{Definition}{Definition}[section]
\newtheorem{Theorem}{Theorem}[section]
\newtheorem{Corollary}{Corollary}[section]
\newtheorem{Proof}{Proof}[section]

\title{\LARGE \bf  PAC-Bayesian bounds for learning LTI-ss systems with input from empirical loss}

\author{Deividas Eringis*, John Leth, Zheng-Hua Tan, Rafal Wisniewski, Mihaly Petreczky% <-this % stops a space
\thanks{This work was not supported by any organization}% <-this % stops a space
\thanks{D. Eringis, J. Leth, Z. Tan, R. Wisniewski is with Department of Electronic Systems,
        Aalborg University, Aalborg, Denmark
        {\tt\small \{der,jjl,zt,raf\}@es.aau.dk}}%
\thanks{Mihaly Petreczky is with Laboratoire Signal et Automatique de Lille (CRIStAL), Lille, France {\tt\small mihaly.petreczky@centralelille.fr}}%
}

\begin{document}

\maketitle

\begin{abstract}
In this paper we derive a Probably Approxilmately Correct(PAC)-Bayesian error  bound for linear time-invariant (LTI) stochastic dynamical systems with inputs. Such bounds
are widespread in machine learning, and they are useful for characterizing the predictive power of models learned from  
finitely many data points.  
In particular, with  the bound derived in this paper relates
future average prediction errors with the prediction error 
generated by the model on the data used for learning.
In turn, this allows us to provide finite-sample  error bounds for
a wide class of learning/system identification algorithms. 
Furthermore, as LTI systems are a sub-class of recurrent neural
networks (RNNs), these error bounds could be a first step towards 
PAC-Bayesian bounds for RNNs. 
\end{abstract}

\section{Introduction}
Linear time invariant (LTI) state-space models have been widely used in control and econometric applications to model time-series and 
have rich literature on learning (classically called identification)\cite{LjungBook}. 

In this paper, we present PAC-Bayesian type bounds on learning LTI systems from data generated by LTI system driven by zero-mean, i.i.d., Gaussian or sub-Gaussian noise. 

The Probably Approximately Correct (PAC)-Bayesian framework, provides theoretical guarantees (with arbitrary high probability) on the difference between learning from infinite amount of data, and learning from finite empirical data, see \cite{guedj2019primer,alquier2021userfriendly,zhang-06,grunwald-2012,alquier-15,nips-16,ShethK17}. 

\textbf{Motivation}
PAC and PAC-Bayesian bounds have been a major tool for analyzing
learning algorithms. 
They provide bounds on the generalization error in terms of the empirical error,
in a manner which is independent of the learning algorithm. Hence, these bounds can be used to analyze and explain a wide variety of learning algorithms. Moreover,
by minimizing the error bound, new, theoretically well-founded learning algorithms can be 
formulated. In particular, PAC-Bayesian error bounds turned out to be useful for 
providing non-vacuous error bounds for neural networks \cite{Dziugaite2017}.
\par
  While there is a wealth of literature on PAC \cite{shalev2014understanding} and PAC-Bayesian \cite{alquier2021userfriendly,guedj2019primer},
  bounds for static models, much less is known on  dynamical systems.
  \par

    Traditionally, the literature on LTI systems \cite{LjungBook} has focused on statistical consistency. More recently, several results have appeared on finite-sample bounds for learning LTI systems, but they are valid only for specific learning algorithms or for very limited subclasses
    \cite{simchowitz2021statistical,Ozay2018,oymak2021revisiting},

   \textbf{Contribution}
     In this paper we consider stochastic LTI state-space representations (LTI systems for short)
     in innovation form.
     In accordance with the standard
     practice in system identification, we view
     stochastic LTI systems as predictors, which take past inputs and
     outputs and generate predictions for the current
     output. We assume that the data used for learning (system identification) are generated by
     stochastic LTI systems in innvation form too.
     Learning/identifying
      an LTI system is then amounts to
     finding the best predictor, i.e., the predictor
     which results in the smallest prediction error
     for the training data, i.e., in the smalled
     \emph{empirical loss}
     However, for decision making (fault detection,control, etc.), the quality of the learned model is determined by the 
     \emph{generalization
     error}, i.e., the average prediction error
     for future, unseen data. The PAC-Bayesian
     bound of this paper says that with a high probability (probability $1-\delta$), the generalization error
     is smaller than the empirical loss plus a 
     an error term. The error term depends on the
     number of data points $N$ and on parameter
     (learning rate $\lambda$). 
     In this paper we provide explicit formulas for
     the error term. We show that the error
     term converges to a constant as $N \rightarrow \infty$. The constant depends on the confidence
     level $\delta$ and the distance  between
     prior and posterior densities on models. 
     If we assume that the data used for learning is generated by an LTI system with \emph{bounded noise}, we can show that the error term
     converges to $0$ as $N \rightarrow \infty$.
     The rate of convergence is $O(\frac{1}{\sqrt{N}})$, which is consistent with most of finite-sample bounds available in the literature for various, not necessarily LTI, models. This suggests that the obtained error
     bound is likely to be asymptotically sharp for bounded signals.
        
    \textbf{Related work}
%\label{related:work}
   The related literature can be divided into the following categories.
  \\
   \textbf{Generalization bounds for RNNs.}
      PAC bounds for RNN were developed in \cite{KOIRAN199863,sontag1998learning,pmlr-v108-chen20d} using VC dimension, and
      in  \cite{pmlr-v161-joukovsky21a, pmlr-v108-chen20d} using Rademacher complexity, and in 
      \cite{pmlr-v80-zhang18g} using PAC-Bayesian bounds approach.
      However, all the cited papers assume noiseless models,  
      a fixed number of time-steps, that the training data are i.i.d sampled time-series, and the signals are bounded.
      In contrast, we consider (1) noisy models, (2) prediction error defined on infinite time horizon, (3) only one single time series available for training data, and (4) unbounded signals. 
      Moreover, several
      papers \cite{KOIRAN199863,sontag1998learning,pmlr-v144-hanson21a} assume
      Lipschitz loss functions, while we use quadratic loss function.

\textbf{Finite-sample bounds for system identification of LTI systems.}
Guarantees for asymptotic convergence of learning algorithms is a classical
topic in system identification \cite{LjungBook}.
Recently, several publications on finite-sample  bounds for learning linear dynamical systems were derived, without claiming completeness \cite{simchowitz2018learning,simchowitz2019learning,simchowitz2021statistical,oymak2021revisiting,lale2020logarithmic,foster2020learning,NEURIPS2018_d6288499,Pappas1,SarkarRD21}. 
% \cite{simchowitz2018learning,simchowitz2019learning}
First, all the cited papers propose a bound which is valid only for models generated by a specific learning algorithm. In particular, these bounds do not relate the generalization loss with the empirical loss for arbitrary models, i.e., they are not PAC(-Bayesian) bounds. This means that in contrast to the results of this paper, the bounds of the cited papers cannot be use for analyzing algorithms others than for which they were derived.
Second, many of the cited papers
do not derive bounds on the infinite horizon prediction error.
More precisely, \cite{oymak2021revisiting,SarkarRD21,lale2020logarithmic,Pappas1,Simchowitz_Foster_2020} provided error bounds for the difference of the first $T$ Markov-parameters of the estimated and true system for a specific
identification algorithm. However, in order to characterize the infinite horizon prediction error,
we need to take $T=\infty$.
For $T=\infty$ the cited bounds become infinite, i.e., vacuous. 
In addition, in contrast to the present paper,  \cite{oymak2021revisiting,SarkarRD21,simchowitz2018learning} deals only with the deterministic part of the stochastic LTI, \cite{Pappas1} deals only with the stochastic part.
% Note that the error bounds of
% the cited papers
%\cite{simchowitz2019learning} ,lale2020logarithmic,NEURIPS2018_d6288499,Pappas1,SarkarRD21}
% converge to their limit at rate $O(\frac{\ln(N)}{\sqrt{N}})$, which is comparable to the rate $O(\frac{1}{\sqrt{N}})$ of this paper.
% This being said, the cited papers 
% provide bounds on the parameter estimation error, and many of them allow marginally stable systems.

\textbf{PAC-Bayesian bounds for state-space representation.}
In \cite{haussmann2021learning} learning of stochastic differential equations without inputs was considered and it was assumed that  several independently sampled time-series were available for learning. 
In contrast, in this paper we deal with discrete-time systems with inputs and the learning takes place from a single time-series.
In \cite{PACMarkov} learning of general Markov-chains was considered, but the state of the Markov-chain was assumed to be observable and no inputs were considered. The learning problem of \cite{PACMarkov} is thus different from the one considered in this paper.

In \cite{CDC21paper} %cdc paper
PAC-Bayesian error bounds were developed for autonomous LTI state-space systems without exogenous input. 
In contrast to \cite{CDC21paper}, in the current paper we consider
systems with exogenous inputs.
Moreover, the error bound of this
paper is much tighter than that of \cite{CDC21paper}: 
in contrast to \cite{CDC21paper}, with the growth of the number of observations,
the error bounds of this paper converge either to zero (in the case of bounded innovation noise) or to a constant involving KL-divergence. 
Finally, the proof technique is completely different from that of
\cite{CDC21paper}.

\textbf{Paper Outline}
We start by defining the problem formulation in Section \ref{sect:Problem_formulation}, where all the assumptions and important quantities are defined. Then we will discuss the PAC-Bayesian framework in Section \ref{sect:pac:gen}, then we will present the main results of the paper in Section \ref{sec:mainResults}, then we will present some auxiliary results for systems driven by bounded noise in Section \ref{sec:boundedResults}, We will finish off with a short numerical example in Section \ref{sec:numEx}. Finally, we will have the conclusion in Section \ref{sect:concl}.

\section{Problem formulation}
\label{sect:Problem_formulation}
\subsection*{ Notation and terminology} 
We occasionally use $\triangleq$ to denote ''defined by''. 
Let $\F$ denote a $\sigma$-algebra on the set $\Omega$ and $\bP$ be a probability measure on $\F$. Unless otherwise stated all probabilistic considerations will be with respect to the probability space $(\Omega,\F,\bP)$, and we let $\bE(\z)$ denote expectation of the stochastic variable $\z$.
We use bold face letters to indicate stochastic variables/processes. 
Each euclidean space is associated with the topology generated by the 2-norm $\|\cdot\|_2$, and the Borel $\sigma$-algebra generated by the open sets. The induced matrix 2-norm is also denoted $\|\cdot\|_2$. 
 We say that a random variable $\z$
taking values in $\mathbb{R}^n$ is essentially bounded, if for some constant $C > 0$, $\|\z\|_2 < C$
holds with probability one.

A \emph{stochastic linear-time invariant (LTI) systems with inputs in state-space form} \cite[Chapter 17]{LindquistBook} is a dynamical system
of the form
\begin{equation}\label{eq:assumedSys}
	\begin{split}
		\x(t+1)=A\x(t)+B\mathbf{u}(t)+\bnu(t),\\
		\y(t)=C\x(t)+D\mathbf{u}(t)+\etab(t)
	\end{split}
\end{equation}
defined for all $t \in \mathbb{Z}$, where 
$A,B,C,D$ are $n \times n$, $n \times n_u$, $n_y \times n$ and $n_y \times n_u$ matrices respectively, $A$ is a Schur matrix (a square matrix with all its eigenvalues inside the unit disk), $\bnu,\etab$ are zero-mean Gaussian i.i.d processes,
$\mathbf{u}$, $\x$, are zero-mean stationary Gaussian processes, $\mathbf{u}(t)$ and $\begin{bmatrix} \etab^T(t),\bnu^T(t) \end{bmatrix}^T$ are independent, and $\x(t)$ and $\begin{bmatrix} \bnu^T(t),\etab^T(t) \end{bmatrix}^T$ are independent. 
The process $\x$ is called the state process, $\bnu$ is called the  process noise and $\etab$ is the 
measurement noise. 
%Note that $\x$ is uniquely determined by the matrices $A,B,E$ and noise $\v$
%The unique stationary  process $\x$ satisfies $\x(t)=\sum_{k=0}^{\infty} A^kB\mathbf{u}(t-k)+\sum_{k=0}^{\infty} A^k\bnu(t-k)$.
If $B,D$ are absent from \eqref{eq:assumedSys}, then we say that \eqref{eq:assumedSys} is an \emph{autonomous stochastic LTI system} \par
Let us fix stochastic processes $\y(t)\in \reals^{n_\y}$, and $\rvu(t)\in  \reals^{n_\rvu}$, that share a time axis $t\in\sZ$, that is, for any $t\in\mathbb{Z}$, $\y(t):\Omega\to\sR^{n_\y};\omega\mapsto\y(t)(\omega)$, and $\mathbf{u}(t):\Omega\to\sR^{n_\rvu};\omega\mapsto\rvu(t)(\omega)$
%; hence  $y(t):\omega\to\y(t)(\omega)$, $w(t):\omega\to\w(t)(\omega)$ 
are random vectors on $(\Omega,\F,\bP)$. The goal is to estimate $\y(t)$ from current and past values of $\rvu(t)$, for this we need a structure connecting $\rvy(t)$ and $\rvu(t)$, thus we have 
\begin{Assumption}\label{as:generator}
	Let $\y(t)$ and $\mathbf{u}(t)$ be generated by an autonomous stochastic LTI system
	\begin{subequations}\label{eq:generator}
		\begin{align}
			\x(t+1) &=A_g\x(t)+K_g\e_g(t), \\
			\begin{bmatrix}\y(t)\\\mathbf{u}(t)\end{bmatrix}&=C_g\x(t)+\e_g(t)
		\end{align}
	\end{subequations}
	where $A_g \in \mathbb{R}^{n \times n},K_g \in \mathbb{R}^{n \times m},C_g \in \mathbb{R}^{m \times n}$ for $n > 0$, $m=n_y+n_u\geq2$ and $\x$, $\y$ and $\e_g$ are stationary, zero-mean, and jointly Gaussian stochastic processes.
	%The processes $\x$ and $\e_g$ are called state and noise process, respectively. 
	%Recall that stationarity and square-integrability imply constant expectation and the covariance matrix $Cov(\y(t),\y(s))\triangleq\bE[(\y(t)-\bE[\y(t)])(\y(s)-\bE[\y(s)])^T]$ only depends on time lag $(t-s)$. 
	% Processes $\y$ and $\w$ are markovian, therefore there exists a splitting subspace $\mathcal{X}_t$, with $x(t)\in\mathcal{X}_t$, such that, $(H_{t-1}^-(y)\vee H_t^-(w))\perp (H_{t-1}^+(y)\vee H_t^+(w)) | \mathcal{X}_t$ or in more standard notation $E[\{z(s)\}_{s=t+1}^\infty|\{z(s)\}_{s=-\infty}^t,x(t)]=E[\{z(s)\}_{s=t+1}^\infty|x(t)]$.
	Furthermore, we require that $A_g$ and $A_g-K_gC_g$ are Schur (all its eigenvalues are inside the open unit circle), that %for any $t,k \in \mathbb{Z}$, $k \geq 0$, $E[\e_g(t)\e_g^T(t\!-\! k\!-\! 1)]=0$, $E[\e_g(t)\x^T(t-k)]=0$, i.e., the stationary Gaussian process 
	$\e_g(t)$ is white noise uncorrelated with $\x(t-k)$, with covariance $\bE[\e_g(t)\e_g^T(t)]=Q_e$, and that $\e_g$ is the innovation process (see \cite{LindquistBook} for definition) of $\begin{bmatrix} \y^T & \mathbf{u}^T \end{bmatrix}^T$. 
	%see \cite{LindquistBook} for the definition of the innovation process.
	We identify the system \eqref{eq:generator} with the tuple
	$\Sigma_{gen}\triangleq(A_g,K_g,C_g,I)$;% note that the state process $\x$ is uniquely defined by
	%the infinite sum $\x(t)=\sum_{k=1}^{\infty} A_g^{k-1}K_g\e_g(t-k)$
\end{Assumption}
% Note that if there is no feedback from $\y$ to $\mathbf{u}$ (see \cite[Definition 17.1.1]{LindquistBook}), then, by \cite{LindquistBook,eringis2021optimal}, Assumption \ref{as:generator} is equivalent to the existence of a stochastic LTI  \eqref{eq:assumedSys} with output $\y$ and input $\mathbf{u}$.\par
\textbf{Note:} For learning, we assume to have the training data set $\train_N = \{\{\y(s),\rvu(s)\}\}_{s=0}^{N-1}$, i.e. a single trajectory of $[\y^T(t),\rvu^T(t)]^T$, but no knowledge of the matrices $A_g,K_g,C_g$ and noise process $\e_g$. The system \eqref{eq:generator} only defines the assumptions on the data generating process. \par
The goal is to use the past and present of $\rvu(t)$, or past of $\y(t)$, to estimate $\y(t)$. 
Note that $\y$ and $\rvu$ are stationary processes by
\cite[Theorem 1.4]{CainesBook}. Moreover,
from classical theory of LTI systems
it follows that $\y(t)$ and $\rvu(t)$, $t \in \mathbb{Z}$
are essentially bounded if the noise 
$\e_g(s)$ is essentially bounded for all $s \in \mathbb{Z}$

That is we wish to consider LTI predictors, 
	\begin{subequations}\label{eq:predictor}
		\begin{align}
			\hat{\x}(t+1)&=\hat{A}\hat{\x}(t)+\hat{B}\rvu(t)+\hat{L}\y(t), ~ \hat{\x}(0)=0 \\
			\hat{\y}(t)&=\hat{C}\hat{\x}(t)+\hat{D}\rvu(t)
			%\hat{\x}(0)&=E[\bar{x}(0)]=0
		\end{align}
	\end{subequations}
where matrices $\hat{A}, \hat{B}, \hat{L}, \hat{C}, \hat{D}$ are of appropriate size, and $\hat{A}$ is Schur (all its eigenvalues are inside the unit disk).

\textbf{Note:} In this paper, we will allow a more general form of predictors, where $\hat{L}$ can be set to 0, i.e. we may wish to estimate $\y(t)$ only from measurements $\rvu(t)$, when past values of the process $\y(t)$ is not available. In order to accommodate this let us define a stochastic process $\w(t)\in \sR^{n_\w}$, by two cases
\begin{itemize}
    \item $\w(t)=\begin{bmatrix} \y^T(t) & \rvu^T(t) \end{bmatrix}^T$, $n_\w=n_\y+n_\rvu$
    \item $\w(t)=\rvu(t)$, $n_\w=n_\rvu$
\end{itemize}
Note that, one can define $\w(t)$, to consist of some of the components of $\y(t)$, i.e. $\w(t)$ does not need to contain all of $\y$. 

% , consisting of $n_\w\leq n_\y+n_\rvu$ components of $\begin{bmatrix} \y^T(t) & \rvu^T(t) \end{bmatrix}^T$, that are used to estimate $\y(t)$. 

% whose components are the components of $\begin{bmatrix} \y^T(t) & \rvu^T(t) \end{bmatrix}^T$ used to estimate $\y(t)$. More explicitly, we will consider two cases

\textbf{Class of predictors (hypotheses)} In this paper, we will be interested in the following hypothesis class, consisting of predictors realizable by LTI systems.
\begin{Assumption}[Parameterised hypothesis class]\label{as:parameterisation}
	The hypothesis class $\mathcal{F}$ is a parametrized set of LTI predictors, with $\Sigma(\theta)=(\hat{A}(\theta),\hat{B}(\theta),\hat{C}(\theta),\hat{D}(\theta))$:
        \begin{subequations} \label{eq:formal_predictor}
            \begin{align}
                \hat{\rvx}(t+1)&=\hat{A}\hat{\rvx}(t)+\hat{B}\rvw(t), \;\hat{\rvx}(0)=0,\\
                f_{\Sigma(\theta)}(\{\rvw(s)\}_{s=0}^{t})&=\hat{C}\hat{\rvx}(t)+\hat{D}\rvw(t).
            \end{align}
        \end{subequations}
	$$\mathcal{F}=\{f_{\Sigma(\theta)} \mid \gamma(\hat{A}(\theta))<1,\;\theta\in\Theta\}$$
	with $\gamma(\hat{A}(\theta))$ the spectral radius of $\hat{A}(\theta)$, i.e. the largest modulus of eigenvalues of $\hat{A}(\theta)$. Set $\Theta\subset \sR^{n_\theta}$ is a compact set, and  
	%for any $\theta\in\Theta$, 
	%the tuple $$\mathbf{\Sigma}(\theta)=(\hat{A}(\theta),\hat{B}(\theta),\hat{C}(\theta),\hat{D}(\theta))$$
	%is such that 
	$\hat{A}(\theta)$,$\hat{B}(\theta),\hat{C}(\theta),\hat{D}(\theta)$ are continuous functions of 
	$\theta$ taking values in the sets of 
	$\hat{n}\times\hat{n}$, $\hat{n}\times n_w$,
	$n_y\times \hat{n}$ and $n_y\times n_w$
	matrices respectively.
        If $\w(t)=[\y^T(t),\rvu^T(t)]^T$, then  $\hat{D}=[0,\hat{D}_\rvu]$ for some $n_y \times n_u$ matrix $\hat{D}_\rvu$, i.e., $\hat{D}\w(t)$  depends only on $\mathbf{u}(t)$\footnote{The latter assumption is necessary, since otherwise we would be using the components of $\y(t)$ to predict $\y(t)$, which is not meaningful.}.
\end{Assumption}
We will identify the system \eqref{eq:formal_predictor}
with the tuple 
$(\hat{A},\hat{B},\hat{C},\hat{D})$.
%\emph{a LTI predictor realization}.
%\end{Definition}
% We will often denote the predictor 
% realizable by the LTI system $(\hat{A},\hat{B},\hat{C},\hat{D})$ by
% $f_{(\hat{A},\hat{B},\hat{C},\hat{D})}$.
For the sake of notation, throughout the paper we will use $f$, to denote $f_{\Sigma(\theta)}$, for some arbitrary $\theta\in\Theta$.

Under assumption \ref{as:parameterisation}, we can use probability densities on the set of predictors $\mathcal{F}$. The latter will be essential for using the PAC-Bayesian framework. 
% Loosely speaking, the  learning problem for stochastic LTI systems is as follows:
% based on a sample of finite length $\{y(t),u(t)\}_{t=1}^{N}$  of $\{\y,\mathbf{u}\}$, estimate the matrices $A,B,C,D$
% of \eqref{eq:assumedSys}. In addition, often the variance of the noises $\bnu$,$\etab$ is also estimated. 

Next, we define the notions of empirical and generalization loss for 
predictors which are realized by LTI systems.
\begin{Assumption}[Quadratic loss function]\label{as:lossfunc} \;\\
	We will consider \emph{quadratic loss functions}
	$\ell : \reals^{n_y}\times \reals^{n_y} \ni (y,y') \mapsto \|y-y'\|_2^2=(y-y')^T(y-y') \in [0,\infty)$.
\end{Assumption}
The empirical loss  of a  predictor for the data $\train_N=\{\y(t),\w(t)\}_{t=0}^{N}$ is defined as follows:
we define the random variable
\[\hyf(t\mid s) \triangleq f(\w(s),\ldots,\w(t))\] 
%for the random variable 
which represents the estimate of $\y(t)$ based on random variables $\{\w(s),\ldots,\w(t)\}$ .
%The quantity $\bE[\ell(\hyf(t\mid s),\y(t))]$, measures the expected difference between the actual process $\y(t)$ and the predicted value $\hyf(t \mid s)$ based on $\{\w(\tau)\}_{\tau=s}^{t}$.
%Equipped with a loss function, we define empirical loss.
%in order to evaluate the predictor on some training set $S_N$.
%\begin{Definition}
	The \emph{empirical loss for a predictor $f$ }and processes $(\y,\w)$ is defined by
	\begin{equation}
		\hat{\mathcal{L}}_{N}(f)\triangleq\frac{1}{N}\sum_{i=0}^{N-1} \ell(\hyf(i \mid 0), \y(i)).\label{eq:EmpLoss}
	\end{equation}
%\end{Definition}
The definition of the generalization loss is a 
bit more involved. Namely, we are using varying number of inputs for predictions and 
hence 
%the size of the predictors, i.e., the number of inputs 
%used for prediction  will impact its quality:
the expectation $\bE[\ell(\hyf(t \mid 0),\y(t))]$ depends on $t$.
This will hold true even if the processes $\y$ and $\w$ are stationary.
Note that this issue is specific for state-space models: autoregressive models 
always use the same number of inputs to make a prediction, see Remark \ref{remark:ARMAdifference}.
%\\
In this paper we will opt for looking at the case when the size of the past used for the prediction is infinite.
%the generalization loss will be the limit of the expected loss as the number of data points used for prediction tends to infinity. 
%This is consistent with the usual practice in system identification
%\cite{LjungBook}. From a practical standpoint it is reasonable, because the longer the predictor
%runs, the more of the past dat
%a it has integrated.  
%We will define this quantity as generalised loss.
To this end, we need the following result from \cite{HannanBook}.
\begin{lemma}[\cite{HannanBook}]
%[Infinite past prediction]
\label{l:ihp}
    \; \\
    The limit 
	%begin{equation}
	%\label{model:inf:eq-1}
	\( \hyf(t)=\lim_{s \rightarrow -\infty} \hyf(t \mid s) \)
	%\end{equation}
	exists
	in the mean-square sense for all $t$, the process $\hyf(t)$ is stationary, and
	%\begin{equation}
	%	\label{model:inf:eq}
	\(	\bE[\ell(\hyf(t),\y(t))]=\lim_{s \rightarrow -\infty} \bE[\ell(\hyf(t \mid s),\y(t))] \).
	%end{equation} 
	%and $\bE[\ell(\hyf(t),\y(t))]$ does not depend on $t$. 
\end{lemma}
This motivates us to introduce
%\begin{Definition}[Generalization loss of a predictor]\;\\
the quantity
	\[ \mathcal{L}(f)=\bE[\ell(\hyf(t),\y(t))]=\lim_{s \rightarrow -\infty} \bE[\ell(\hyf(t \mid s),\y(t))]
	\]
	which is called the \emph{generalization loss} of the predictor $f$ when applied to
	process $(\y,\w)$.% or simply generalization loss, when $\y$ is clear from the context. 
%\end{Definition}
\par
Intuitively, $\hyf(t)$ can be interpreted as the prediction of
$\y(t)$ generated by the predictor $f$ based on all (infinite) past and present values of $\w$. As stated in Lemma~\ref{l:ihp} we consider the special case when $\hyf(t)$ is the mean-square limit of $\hyf(t \mid s)$ as
$s \rightarrow -\infty$. Clearly, for large enough $t-s$, the empirical loss, is close to the generalization loss. 
%More precisely, from the proof of Lemma II.2 of \cite{CDC21paper}
%it follows that $\lim_{N \rightarrow \infty} \bE[\hat{\mathcal{L}}_N(f)]=\mathcal{L}(f)$.
%Note that the generalised loss is independent of the training data, and as such is suitable to quantify the "quality" of a predictor.
In fact, it is standard practice in learning dynamical systems \cite{LjungBook} to use $\mathcal{L}(f)$ as the measure of fitness
of the predictor.
%In order to define the learning problem, we will fix the class of predictors (hypotheses).
%\textbf{Learning problem}
With these definitions in mind, the learning problem considered in this paper can be stated as follows.
\begin{Problem}[Learning problem]
\label{learn:prob}
Compute a predictor $f \in \mathcal{F}$ from a sample 
$\train_N=\{\y(t)(\omega),\rvw(t)(\omega)\}_{t=0}^{N}$ 
of the random variables $\{\y(t),\w(t)\}_{t=0}^{N}$ 
such that the generalization loss $\mathcal{L}(f)$ is small.
\end{Problem}

\begin{Remark} \label{remark:ARMAdifference}
%Note that autoregressive models use fixed number of past values of predictor %variables to predict the output, while for state-space representations this number increases with time. Indeed, 
%$\hyf(t|s)=\sum_{j=s}^{t-1} \hat{C}\hat{A}^{t-j-1}\hat{B}\w(j)+\hat{D}\w(t)$,
%i.e., the number of past values of $\w$ is $t-s$. 
It is known
\cite[Section 4.2]{LjungBook} that the LTI system 
\eqref{eq:predictor} can be
rewritten as an ARX model: %\text{LTI of order n predictors:}\quad
\begin{equation}
\label{lti:arx}
\hat{\y}_f(t|s)=\sum_{i=1}^{n}\hat{\gamma}_i \hat{\y}_f(t-i|s) + \sum_{i=0}^{n-1}\hat{\eta}_i \w(t-i)
\end{equation}
At a first glance this is similar to classical ARX predictors, where
\(\hat{\y}(t)=\sum_{k=1}^n\hat{\alpha}_k \y(t-k) + \sum_{i=0}^{n-1}\hat{\beta}_i \w(t-i) \)
where $\y$ is predicted based on the last $n$ values of $\y$ and $\w$.
However, in contrast to classical ARX models, in \eqref{lti:arx} we do not use the past values of $\y$, but the past values
of the prediction $\hat{\y}_f$.
This difference has significant consequences, in particular, it means that
the previous results \cite{shalaeva2019improved} do not apply.  Note that \cite{alquier2012pred,alquier2013prediction} studied autoregressive models
without inputs (nonlinear AR models), 
so those results are not applicable either.
%In the light of the relationship between LTI predictors and representations
%of $\y$ by stochastic LTI systems with inputs, this is hardly surprising. 
In fact, the problem of learning LTI systems with inputs, or, which is almost equivalent, learning LTI predictors,  is essentially equivalent to learning ARMA models, and the latter is much more involved than learning
ARX models.
%hile, state-space predictors more closely follow Output Error model structure \cite{LjungBook}, where, one now computes $\hat{\y}$, instead of measuring $\y$ %p. 85 (p. 108 in pdf)
%\
%ARMA predictors follow ARMA representation of the process $\y$, with estimates of the process parameters $\alpha_k,\beta_k$, one obtains the prediction of the process
\end{Remark}

\section{PAC-Bayesian Framework}
\label{sect:pac:gen}
Below we present the adaptation of the PAC-Bayesian framework for LTI systems.
%is closely related to the PAC-Bayesian framework for autonomous LTI system described in %\cite{CDC21paper}. 
To this end, let $B_{\Theta}$ be the $\sigma$-algebra of Lebesque-measurable subsets of the parameter set $\Theta\subseteq\mathbb{R}^{n_\theta}$, and $m$ denote the Lebesque measure on $\mathbb{R}^{n_\theta}$. We then define
\begin{equation}
	\underset{f \sim \rho}{E} g(f)\triangleq\int_{\theta \in \Theta} \rho(\theta)g(f_{\Sigma(\theta)})dm(\theta)\label{eq:Edef}
\end{equation}
with $\rho$ a probability density function on the measure space $(\Theta,B_{\theta},m)$, and $g:\mathcal{F} \rightarrow \mathbb{R}$ a map such that
$\Theta \ni \theta \mapsto g(f_{\theta})$ is measurable and absolutely integrable. 
The essence of the PAC-Bayesian approach is to prove 
that for any density $\pi$ on $\mathcal{F}$, and
any $\delta\in(0,1]$, 
% \begin{align} \label{T:pac:gen}
% 		\bP \Bigg( \Bigg\{ \omega \in& \Omega  \mid  
% 		E_{f\sim \hat{\rho}} \mathcal{L} (f) 
% 		 \nonumber \le E_{f\sim \hat{\rho}} \hat{\mathcal{L}}_{N}(f)(\omega) \\
% 		 &
% 		  + r_N 
% 		  %\dfrac{1}{\lambda}\!\left[ KL(\hat{\rho} \|\pi) +
% 		%\ln\dfrac{1}{\delta}
% 		%+ \Psi_{\ell,\pi}(\lambda,N)  \right]  
% 		\Bigg \}\Bigg) > 1-\delta \
% 	\end{align}
\begin{align} 
\label{T:pac:gen}
	\bP \Big( \Big\{ \omega \in \Omega  \mid  \forall \hat{\rho}\in \mathcal{M}_\pi : 
	\underset{f\sim \hat{\rho}}{E} \mathcal{L} (f) \le \kappa(\omega)&
	\Big \}\Big) > 1-\delta,
\end{align}
with
$$
\kappa(\omega)=\underset{f\sim \hat{\rho}}{E} \hat{\mathcal{L}}_{N}(f)(\omega) + r_N
$$
$\mathcal{M}_\pi$ the set of all absolutely continuous densities w.r.t $\pi$, and $r_N=r_N(\pi,\hat{\rho},\delta)$ an error term. That is, the PAC-Bayesian bound holds for every posterior $\hat{\rho}$ in $\mathcal{M}_\pi$, simultaneously.

We may think of $\pi$ as a prior distribution density function and $\hat{\rho}$ as any candidate to a posterior distribution on the space of predictors. 
The inequality \eqref{T:pac:gen} says that the average generalization loss for models sampled from the posterior distribution is smaller than the average empirical loss for the posterior distribution plus the error terms $r_N$. \par
%For a discussion on how to use PAC-Bayesian for learning, see \cite{alquier2021userfriendly}. In a nutshell, 
A learning algorithm
can be thought of as fixing a prior $\pi$ and then
choosing a posterior $\hat{\rho}$ for which
$\kappa(\omega)$ is small.
%i.e., the sum
%$E_{f\sim \hat{\rho}} \hat{\mathcal{L}}_{N}(f)(\omega)
%		 + r_N(\pi,\hat{\rho},\delta)$ is small.
Moreover, $\kappa(\omega)$ can be viewed as a cost function involving the
empirical loss and the regularization term $r_N$.
The learned model
is either sampled from the posterior density $\hat{\rho}$, or it is chosen
as the one with maximal likelihood w.r.t. $\hat{\rho}$. 
Inequality \eqref{T:pac:gen} then gives guarantees on the 
generalization loss of the learned model.
For more details on using PAC-Bayesian bounds see \cite{alquier2021userfriendly}
%As it was pointed out in %\cite{alquier2021userfriendly}, 
%\eqref{T:pac:gen} could be used to derive
%so called \emph{oracle inequalities}. An oracle inequality relates
%generalization error of the posterior with 
%the best possible generalization error achievable for the given hypothesis class.
For \eqref{T:pac:gen} to be useful, 
the term  $r_N$ 
%to be
%decreasing in $\delta$ and in $N$. Ideally, wewould like $r_N$ 
should converge to a small constant, preferably zero,  as $N \rightarrow \infty$, and to be decreasing in $\delta$. The most common way of expressing the error term $r_N$, is based on Donsker-Varadhan’s change of measure \cite[Theorem 3]{nips-16}:
%\cite{alquier2021userfriendly}, the error term $r_N$ is of the form
    \begin{align} 
	r_N=
		 \dfrac{1}{\lambda}\!\left[ \KL(\hat{\rho} \|\pi) +
	\ln\dfrac{1}{\delta}+ \Psi_{\pi}(\lambda,N)  \right] \label{T:pac},
	\end{align}
where $\lambda > 0$ and
	$\KL(\hat{\rho} \mid \pi)\triangleq E_{f\sim\hat{\rho}} \ln \frac{\hat{\rho}(f)}{\pi(f)}$ is the KL-divergence between
	$\pi$ and $\hat{\rho}$, and 
	\begin{equation}
	\label{T:pac:2}
	   \Psi_{\pi}(\lambda,N) \triangleq \ln E_{f\sim\pi} \bE[e^{\lambda(\mathcal{L}(f)-\hat{\mathcal{L}}_{N}(f))}]
	\end{equation}
That is, $r_N$ involves the KL-divergence and a free parameter $\lambda$.
The density which minimizes $\kappa(\omega)$, with $r_N$ from \eqref{T:pac} is known as the Gibbs-posterior
\cite{alquier2021userfriendly} and it can be explicitly computed, i.e. \begin{align}
\rho_{\text{Gibbs}}(f)&\triangleq Z^{-1}\pi(f)\exp(-\lambda \hat{\mathcal{L}}_N(f)),\label{eq:gibbs}\\ 
Z&\triangleq E_{f\sim\pi}\exp(-\lambda \hat{\mathcal{L}}_N(f)). \nonumber
\end{align}
%Moreover, $\lambda$ i. \\
The disadvantage of this approach is that it is difficult to bound
$\Psi_{\pi}(\lambda,N)$, since it involves bounding higher-order moments
\begin{align}
    \bE[|\mathcal{L}(f)-\hat{\mathcal{L}}_N(f)|^r],\quad  r\in\sN
\end{align}

% \textcolor{blue}{
% \begin{Problem}[Bayesian learning problem]
%     find the probability density function $\hat{\rho}(f_{\Sigma(\theta)}|\train_N):\mathcal{F}\to \sR_+$ from a sample $\train_N=\{y(t),w(t)\}_{t=0}^{N}$, s.t. $E_{f\sim\hat{\rho}}\mathcal{L}(f)$ is minimised, then either
%     \begin{itemize}
%         \item sample $\hat{f}$ according to $\hat{\rho}$, (more likely to sample predictors which yield smaller generalization loss)
%         \item take the maximum likelihood predictor $\hat{f}=\argmax_f \hat{\rho}(f)$
%     \end{itemize}
% \end{Problem}
% }
One can also use PAC-Bayesian bounds, in order to choose the prior $\pi$ or the hypothesis class $\mathcal{F}$, s.t. the difference between generalised loss and empirical loss is within some acceptable level, i.e.  
\begin{align}
    E_{f\sim\rho}\left ( \mathcal{L}(f)-\hat{\mathcal{L}}_N(f) \right ) \leq r_N(\lambda,\pi) \leq \epsilon
\end{align}
then it is only a matter of choosing $\pi,\lambda,\mathcal{F}$, s.t. $r_N(\lambda,\pi)\leq \epsilon$, after which one can proceed with more standard Bayesian learning approach on just the empirical loss $\hat{\mathcal{L}}_N(f)$.

In the next section, we will apply a simple trick, which will allow us to upper-bound higher-order moments. 
\section{Main Results} \label{sec:mainResults}
In this paper we derive PAC-Bayesian bounds  \eqref{T:pac:gen}
for LTI systems. The main idea is to use the change of measure inequality from \cite[Theorem 3]{nips-16}. The major
challenge is to bound the corresponding moment generating function/higher-order moments of $(\mathcal{L}(f)-\hat{\mathcal{L}}_N(f))$. However this brings some technical challenges. Namely, the processes involved are not i.i.d.. Moreover, they are not bounded, and the quadratic loss function is not Lipschitz.
In addition, the empirical loss $\hat{\mathcal{L}}_N(f)$ is not
an unbiased estimate of the generalization loss $\mathcal{L}(f)$. This 
is specific to state-space representations, for auto-regressive models 
considered in \cite{alquier2012pred,alquier2013prediction,alquier:hal-01385064} this problem does not occur.  All these issues make it impossible to directly apply existing techniques \cite{alquier2012pred,alquier2013prediction,alquier:hal-01385064}. \\
As the first step, temporarily we replace the empirical loss $\hat{\mathcal{L}}_{N}(f)$ by
\begin{equation}
	\label{inf:emp:pred}
	V_N(f)\triangleq\frac{1}{N}\sum_{i=0}^{N-1}(\y(i)-\hyf(i))^2
\end{equation}
where the finite-horizon prediction $\hyf(t\mid 0)$ is replaced by the 
infinite horizon prediction $\hyf(t)$ defined in Lemma \ref{l:ihp}.
%Intuitively, $V_N(f)$ represents the empirical prediction error
%if the whole past of $\w$ was used for prediction, as opposed
%to a finite portion of the past of $\w$. \\
The advantage of $V_N(f)$ over $\hat{\mathcal{L}}_{N}(f)$
is that $V_N(f)$ is an unbiased estimate of the generalization loss
$\mathcal{L}(f)$, i.e.,
$\bE[V_N(f)]=\mathcal{L}(f)$.
Indeed, since $\y(t)-\hat{\y}_f(t)$ is a stationary process, $E[\|\y(i)-\hat{\y}_f(i)\|^2_2]=\mathcal{L}(f)$ 
does not depend on $i$, and hence 
$\bE[V_N(f)]=\frac{1}{N} \sum_{i=0}^{N-1} E[\|\y(i)-\hat{\y}_f(i)\|^2_2]= \mathcal{L}(f)$.
hence, usual techniques for deriving error bounds are easier to
extend to $V_N(f)$ than to $\hat{\mathcal{L}}_{N}(f)$.
Moreover, , from Lemma B.7 in Appendix B of the supplementary
material, it follows that $\hat{\mathcal{L}}_{N}(f)-V_N(f)$ converges to zero as $N \rightarrow \infty$
%
%from the proof of (Lemma II.2) in \cite[Lemma II.2]{CDC21paper} it follows that
%$\lim_{N \rightarrow \infty} \hat{\mathcal{L}}_{N}(f)-V_N(f)=0$
%where the limit is understood 
in the mean sense. 
% This suggests that for large enough $N$, the empirical prediction loss $\hat{\mathcal{L}}_{N}(f)$ could be replaced by the infinite past prediction error $V_N(f)$\\
%, and we could use the latter in the PAC-Bayesian error bound. \\
In order to derive upper bounds on the errors of the type \eqref{T:pac}, we will first derive upper bounds of the type \eqref{T:pac}, for $\mathcal{L}(f)-V_N(f)$, secondly we will derive upper bounds for $V_N(f)-\hat{\mathcal{L}}_N(f)$, then we will combine them using union bound. 
%we decompose the moments $(\mathcal{L}(f)-\hat{\mathcal{L}}_N(f))^r=(\mathcal{L}(f)-V_N(f) + V_N(f)-\hat{\mathcal{L}}_N(f))^r$, which then by convexity of $\phi(x)=x^r$, is bounded by
% \begin{multline}
%     (\mathcal{L}(f)-\hat{\mathcal{L}}_N(f))^r\leq 2^{r-1} (\mathcal{L}(f)-V_N(f))^r\\
%     +2^{r-1}(V_N(f)-\hat{\mathcal{L}}_N(f))^r
% \end{multline}
Doing this might seem counter-productive, however it is significantly easier to bound moments, $\bE[(\mathcal{L}(f)-V_N(f))^r]$, and $\bE[(V_N(f)-\hat{\mathcal{L}}_N(f))^r]$

% we apply change of measures on $\lambda|\mathcal{L} (f) -  V_N(f)|$ instead of $\lambda|\mathcal{L} (f) -  \hat{\mathcal{L}}_{N}(f))|$. Then we apply another change of measure on $\lambda|V_N(f)-\hat{\mathcal{L}}_N(f)|$. By combining these two bounds we will finally derive a PAC-Bayesian error bound for $\hat{\mathcal{L}}_{N}(f)$.
For every predictor $f$ 
 we define the following constants. %% $G(f)$ and $G_e(f)$. 
	\begin{Definition}[Constants $\bar{G}_f(f),G_e(f)$]
	\label{def:constants}
	 Let $f=(\hat{A},\hat{B},\hat{C},\hat{D})$ be a predictor.
	 % The constants $G(f)$, $G_e(f)$ are define as follows.
     Let
	$A_g,K_g,C_g$ be the matrices of the data generator  from Assumption \ref{as:generator}. Define
	the matrices $(A_e,K_e,C_e,D_e)$ as
	  $ D_e= I-\hat{D}_w$,
	\begin{align*}
	   & A_e=
	   \begin{bmatrix} A_g & 0 \\ \hat{B}C_w & \hat{A}  \end{bmatrix}, 
	   ~
	   K_e=
	    \begin{bmatrix} K_g \\
	    \hat{B}_w
	    \end{bmatrix},
	     ~ C_e=
	         \begin{bmatrix} (C_1-\hat{D}C_w)^T \\ -\hat{C}^T \end{bmatrix}^T,
	    \end{align*}
	where $C_g=\begin{bmatrix} C_1^T & C_2^T \end{bmatrix}^T$ and $C_1$ has $n_y$ rows and $C_2$ has $n_u$ rows; and
	$(C_w,\hat{B}_w,\hat{D}_w)=(C_2,\begin{bmatrix} 0 & \hat{B} \end{bmatrix}, \begin{bmatrix} 0 & \hat{D} \end{bmatrix})$
	if $\w=\mathbf{u}$, and 
	$(C_w,\hat{B}_w,\hat{D}_w)=(C_g,\hat{B},\hat{D})$, if $\w=\begin{bmatrix} \y^T & \mathbf{u}^T \end{bmatrix}^T$.
 Choose for all $f\in\mathcal{F}$, $\hat{M}(f)>1$, and $\hat{\gamma}(f)\in[\hat{\gamma}^*(f),1)$, such that $\|\hat{A}^k\|_2\leq \hat{M}(f)\hat{\gamma}^k(f)$, with $\hat{\gamma}^*(\hat{A})$ the spectral radius of $\hat{A}$.
	With these definitions, 
  \begin{align*}
	    &G_e(f)\hspace{-2pt}=\hspace{-2pt}\|(A_e,K_e,C_e,D_e)\|_{\ell_1}\hspace{-2pt}\triangleq\hspace{-2pt}\|D_e\|_2 \hspace{-2pt}+\hspace{-2pt}\sum_{k=0}^{ \infty} \|C_eA_e^{k}K_e\|_2 \nonumber \\
        &\|\Sigma_{gen}\|_{\ell_1}=1+\sum_{k=0}^\infty \|C_gA_g^{k-1}K_g\|_2\\
        &\bar{G}_{gen}=\|\Sigma_{gen}\|_{\ell_1}^{2} \mu_{\max}(Q_e)\\
        &\bar{G}_f(f)=\left (1+\|\hat{D}\|+ \frac{\hat{M}\|\hat{B}\|\|\hat{C}\|}{1-\hat{\gamma}} \right ) \frac{\hat{M}\|\hat{C}\| \|\hat{B}\|}{(1-\hat{\gamma})^{1.5}}
   \end{align*}
\end{Definition}
The interpretation of the various terms appearing in  Definition \ref{def:constants} is as follows.
\begin{Remark}[Interpretation of constants]\text{}
\\
\textbf{The matrices $A_e,K_e,C_e,D_e$} represent the LTI system driven
  by the innovation process $\e_g$ of $(\y^T,\w^T)^T$, output
  of which is $\y-\hat{\y}_f$, i.e.,
\begin{equation}
\label{error-sys1}
\begin{split}
   \tilde{\x}(t+1)=A_e\tilde{\x}(t)+K_e\e_g(t), \\
   \y(t)-\hat{\y}_f(t)=C_e\tilde{\x}(t)+D_e\e_g(t)
\end{split}  
\end{equation}

\textbf{The term $\bar{G}_{\text{gen}}$} depends only on the data generator system \eqref{eq:generator}, and characterises the scaling of $\y,\rvu$  

\textbf{The term $\bar{G}_f(f)$} depends only the predictor $f$, and should be interpreted similarly to $\|(\hat{A},\hat{B},\hat{C},\hat{D})\|_{\ell_1}^2$.
\end{Remark}

\begin{Theorem}\label{thm:unbounded} Let $\mathcal{M}_\pi$ denote the set of all absolutely continuous densities w.r.t $\pi$. Then for any density $\pi$ on hypothesis class $\mathcal{F}$, any $\delta\in(0,1]$, and
    \begin{multline} 0<\lambda < \Big (\sup_{f\in\mathcal{F}} \max\{8 (n_u+n_y) \bar{G}_{gen}  \bar{G}_f(f),  \\ 
    6(n_u+n_y+1) n_y \mu_{\max}(Q_e)G_e(f)^{2} \}\Big )^{-1} \label{eq:lambdaBound}
    \end{multline}
     the following inequality holds with probability at least $1-2\delta$
    \begin{multline}
         \forall\rho\in\mathcal{M}_\pi:\quad E_{f\sim \hat{\rho}} \mathcal{L} (f) \le \E_{f\sim \hat{\rho}} \hat{\mathcal{L}}_N(f) + r_N(\lambda,N), \label{eq:NewKLBound}
    \end{multline}
      with
      \begin{align}
          r_N(\lambda,N)&\triangleq \dfrac{1}{\lambda}\!\left[\KL(\hat{\rho} \|\pi) + \ln\dfrac{1}{\delta}	+ \widehat{\Psi}_{\pi}(\lambda,N) \right ]\\
          \widehat{\Psi}_{\pi}(\lambda,N)&\triangleq \frac{1}{2}\left (\widehat{\Psi}_{\pi,1}(\lambda,N)+\widehat{\Psi}_{\pi,2}(\lambda,N)\right ) \\
          \widehat{\Psi}_{\pi}(\lambda,N) &\geq \Psi_{\pi}(\lambda,N)= \ln E_{f\sim\pi} \bE[e^{\lambda(\mathcal{L}(f)-\hat{\mathcal{L}}_{N}(f))}] \nonumber
      \end{align}
      and
      \begin{align}
        \widehat{\Psi}_{\pi,1}(\tilde{\lambda},N) &\triangleq \ln E_{f\sim\pi} \left ( 1+\frac{1}{N}C_1(f,\lambda) \right )\\
        \widehat{\Psi}_{\pi,2}(\tilde{\lambda},N) &\triangleq \ln E_{f\sim\pi } \left ( 1+ \frac{1}{\sqrt{N}} C_2(f,\lambda) \right )\\
        C_1(f,\lambda)&\triangleq \frac{2(m+1)! \left (6\lambda n_y\mu_{\max}(Q_e)G_e(f)^{2}\right )^2}{(1-6(m+1)\lambda n_y\mu_{\max}(Q_e)G_e(f)^{2})} \\ 
        C_2(f,\lambda)&\triangleq \frac{8 (m!)\lambda \bar{G}_{gen}  \bar{G}_f(f)}{1-8 \lambda m\bar{G}_{gen}  \bar{G}_f(f)}
      \end{align}
\end{Theorem}
For proof of Theorem \ref{thm:unbounded}, see Proof \ref{proof:thm:unbounded}, in the Appendix. 

Note that, as $N\to \infty $ the PAC-Bayesian error $r_N\to \frac{1}{\lambda}\left (\KL(\rho|\pi)+\ln \left (\frac{1}{\delta}\right ) \right )$. That is, irrespective of $\rho,\pi$, the error $r_N\geq \frac{1}{\lambda}\ln \left (\frac{1}{\delta}\right )$. Usually, one chooses $\lambda=\lambda(N)$ as an increasing function of $N$, which then allows the PAC-Bayesian error to converge to 0. However, since by Theorem \ref{thm:unbounded}, $\lambda$ is bounded by a constant, we can not control the term $\frac{1}{\lambda}\ln \left (\frac{1}{\delta}\right )$, and $r_N>0$ always. 
\begin{Remark}
    Theorem \ref{thm:unbounded}, holds under assumption \ref{as:generator}, for any distribution of $\e_g(t)$, as long as
    \begin{itemize}
        \item $\e_g(t)\in\reals^m$ is zero-mean, i.i.d.,
        \item $\bE\left [ \|\e_g(t)\|^{2r} \right ]\leq 2^r\mu_{\max}(Q_e)^{r}(m+r-1)!$,
        \item $\sigma(r)\leq 3^r\mu_{\max}(Q_e)^{r}(m+r-1)!$,
    \end{itemize}
    with
    \begin{multline*}
        \sigma(r)=\sup_{t,k,j} \bE\left [\| \e(t,k,j) \|^r_2   \right ],\\
        \e(t,k,j)\triangleq\bE[\e_g(t-k)\e_g^T(t-j)] - \e_g(t-k)\e_g^T(t-j)
    \end{multline*}
    That is, Theorem \ref{thm:unbounded} holds, for $\e_g(t)$, zero mean, i.i.d. with any sub-gaussian distribution. 
\end{Remark}

\section{Bounded case} \label{sec:boundedResults}
If we drop the assumption that $\e_g(t)$ has a Gaussian distribution, and only assume that $\e_g(t)$ is bounded, we get quite straight-forward PAC-Bayesian bounds. 
\begin{Assumption}\label{assumption}
    $\e_g(t)$ is a zero mean i.i.d. stochastic process, with arbitrary distribution, but for all components $\e_{g,i}(t)$ of $\e_g(t)$
 %   \begin{align}
        $|\e_{g,i}(t)|\leq c_e$,
  %  \end{align}
     for some $c_e>0$.
\end{Assumption} 
\begin{Theorem}\label{thm:bounded} Let $\mathcal{M}_\pi$ denote the set of all absolutely continuous densities w.r.t $\pi$. Under assumption \ref{assumption} it holds true that for any density $\pi$ on hypothesis class $\mathcal{F}$, any $\delta\in(0,1]$, and $\lambda>0$ the following inequality holds with probability at least $1-2\delta$
    \begin{multline}
        \forall \rho\in\mathcal{M}_\pi:\quad E_{f\sim\rho} \mathcal{L}(f) \leq E_{f\sim\rho }\hat{\mathcal{L}}_N(f)+\bar{r}_N(\lambda,N) 
    \end{multline}
    with 
    \begin{align}
        \bar{r}_N(\lambda,N) &\triangleq \frac{1}{\lambda}\left [\KL(\rho||\pi)+\ln\frac{1}{\delta} + \widehat{\Psi}_{c_e,\pi}(\lambda,N)\right ]\\
        \widehat{\Psi}_{c_e,\pi}(\lambda,N) &\triangleq \frac{1}{2} \left ( \widehat{\Psi}_{c_e,\pi,1}(\lambda,N)+\widehat{\Psi}_{c_e,\pi,2}(\lambda,N)\right )\\
        \widehat{\Psi}_{c_e,\pi,1}(\lambda,N) &\triangleq \ln E_{f\sim\pi } \left ( 1+\frac{1}{N}e^{ \lambda G_{gen,1} G_e(f)^2} \right )\\
        \widehat{\Psi}_{c_e,\pi,2}(\lambda,N) &\triangleq \ln E_{f\sim\pi } \left (1+\frac{1}{\sqrt{N}}e^{\lambda G_{gen,2} \bar{G}_f(f) } \right )
    \end{align}
    and
    \begin{align}
        G_{gen,1}&\triangleq 8c_e^2n_y(n_y+n_u)\\
        G_{gen,2}& \triangleq 4\|\Sigma_{gen}\|_{\ell_1}^2 c_e^2 (n_y+n_u)
    \end{align}
% with probability at least $1-\delta$
%     \begin{multline}
%          \forall\rho\in\mathcal{M}_\pi :\quad E_{f\sim \hat{\rho}} \mathcal{L} (f) \leq   E_{f\sim \hat{\rho}} \hat{\mathcal{L}}_N(f) \\
%          +\dfrac{1}{\lambda}\!\left[\KL(\hat{\rho} \|\pi) + \ln\dfrac{1}{\delta}	+ \Psi_{\pi,c_e}(\lambda,N) \right ], \label{eq:NewKLBound}
%     \end{multline}
%       with
%     \begin{align}
%         &\Psi_{\pi,c_e}(\lambda,N)=	\ln E_{f\sim\pi} \bE[e^{\lambda(\mathcal{L}(f)-\hat{\mathcal{L}}_N(f))}] \nonumber \\
%         &\leq \ln  E_{f\sim\pi}\left ( 1+ \frac{1}{2N} e^{\lambda G_{gen,1} G_e(f)^{2}} + \frac{1}{2\sqrt{N}} e^{\lambda \bar{G}_f(f) G_{gen,2}} \right )
%     \end{align}
%     and
%     \begin{align}
%         G_{gen,1}&\triangleq 8c_e^2 n_y (n_y+n_u) \\
%         G_{gen,2}&\triangleq 4  c_e^2 (n_y+n_u) \|\Sigma_{gen}\|_{\ell_1}^2
%     \end{align}
\end{Theorem}
For proof of Theorem \ref{thm:bounded}, see Corollary \ref{cor:thm:bounded}, in the Appendix.
Note that, in this case $\lambda$ is not bounded, and as such we can choose $\lambda=\lambda(N)$ an increasing function of $N$, in order to control the term $\frac{1}{\lambda(N)}\ln \delta^{-1}$. More specifically one can choose 
\begin{align}
    \lambda(N)&=\frac{\ln\sqrt{N}}{\sup_{f\in\mathcal{F}} \max \{G_{gen,1} G_e(f)^2, G_{gen,2} \bar{G}_f(f) \}},
\end{align}
for which, it can be shown that $\lambda^{-1}(N)\Psi_{\pi,c_e}(\lambda(N),N)\to 0$, and $\lambda^{-1}(N)\ln\delta^{-1}\to 0$. If one considers $\rho$ independently of $\lambda$, then $\lambda^{-1}(N)\KL(\hat{\rho} \|\pi)\to 0$, however if one considers Gibbs posteriors \eqref{eq:gibbs}, which do depend on $\lambda$, then it is hard to say what will happen with  $\lambda^{-1}(N)\KL(\hat{\rho} \|\pi)$. Simulations seem to indicate that if $\lambda(N)$ is any reasonable increasing function of $N$, then $\lambda(N)$ will converge to some problem dependant constant.

The bound above has all the desired properties,
but its rate of convergence to zero as $N \rightarrow +\infty$ is very slow. In fact, using
\cite{alquier2013prediction}, the results of Theorem \ref{thm:bounded} can be sharpened as follows.
%\section{Alternative Bounded case, based on µ\cite{alquier2013prediction}}
%\textcolor{red}{TO DO: intro to alt case}
\begin{Theorem}\label{thm:bounded_alt} Let $\mathcal{M}_\pi$ denote the set of all absolutely continuous densities w.r.t $\pi$. Under assumption \ref{assumption} it holds true that for any density $\pi$ on hypothesis class $\mathcal{F}$, any $\delta\in(0,1]$, and $\lambda>0$ the following inequality holds with probability at least $1-2\delta$
    \begin{multline}
        \forall \rho\in\mathcal{M}_\pi:\quad E_{f\sim\rho} \mathcal{L}(f) \leq E_{f\sim\rho }\hat{\mathcal{L}}_N(f)+\tilde{r}_N(\lambda,N) 
    \end{multline}
    with 
    \begin{align}
        \tilde{r}_N(\lambda,N) &\triangleq \frac{1}{\lambda}\left [\KL(\rho||\pi)+\ln\frac{1}{\delta} + \tilde{\Psi}_{c_e,\pi}(\lambda,N)\right ]\\
        \tilde{\Psi}_{c_e,\pi}(\lambda,N) &\triangleq \frac{1}{2} \left ( \tilde{\Psi}_{1}(\lambda,N)+\tilde{\Psi}_{2}(\lambda,N)\right )\\
        \tilde{\Psi}_{1}(\lambda,N) &\triangleq \ln E_{f\sim\pi } \left ( 1-C_{1,2}(f)+ C_{1,2}(f) e^{\frac{\lambda}{N}C_{1,1}(f)}  \right )\\
        \tilde{\Psi}_{2}(\lambda,N) &\triangleq \ln E_{f\sim\pi} \left(
        e^{\frac{\lambda^2}{N}C_2(f) }\right) 
    \end{align}
    and, with $C\triangleq c_e\sqrt{n_u+n_y}$,
    \begin{align}
        C_{1,1}(f)&\triangleq 2\|\Sigma_{gen}\|_{\ell_1} C \bar{G}_{f,2}(f)\\
        C_{1,2}(f)&\triangleq \bar{G}_{f,1}(f) \|\Sigma_{gen}\|_{\ell_1} C \\
        C_2(f)&\triangleq 8(G_e(f)+G_{e,1}(f))^2C^2 (4G_e(f)C+1)^2\\
        G_{e,1}&\triangleq \|D_e\|_2 + \sum_{k=0}^\infty (k+1)\|C_eA_e^kK_e\|_2
    \end{align}
\end{Theorem}
For proof of Theorem \ref{thm:bounded_alt}, see Proof \ref{proof:thm:bounded_alt}, in the Appendix. 
If $\lambda_N=\sqrt{N}$ is chosen, then 
the error bound $\bar{r}_N(\lambda_N)$
above converges to zero as $N \rightarrow \infty$ 
at a rate $O(\frac{1}{\sqrt{N}})$.

\section{Numerical example}\label{sec:numEx}
For the sake of illustration let us assume that data is generated by
\begin{align*}
    \x(t+1)&=\begin{bmatrix} 0.16 & -0.3 \\ 0 & -0.05 \end{bmatrix} \x(t) + \begin{bmatrix} 0.33 & -0.75 \\ 0 & -0.09 \end{bmatrix} \e_g(t)\\
    \begin{bmatrix} \y(t) \\ \rvu(t) \end{bmatrix} &= \begin{bmatrix}
        1 & 1 \\ 0 & 1 \end{bmatrix} \x(t) + \e_g(t),
\end{align*}
Following the two theorems in the paper, we will consider two cases
\begin{itemize}
    \item Unbounded innovation noise: $\e_g(t)\sim\mathcal{N}(0,Q_e)$, 
    \begin{align}
        Q_e=\begin{bmatrix} 0.054 &  0.018\\ 0.018 & 0.248\end{bmatrix}
    \end{align}
    \item Bounded innovation noise: $\e_g(t)$ is distributed according to zero-mean truncated gaussian, s.t. $c_e=1$, and 
    \begin{align}
        \bE[\e_g(t)\e_g^T(t)]\approx Q_e
    \end{align}
    
\end{itemize}
We will assume that the predictors are fully parameterised, i.e. for the case of $\w(t)=\rvu(t)$
\begin{align*}
    \hat{A}(\theta)=\begin{bmatrix} \theta_1&\theta_2\\ \theta_3&\theta_4\end{bmatrix} \;
    \hat{B}(\theta)=\begin{bmatrix} \theta_5\\ \theta_6\end{bmatrix} \\
    \hat{C}(\theta)=\begin{bmatrix} \theta_7&\theta_8\end{bmatrix} \;
    \hat{D}(\theta)=\begin{bmatrix} \theta_9\end{bmatrix}
\end{align*}
for the case of $\w(t)=[\y^T(t), \rvu^T(t)]^T$
\begin{align*}
    \hat{A}(\theta)=\begin{bmatrix} \theta_1&\theta_2\\ \theta_3&\theta_4\end{bmatrix} \;
    \hat{B}(\theta)=\begin{bmatrix} \theta_{10} & \theta_5\\ \theta_{11}& \theta_6\end{bmatrix} \\
    \hat{C}(\theta)=\begin{bmatrix} \theta_7&\theta_8\end{bmatrix} \;
    \hat{D}(\theta)=\begin{bmatrix} 0& \theta_9\end{bmatrix}
\end{align*}
Thus, with $\Sigma(\theta)=(\hat{A}(\theta),\hat{B}(\theta),\hat{C}(\theta),\hat{D}(\theta))$, we will define our hypothesis class to be 
$$\mathcal{F}=\{f_{\Sigma(\theta)}|\gamma(\hat{A}(\theta))<1, \bar{G}_f(f)<10, \theta\in\sR^{11}\}$$
The prior is given by
\begin{align}
    \pi(f)=Z_\pi \exp(-\bar{G}_f(f))
\end{align}
with $Z_\pi$ the normalisation term. This prior will act as regularisation, penalising predictors with high $\ell_1$ norms.
We will use the Gibbs posterior
\begin{align}
    \rho(f|N)=Z_\rho \pi(f) \exp(-\lambda(N) \hat{\mathcal{L}}_N(f))
\end{align}
In order to compute the numerical value of $r_N$, we can use Markov-Chain Monte-Carlo methods, which means that we only need to be able to evaluate
\begin{align}
    \hat{\pi}(f)=\exp(-\bar{G}_f(f)) \propto \pi(f)\\
    \hat{\rho}(f)=\hat{\pi}(f)\exp(-\lambda \hat{\mathcal{L}}_N(f))\propto \rho(f)
\end{align}
More precisely one can approximate $r_N$, by only being able to evaluate $\hat{\pi}(f)$ and 
$
    \beta(f)\triangleq\frac{\hat{\rho}(f)}{\hat{\pi}(f)} \propto \frac{\rho(f)}{\pi(f)}
$

\begin{figure}[ht]
    \centering
    \includegraphics[width=0.99\linewidth]{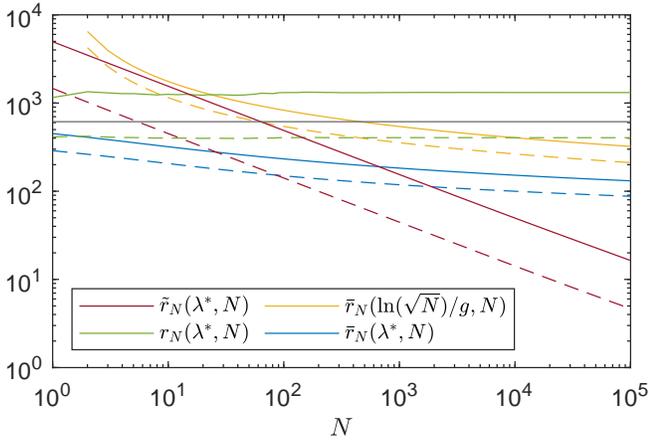}
    \caption{Numerical simulation of both cases (bounded and unbounded noise), solid lines depict case of $\w=\rvu$, dashed lines show case of $\w=[\y^T,\rvu^T]^T$, $\lambda^*$ is found by numerical optimisation, i.e. $\lambda^*=\arg\min_\lambda r_N(\lambda,N)$, the black horizontal line denotes a vacuous bound for the bounded noise case, i.e. any bounds above that line are vacuous}
    \label{fig:rN}
\end{figure}
In Figure \ref{fig:rN} we see the convergence of the error term, for the case of bounded noise. Note that the proposed function $\lambda(N)$ is close to numerically optimal (blue line in Figure \ref{fig:rN}), asymptotically $\lambda(N)\propto \ln\sqrt{N}$, seem to be optimal, one could try to find a less conservative scaling $g(\mathcal{F})< \sup_{f\in\mathcal{F}} \max \{G_{gen,1} G_e(f)^2, G_{gen,2} \bar{G}_f(f) \}$. For the proposed PAC-Bayesian bounds to be useful, the bounds should convergence faster than $\mathcal{O}(\frac{1}{\ln\sqrt{N}})$, since in most applications collecting $N=10^{10}$ data points is not feasible. 
Note that for $N\leq 460$, for this system Theorem \ref{thm:bounded}, yields vacuous bounds, i.e. $\bar{r}_N\geq 2(C\sup_{f\in\mathcal{F}}G_e(f))^2$. However for Theorem \ref{thm:bounded_alt}, only for $N\leq 64$, is the bound vacuous.

 For the case of unbounded innovation noise, as stated before we see in Figure \ref{fig:rN} that it converges to a constant. Unfortunately, since $\lambda$ is bounded not much can be done. However, since the noise is unbounded it is difficult to determine if the bound is vacuous.

% \begin{figure}[ht]
%     \centering
%     \includegraphics[width=0.99\linewidth]{Psi_plot_both.eps}
%     \caption{Numerical simulation of the bounded and unbounded case, solid lines depict case of $\w=\rvu$, dashed lines show case of $\w=[\y^T,\rvu^T]^T$, $\lambda^*$ is found by numerical optimisation, i.e. $\lambda^*=\arg\min_\lambda r_N(\lambda,N)$}
%     \label{fig:Psi}
% \end{figure}
% Focusing on the problem dependant terms $\widehat{\Psi}$, in Figure \ref{fig:Psi}, we see that they converge to $0$ at the rate of $\mathcal{O}(\frac{1}{\sqrt{N}})$, meaning that the slow convergence of $\mathcal{O}(\frac{1}{\ln\sqrt{N}})$, is induced more from the general framework of the PAC-Bayesian bounds, i.e. $\frac{1}{\lambda}\KL(\rho|\pi)$, and $\frac{1}{\lambda}\ln\frac{1}{\delta}$. 

\section{Conclusion}
\label{sect:concl}
In this paper we have 
derived two PAC-Bayesian error bounds
for stochastic LTI systems with inputs. For data generated by an LTI system with sub-gaussian noise, we see that the difference between empirical and generalised loss is bounded from below, which intuitively should not be the case. Thus, more work needs to be done, to obtain less conservative bounds, or use a difference approach, i.e. one can derive PAC-Bayesian type bounds based on different change of measure inequalities. 

For data generated by an LTI system with bounded innovation noise, we have that the difference between empirical and generalised loss will convergence to 0, slowly at the rate of $\mathcal{O}(\frac{1}{\ln\sqrt{N}})$. That is the problem of minimising the empirical loss, becomes equivalent to minimising the generalised loss, at the aforementioned rate. 

Future research will be directed towards extending these results to more general state-space representations and using the results of the paper for deriving oracle inequalities \cite{alquier2021userfriendly}.

% \bibliography{bib}
% \bibliographystyle{ieee}
\printbibliography

\newpage

\onecolumn
\appendix
\renewcommand{\theequation}{A.\arabic{equation}} 
\renewcommand{\theProof}{A.\arabic{Proof}} 
\renewcommand{\thelemma}{A.\arabic{lemma}} 
\renewcommand{\theRemark}{A.\arabic{Remark}}
\renewcommand{\theTheorem}{A.\arabic{Theorem}}
\renewcommand{\theCorollary}{A.\arabic{Corollary}}

\subsection{Proofs}
In this section we provide the proofs of theorem \ref{thm:unbounded} and \ref{thm:bounded} under the assumptions stated in the main text. To do so we first prove a series of lemmas. 
%In this section we state, necessary lemmas with proofs, following assumptions made in main text, in order to state proofs of theorem \ref{thm:pac:altKL} and \ref{thm:pac:alt1}.
\newcommand{\Efrho}{E_{f\sim\rho}}
\newcommand{\Efpi}{E_{f\sim\pi}}

%	\begin{equation}
%	P\left (\omega\in\Omega \mid E_{\theta\sim\rho}\mathcal{L}(\theta) \leq E_{\theta\sim\rho}V(\theta)+\left (E_{\theta\sim\pi} \left ( \frac{\rho(\theta)}{\pi(\theta)} \right )^\alpha \right)^\frac{1}{\alpha}\left (E_{\theta\sim\pi}\bE \left [ (\mathcal{L}(\theta)-V(\theta))^\frac{\alpha}{\alpha-1} \right ] \right )^{\frac{\alpha-1}{\alpha}} \right )
%\end{equation} 
% with $\phi(f)=\lambda(\mathcal{L}(f)-V_N(f))$, $\lambda\in\reals,\lambda>0$ it follows

\begin{lemma} \label{lemma:etoz} For random variable $\e_g(t)\sim\mathcal{N}(0,Q_e)$, the following holds
	\begin{align*} 
		\bE[\|\e_g(t)\|_2^r]\leq \mu_{\max}(Q_e)^{\frac{r}{2}}\bE[\|\z(t)\|_2^r]\\
		\z(t)\sim \mathcal{N}(0,I),
	\end{align*}
	where $Q_e=\bE[\e_g(t)\e_g^T(t)]$, and $\mu_{\max}(Q_e)$ denotes the maximal eigen value of $Q_e$.
\end{lemma}
\begin{Proof}[Proof of Lemma \ref{lemma:etoz}]
	First, note $\z(t)=Q_e^{-\frac{1}{2}}\e_g(t)$, and 
	\begin{align*}
		\|\e_g(t)\|_2^2=\e_g^T(t)\e_g(t)=\z^T(t)Q_e^{\frac{1}{2}}Q_e^{\frac{1}{2}}\z(t)=\z^T(t)Q_e\z(t)
	\end{align*}
	therefore
	\begin{align*}
		\|\e_g(t)\|_2^2 \leq \mu_{\max}(Q_e)\|\z(t)\|_2^2\\
		\|\e_g(t)\|_2^r\leq\mu_{\max}(Q_e)^{\frac{r}{2}}\|\z(t)\|_2^r\\
		\bE[\|\e_g(t)\|_2^r]\leq\mu_{\max}(Q_e)^{\frac{r}{2}}\bE[\|\z(t)\|_2^r]\\
	\end{align*}
	Finally, note that $\z(t) \sim  \mathcal{N}(0,I)$.
\end{Proof}
\begin{lemma} \label{lemma:zMomentsSq}
	If $\z(t)\sim \mathcal{N}(0,I_m)$, then
	\begin{align*}
		\bE[\|\z(t)\|_2^r]^2\leq 4((m+r-1)!)
	\end{align*}
\end{lemma}

\begin{Proof}[Proof of Lemma \ref{lemma:zMomentsSq}]
	First, notice that the distribution of $\|\z(t)\|_2=\sqrt{\sum_{i=1}^m \z_i^2(t)}$ is chi- distribution, as such
	\begin{align}
		\bE[\|\z(t)\|_2^r]=2^{\frac{r}{2}}\dfrac{\Gamma(\frac{m+r}{2})}{\Gamma(\frac{m}{2})}
	\end{align}
	We will use mathematical induction to prove the lemma.\\
	\textbf{For $r=0$}, lemma holds, since
	\begin{align}
		\bE[\|\z(t)\|_2^0]^2=\left (2^{\frac{0}{2}}\dfrac{\Gamma(\frac{m+0}{2})}{\Gamma(\frac{m}{2})}\right )^2=1\leq 4(m-1)! 		,\quad \forall m\in\mathbb{N}.
	\end{align}
	\textbf{for $r=1$}, lemma holds, as
	\begin{align*}
		\bE[\|\z(t)\|_2^1]=2^{\frac{1}{2}}\dfrac{\Gamma(\frac{m+1}{2})}{\Gamma(\frac{m}{2})}.
	\end{align*}
	Notice that, for scalar $\x\sim\mathcal{N}(0,1)$
	\begin{align*}
		\bE[|\x|^k]=2^{\frac{k}{2}}\dfrac{\Gamma(\frac{k+1}{2})}{\sqrt{\pi}}
	\end{align*}
	It is also known that 
	\begin{align*}
		\bE[|\x|^k]=\begin{cases} (k-1)!!\sqrt{\frac{2}{\pi}},& k\text{ odd}\\ (k-1)!!,& k\text{ even} \end{cases}
	\end{align*}
	therefore,
	\begin{align*}
		2^{\frac{k}{2}}\dfrac{\Gamma(\frac{k+1}{2})}{\sqrt{\pi}}=\begin{cases} (k-1)!!\sqrt{\frac{2}{\pi}},& k\text{ odd}\\ (k-1)!!,& k\text{ even} \end{cases}
	\end{align*}
	Applying this to $k=m$ and $k=m-1$, we obtain
	\begin{align*}
		2^{\frac{m}{2}}\dfrac{\Gamma(\frac{m+1}{2})}{\sqrt{\pi}}=\begin{cases} (m-1)!!\sqrt{\frac{2}{\pi}},& m\text{ odd}\\ (m-1)!!,& m\text{ even} \end{cases}\\
		2^{\frac{m-1}{2}}\dfrac{\Gamma(\frac{m}{2})}{\sqrt{\pi}}=\begin{cases} (m-2)!!\sqrt{\frac{2}{\pi}},& (m-1)\text{ odd}, (m \text{ even})\\ (m-2)!!,& (m-1)\text{ even}, (m \text{ odd}) \end{cases}
	\end{align*}
	Now notice,
	\begin{align*}
		\bE[\|z(t)\|_2^1]=2^{\frac{1}{2}}\dfrac{\Gamma(\frac{m+1}{2})}{\Gamma(\frac{m}{2})} = \frac{ 2^{\frac{m}{2}}\dfrac{\Gamma(\frac{m+1}{2})}{\sqrt{\pi}}}{2^{\frac{m-1}{2}}\dfrac{\Gamma(\frac{m}{2})}{\sqrt{\pi}}}=\frac{(m-1)!!}{(m-2)!!}c_m\\
		c_m=\begin{cases}\sqrt{\frac{2}{\pi}} ,& m\text{ even}\\\sqrt{\frac{\pi}{2}} ,& m\text{ odd} \end{cases}
	\end{align*}
	notice that $c_m\leq 2$ for all $m$, and therefore
	\begin{align}
		\bE[\|\z(t)\|_2^1]\leq 2\frac{(m-1)!!}{(m-2)!!}\leq 2(m-1)!!
	\end{align}
	Then
	\begin{align*}
		\bE[\|\z(t)\|_2^1]^2\leq 4((m-1)!!)^2
	\end{align*}
	Note that $((m-1)!!)^2\leq m!$. We can see that by contradiction: assume that
	%\begin{align*}
		$((m-1)!!)^2\geq m !$. Notice that $m!=m!!(m-1)!!$ and hence $((m-1)!!)^2\geq m !$
		implies 
		%$((m-1)!!)^2\geq m!!(m-1)!!$
		$(m-1)!!\geq m!!$.
	%\end{align*}
	As $(m-1)!!$ must be less than $m!!$ we have a contradiction. Therefore $((m-1)!!)^2\leq m!$ holds and we have
	\begin{align*}
		\bE[\|\z(t)\|_2^1]^2\leq 4 m!.
	\end{align*}
	That is, we have shown that for $r=0$ and $r=1$ Lemma \ref{lemma:zMomentsSq} holds. \\
	 Now suppose that for all $k \geq 2$ and for all  $0\leq r \leq k$
	\begin{align}
		2^{\frac{r}{2}}\dfrac{\Gamma(\frac{m+r}{2})}{\Gamma(\frac{m}{2})}\leq 4(m+r-1)!,
		\label{pf:b2:eq-1}
	\end{align}
	We will show that \eqref{pf:b2:eq-1} holds for $r=k+1$ too.
	To this end, notice that 
	\begin{align*}
		\Gamma \left (\frac{m+k}{2} \right )=\Gamma \left(\frac{m+k-2}{2}+1 \right)=\frac{m+k-2}{2}\Gamma \left(\frac{m+k-2}{2} \right)
	\end{align*}
	Using this relation we obtain
	\begin{equation}
	\label{pf:b2:eq1}
		\begin{split}
			\left (2^{\frac{k}{2}}\dfrac{\Gamma(\frac{m+k}{2})}{\Gamma(\frac{m}{2})}\right )^2=\left ( \left ( 2^{\frac{k-2}{2}}\dfrac{\Gamma(\frac{m+k-2}{2})}{\Gamma(\frac{m}{2})} \right )\left (2\frac{m+k-2}{2} \right) \right )^2 \\
			=\left ( 2^{\frac{k-2}{2}}\dfrac{\Gamma(\frac{m+k-2}{2})}{\Gamma(\frac{m}{2})} \right )^2\left (2\frac{m+k-2}{2}  \right )^2.
		\end{split}
	\end{equation}	
	Now $k-2\in[0,k]$, so we can apply to it the induction hypothesis. That is,
	for $r=k-2$, \eqref{pf:b2:eq-1} holds, i.e., 
	\begin{align*}
		\left ( 2^{\frac{r}{2}}\dfrac{\Gamma(\frac{m+r}{2})}{\Gamma(\frac{m}{2})} \right )\leq 4(m+r-1)!=4(m+k-3)!.
	\end{align*}
	%then we have $4(m+r-1)!\leq 4(m+k-3)!$, as such currently we have 
	and therefore
	\begin{align*}
		\left (2^{\frac{k}{2}}\dfrac{\Gamma(\frac{m+k}{2})}{\Gamma(\frac{m}{2})}\right )^2\leq 4(m+k-3)! \left ( 4\frac{(m+k-2)^2}{4} \right )\\
		=4(m+k-3)!(m+k-2)(m+k-2).
	\end{align*}
	%Substituting this into \eqref{pf:b2:eq1} a
    Using $(m+k-2)\leq (m+k-1)$, it follows that 
	\begin{align*}
	 \left ( 2^{\frac{k-2}{2}}\dfrac{\Gamma(\frac{m+k-2}{2})}{\Gamma(\frac{m}{2})} \right )^2\left (2\frac{m+k-2}{2}  \right )^2 \le 
		4(m+k-3)!(m+k-2)(m+k-2)	\leq 4(m+k-1)!
	\end{align*}
	 Substituting the last inequality into \eqref{pf:b2:eq1}, it follows that
	 \eqref{pf:b2:eq-1} holds for $r=k+1$. 
\end{Proof}
\begin{lemma}\label{lemma:evenzmoments} For random variable $\z\sim\mathcal{N}(0,I_m)$, the even moments of $\|\z\|_2$ are bounded by
	\begin{align*}
		\bE[\|\z\|_2^{2r}]\leq 2^r(m+r-1)!
	\end{align*}
	\begin{Proof}[Proof of Lemma \ref{lemma:evenzmoments}]
		Clearly $\|\z\|_2$ has the chi distribution,
		\begin{align*}
			\bE[\|\z\|_2^{2r}]=2^{\frac{2r}{2}}\dfrac{\Gamma(\frac{m+2r}{2})}{\Gamma(\frac{m}{2})}=2^{r}\dfrac{\Gamma(\frac{m}{2}+r)}{\Gamma(\frac{m}{2})}
		\end{align*}
		\begin{align*}
			\Gamma\left (\frac{m}{2}+r \right )=\Gamma \left (\frac{m}{2}+(r-1) +1 \right )= \left (\frac{m}{2}+(r-1) \right )\Gamma \left (\frac{m}{2}+(r-1) \right )\\
			=\left (\frac{m}{2}+(r-1) \right )\left (\frac{m}{2}+(r-2) \right ) \dots \frac{m}{2}\Gamma\left ( \frac{m}{2} \right )
		\end{align*}
		\begin{align*}
			\bE[\|\z\|_2^{2r}]=2^r \frac{\left (\frac{m}{2}+(r-1) \right )\left (\frac{m}{2}+(r-2) \right ) \dots \frac{m}{2}\Gamma\left ( \frac{m}{2} \right )}{\Gamma\left ( \frac{m}{2} \right )}
		\end{align*}
		notice $\frac{m}{2}\leq m$, then
		\begin{align*}
			\bE[\|\z\|_2^{2r}]\leq 2^r\frac{(m+r-1)!}{m!} \leq 2^r (m+r-1)!
		\end{align*}
	\end{Proof}
\end{lemma}
Combining Lemmas (\ref{lemma:etoz} and \ref{lemma:zMomentsSq}), we obtain the following lemma.
\begin{lemma} \label{lemma:evenEmoments} Let $r\in\mathbb{N}$
	\begin{align*}
		\bE[\|\e_g(t)\|_2^{2r}]\leq \mu_{\max}(Q_e)^{r}2^r(m+r-1)!
	\end{align*}
\end{lemma}
Combining Lemmas (\ref{lemma:etoz} and \ref{lemma:evenzmoments}), we obtain the following lemma.
\begin{lemma} \label{lemma:Emoments} Let $r\in \{1,3,5,\dots\}$
	\begin{align*}
		\bE[\|\e_g(t)\|_2^r]\leq 2\mu_{\max}(Q_e)^{\frac{r}{2}}\sqrt{(m+r-1)!}
	\end{align*}
\end{lemma}

\begin{lemma}\label{lemma:stationary4thMoment} Let $\z(t)$ be any stationary process, and $r\in\mathbb{N}$, then for a stochastic process $\mathbf{s}(t)=\sum_{k=0}^\infty \alpha_k \z(t-k),$
with $\sum_{k=0}^\infty \|\alpha_k\|\leq +\infty$, the following holds
\begin{equation}
    \bE[\|\mathbf{s}(t)\|^r]\leq \left ( \sum_{k=0}^\infty \| \alpha_k\| \right )^r \bE[\|\z(t)\|^r]
\end{equation}
\end{lemma}
\begin{Proof}[of Lemma \ref{lemma:stationary4thMoment}]
    \newcommand{\m}{r} %in case proof holds for any moment
    \begin{multline}
        \bE[\|\mathbf{s}(t)\|^\m]=\bE \left [\|\sum_{k=0}^\infty \alpha_k \z(t-k)\|^\m \right ] \leq \bE\left [\left (\sum_{k=0}^\infty \|\alpha_k\| \|\z(t-k)\| \right )^\m \right ]\\
        =\bE\left [ \sum_{k_1=0}^\infty \dots \sum_{k_\m=0}^\infty \left ( \prod_{i=1}^\m \|\alpha_{k_i}\| \prod_{i=0}^\m \|\z(t-k_i)\| \right ) \right ] =  \sum_{k_1=0}^\infty \dots \sum_{k_\m=0}^\infty \left ( \prod_{i=1}^\m \|\alpha_{k_i}\| \bE\left [ \prod_{i=0}^\m  \|\z(t-k_i)\|  \right ] \right )
    \end{multline}
    By the inequality of arithmetic and geometric means
    \begin{equation}
        \prod_{i=0}^\m  \|\z(t-k_i)\| \leq \frac{1}{\m}\sum_{i=1}^\m \|\z(t-k_i)\|^\m
    \end{equation}
    then
    \begin{equation}
        \bE\left [ \prod_{i=0}^\m  \|\z(t-k_i)\|  \right ] \leq \bE\left [ \frac{1}{\m}\sum_{i=1}^\m \|\z(t-k_i)\|^\m  \right ] =  \frac{1}{\m}\sum_{i=1}^\m \bE\left [\|\z(t-k_i)\|^\m  \right ]
    \end{equation}
    By assumption $\z(t)$ is stationary, therefore
    $\bE[\|\z(t-k_i)\|^\m]=\bE[\|\z(t)\|^\m]$, i.e. $\bE[\|\z(t)\|^\m]$ does not depend on $k_i$, and so we obtain the statement of the lemma
    \begin{equation}
         \bE[\|\mathbf{s}(t)\|^\m] \leq \bE[\|\z(t)\|^\m] \sum_{k_1=0}^\infty \dots \sum_{k_\m=0}^\infty \left ( \prod_{i=1}^\m \|\alpha_{k_i}\|  \right ) = \left ( \sum_{k=0}^\infty \|\alpha_k\| \right )^\m\bE[\|\z(t)\|^\m]
    \end{equation}
\end{Proof}

\begin{lemma}\label{lemma:E|zinf-zf|^r}
Let $r\in\mathbb{N}$, then with notation as above the following holds
    \begin{align}
        \bE[\|\z_\infty(t)-\z_f(t)\|^r] \leq \hat{\gamma}^{rt}\left (  \frac{\hat{M} \| \hat{C} \| \|\hat{B}\|}{1-\hat{\gamma}} \right )^r \bE \left [ \left \|\begin{bmatrix} \y(t) \\ \mathbf{u}(t) \end{bmatrix} \right \|^r \right ] 
    \end{align}
\end{lemma}
\begin{Proof}[of Lemma \ref{lemma:E|zinf-zf|^r}]
    Notice that the process $\mathbf{s}(t)=\z_\infty(t)-\z_f(t)=\hat{\y}_f(t|0)-\hat{\y}_f(t)$ can be expressed as:
    \begin{align}
        \mathbf{s}(t)&=\left (\sum_{k=1}^t \hat{C}\hat{A}^{k-1}\hat{B}\w(t-k)+\hat{D}\w(t) \right ) - \left ( \sum_{k=1}^\infty \hat{C}\hat{A}^{k-1}\hat{B}\w(t-k)+\hat{D}\w(t) \right)\\
        &=-\sum_{k=t+1}^\infty \hat{C}\hat{A}^{k-1}\hat{B}\w(t-k) 
    \end{align}
    in the case of $\w(t)=\mathbf{u}(t)$
    \begin{align}
        \mathbf{s}(t)=-\sum_{k=t+1}^\infty \hat{C}\hat{A}^{k-1}\hat{B}\mathbf{u}(t-k) = \sum_{k=0}^\infty \alpha_{k,t}(s,1) \begin{bmatrix} \y(t-k) \\ \mathbf{u}(t-k) \end{bmatrix}
    \end{align}
    with
    \begin{align}
        \alpha_{k,t}(s,1)=\begin{cases}\begin{bmatrix} 0 & -\hat{C}\hat{A}^{k-1}\hat{B} \end{bmatrix}, & k\geq t+1 \\ 0,& k<t+1 \end{cases}
    \end{align}
    In the case of $\w(t)=\begin{bmatrix} \y^T(t) & \mathbf{u}^T(t) \end{bmatrix}^T$
    \begin{align}
        \mathbf{s}(t)=-\sum_{k=t+1}^\infty \hat{C}\hat{A}^{k-1}\hat{B}\begin{bmatrix} \y(t-k) \\ \mathbf{u}(t-k) \end{bmatrix} = \sum_{k=0}^\infty \alpha_{k,t}(s,2) \begin{bmatrix} \y(t-k) \\ \mathbf{u}(t-k) \end{bmatrix}
    \end{align}
    with
    \begin{align}
        \alpha_{k,t}(s,2)=\begin{cases} -\hat{C}\hat{A}^{k-1}\hat{B}, & k\geq t+1 \\ 0,& k<t+1 \end{cases}
    \end{align}
    Notice that in both cases we can upper-bound with the same quantity $\|\alpha_{k,t}(s,1)\|\leq \|\alpha_{k,t}(s)\|$, and $\|\alpha_{k,t}(s,2)\|\leq \|\alpha_{k,t}(s)\|$ with
    \begin{align}
        \|\alpha_{k,t}(s)\|=\begin{cases} \|\hat{C}\hat{A}^{k-1}\hat{B}\|,& k\geq t+1 \\ 0, & k<t+1 \end{cases}
    \end{align}
    Since $\w(t)$ is a stationary process, and by assumption predictors are stable, i.e. all eigenvalues of $\hat{A}$ are inside unit circle, thus $\sum_{k=0}^\infty \|\alpha_{k,t}(s)\| \leq +\infty, \forall t\geq0$, we apply Lemma \ref{lemma:stationary4thMoment}, and obtain
    \begin{align}
        \bE[\|\mathbf{s}(t)\|^r]=\bE[\|\z_\infty(t)-\z_f(t)\|^r]&\leq \left ( \sum_{k=0}^\infty \|\alpha_{k,t}(s)\| \right )^r \bE\left [ \left \|\begin{bmatrix} \y(t) \\ \mathbf{u}(t) \end{bmatrix} \right \|^r \right ]\\
        &\leq \left ( \sum_{k=t+1}^\infty \|\hat{C}\| \|\hat{A}^{k-1}\| \| \hat{B}\| \right )^r \bE\left [ \left \|\begin{bmatrix} \y(t) \\ \mathbf{u}(t) \end{bmatrix} \right \|^r \right ]
    \end{align}
    with $\|\hat{A}^k\|\leq \hat{M}\hat{\gamma}^k$, for some $M>1$ and $\hat{\gamma}\in [\hat{\gamma}^*,1)$, where $\hat{\gamma}^*$ is the spectral radius of $\hat{A}$, then with a sum of geometric series, we get the statement of the lemma
    \begin{align}
        \bE[\|\z_\infty(t)-\z_f(t)\|^r] \leq \left ( \hat{M} \| \hat{C} \| \|\hat{B}\| \frac{\hat{\gamma}^t}{1-\hat{\gamma}} \right )^r \bE\left [ \left \|\begin{bmatrix} \y(t) \\ \mathbf{u}(t) \end{bmatrix} \right \|^r \right ].
    \end{align}

\end{Proof}

\begin{lemma}\label{lemma:E|z_inf|^r} Let $r\in\mathbb{N}$, then with notation as above the following holds
    \begin{align}
        \bE \left [ \|\z_\infty(t)\|^r \right ] &\leq \left (1+\|\hat{D}\|+ \frac{\hat{M}\|\hat{B}\|\|\hat{C}\|}{1-\hat{\gamma}} \right )^r \bE\left [ \left \|\begin{bmatrix} \y(t) \\ \mathbf{u}(t) \end{bmatrix}\right \|^r \right ]
    \end{align}
\end{lemma}
\begin{Proof}[of Lemma \ref{lemma:E|z_inf|^r}]
    Notice that $\z_\infty(t)=\y(t)-\hat{\y}_f(t)$ can be expressed as

    In the case of $\w(t)=\mathbf{u}(t)$, 
    \begin{align}
        \z_\infty(t)=\y(t)-\sum_{k=1}^\infty \hat{C}\hat{A}^{k-1}\hat{B}\mathbf{u}(t-k)-\hat{D}\mathbf{u}(t) = \sum_{k=0}^\infty \alpha_k(\z_\infty,1) \begin{bmatrix} \y(t-k) \\ \mathbf{u}(t-k) \end{bmatrix}
    \end{align}
    with 
    \begin{align}
        \alpha_k(\z_\infty,1) = \begin{cases} \begin{bmatrix} I & -\hat{D}\end{bmatrix},& k=0 \\ \begin{bmatrix} 0 & -\hat{C}\hat{A}^{k-1}\hat{B}\end{bmatrix}, & k>0  \end{cases}
    \end{align}
    in the case of $\w(t)=[\y^T(t), \mathbf{u}^T(t)]^T$
    \begin{align}
        \z_\infty(t)=\y(t)-\sum_{k=1}^\infty \hat{C}\hat{A}^{k-1}\hat{B}\begin{bmatrix} \y(t-k)\\ \mathbf{u}(t-k)\end{bmatrix} -\hat{D}\begin{bmatrix} \y(t)\\ \mathbf{u}(t)\end{bmatrix}=\sum_{k=0}^\infty \alpha_k(\z_\infty,2) \begin{bmatrix} \y(t-k) \\ \mathbf{u}(t-k) \end{bmatrix}
    \end{align}
    Recall that in this case, we assume $\hat{D}=[0,\hat{D}_\mathbf{u}]$, note that $\|\hat{D}\|=\|\hat{D}_\mathbf{u}\|$ and thus
    \begin{align}
        \alpha_k(\z_\infty,2) = \begin{cases} \begin{bmatrix} I & -\hat{D}_\mathbf{u}\end{bmatrix},& k=0 \\ -\hat{C}\hat{A}^{k-1}\hat{B}, & k>0  \end{cases}
    \end{align}
    Note that in both cases we can upper-bound with the same quantity, i.e. $\|\alpha_k(\z_\infty)\|\leq \|\alpha_k(\z_\infty)\|$, and $\|\alpha_k(\z_\infty,2)\|\leq \|\alpha_k(\z_\infty)\|$, with
    \begin{align}
        \|\alpha_k(\z_\infty)\|\leq \begin{cases} 1 + \|\hat{D}\|,& k=0\\
        \|\hat{C}\hat{A}^{k-1}\hat{B}\|,& k>0 \end{cases} \label{eq:alpha_k(z_inf)}
    \end{align}
    Since, in both cases, $\sum_{k=0}^\infty \|\alpha_k(\z_\infty)\|\leq +\infty$, due to stability of the predictor, and $\begin{bmatrix} \y^T(t)& \mathbf{u}^T(t) \end{bmatrix}^T$ is stationary, we apply Lemma \ref{lemma:stationary4thMoment}, to both cases, and upper bound by \eqref{eq:alpha_k(z_inf)}, to obtain an upper-bound for both cases:
    \begin{align}
        \bE \left [ \|\z_\infty(t)\|^r \right ] &\leq \left (\sum_{k=0}^\infty \|\alpha_k(\z_\infty,1)\| \right )^4 \bE\left [ \left \|\begin{bmatrix} \y(t) \\ \mathbf{u}(t) \end{bmatrix}\right \|^r \right ]\\
        &\leq\left (\|I\|+\|\hat{D}\|+\sum_{k=1}^\infty \|\hat{C}\hat{A}^{k-1}\hat{B}\| \right )^r \bE\left [ \left \|\begin{bmatrix} \y(t) \\ \mathbf{u}(t) \end{bmatrix}\right \|^r \right ]\\
        &\leq \left (1+\|\hat{D}\|+ \frac{\hat{M}\|\hat{B}\|\|\hat{C}\|}{1-\hat{\gamma}} \right )^r \bE\left [ \left \|\begin{bmatrix} \y(t) \\ \mathbf{u}(t) \end{bmatrix}\right \|^r \right ]
    \end{align}

\end{Proof}

\begin{lemma}\label{lemma:E|z_f|^r} Let $r\in\mathbb{N}$, then with notation as above, the following holds
    \begin{align}
        \bE \left [ \|\z_f(t)\|^r \right ] &\leq \left ( \|I\| + \|\hat{D}\| +  \frac{\hat{M}\|\hat{B}\|\|\hat{C}\|}{1-\hat{\gamma}} \right )^r \bE\left [ \left \|\begin{bmatrix} \y(t) \\ \mathbf{u}(t) \end{bmatrix}\right \|^r \right ]
    \end{align}
\end{lemma}

\begin{Proof}[of Lemma \ref{lemma:E|z_f|^r}]
    Notice that the process $\z_f(t)=\y(t)-\hat{\y}(t|0)$ can be expressed as:\\
    In the case of $\w(t)=\mathbf{u}(t)$
    \begin{align}
        \z_f(t)= \y(t)-\sum_{k=1}^t \hat{C}\hat{A}^{k-1}\hat{B}\mathbf{u}(t-k) - \hat{D}\mathbf{u}(t) = \sum_{k=0}^\infty \alpha_k(\z_f,1) \begin{bmatrix} \y(t-k) \\ \mathbf{u}(t-k) \end{bmatrix}
    \end{align}
    with
    \begin{equation}
        \alpha_k(\z_f,1)=\begin{cases} \begin{bmatrix}I&-\hat{D} \end{bmatrix},& k=0 \\ \begin{bmatrix} 0 & -\hat{C}\hat{A}^{k-1}\hat{B} \end{bmatrix}, & 0<k\leq t \\ 0,& k>t  \end{cases}
    \end{equation}
    In the case of $\w(t)=[\y^T(t), \mathbf{u}^T(t)]^T$, 
    \begin{align}
        \z_f(t)= \y(t)-\sum_{k=1}^t \hat{C}\hat{A}^{k-1}\hat{B}\begin{bmatrix} \y(t-k) \\ \mathbf{u}(t-k) \end{bmatrix} - \hat{D}\begin{bmatrix} \y(t) \\ \mathbf{u}(t) \end{bmatrix} = \sum_{k=0}^\infty \alpha_k(\z_f,2) \begin{bmatrix} \y(t-k) \\ \mathbf{u}(t-k) \end{bmatrix}
    \end{align}
    with
    \begin{equation}
        \alpha_k(\z_f,2)=\begin{cases} \begin{bmatrix}I&0\end{bmatrix} -\hat{D},& k=0 \\ -\hat{C}\hat{A}^{k-1}\hat{B}, & 0<k\leq t \\ 0,& k>t  \end{cases}
    \end{equation}
    Note that for both cases we can upper-bound by the same quantity $\|\alpha_k(\z_f,1)\|\leq \|\alpha_k(\z_f)\|$, and $\|\alpha_k(\z_f,2)\|\leq \|\alpha_k(\z_f)\|$, with
    \begin{align}
        \|\alpha_k(\z_f)\|=\begin{cases} 1+\|\hat{D}\|,& k=0\\ \|\hat{C}\hat{A}^{k-1}\hat{B}\|,& 0<k\leq t \\
        0,& k>t \end{cases}
    \end{align}
    Since by assumption predictors are stable, we apply Lemma \ref{lemma:stationary4thMoment} and obtain
    \begin{align}
        \bE \left [ \|\z_f(t)\|^r \right ] &\leq \left ( \sum_{k=0}^\infty \|\alpha_k(\z_f)\| \right )^r \bE\left [ \left \|\begin{bmatrix} \y(t) \\ \mathbf{u}(t) \end{bmatrix}\right \|^r \right ]\\
        &\leq \left ( \|I\| + \|\hat{D}\| + \sum_{k=1}^t \|\hat{C}\hat{A}^{k-1}\hat{B}\| \right )^r \bE\left [ \left \|\begin{bmatrix} \y(t) \\ \mathbf{u}(t) \end{bmatrix}\right \|^r \right ]\\
        &\leq \left ( \|I\| + \|\hat{D}\| + \hat{M}\|\hat{B}\|\|\hat{C}\| \sum_{k=1}^t \hat{\gamma}^{k-1} \right )^r \bE\left [ \left \|\begin{bmatrix} \y(t) \\ \mathbf{u}(t) \end{bmatrix}\right \|^r \right ]\\
        &= \left ( \|I\| + \|\hat{D}\| + \hat{M}\|\hat{B}\|\|\hat{C}\| \frac{1-\hat{\gamma}^t}{1-\hat{\gamma}} \right )^r \bE\left [ \left \|\begin{bmatrix} \y(t) \\ \mathbf{u}(t) \end{bmatrix}\right \|^r \right ]
    \end{align}
    Notice that $\hat{\gamma}^t>0, \forall t$, thus we obtain the statement of the lemma
    \begin{align}
        \bE \left [ \|\z_f(t)\|^r \right ] &\leq \left ( \|I\| + \|\hat{D}\| +  \frac{\hat{M}\|\hat{B}\|\|\hat{C}\|}{1-\hat{\gamma}} \right )^r \bE\left [ \left \|\begin{bmatrix} \y(t) \\ \mathbf{u}(t) \end{bmatrix}\right \|^r \right ].
    \end{align}
\end{Proof}

\begin{lemma}\label{lemma:E|yw|^r} Let $r\in\mathbb{N}$, then with notation as above, the following holds.
    \begin{align}
        \bE\left [ \left \|\begin{bmatrix} \y(t) \\ \mathbf{u}(t) \end{bmatrix}\right \|^r \right ] \leq \|\Sigma_{gen}\|_{\ell_1}^r G_r(\e_g)
    \end{align}
    with
    \begin{align}
        \|\Sigma_{gen}\|_{\ell_1}&=\|I\|+\sum_{k=1}^\infty \|C_gA_g^{k-1}K_g\|\\
        G_r(\e_g)&=\begin{cases} 2^{\frac{r}{2}}\mu_{\max}(Q_e)^{\frac{r}{2}}(n_u+n_y+\frac{r}{2}-1)!,& r \text{ is even}\\ 2\mu_{\max}(Q_e)^{\frac{r}{2}}\sqrt{(n_u+n_y+r-1)!},& r \text{ is odd} \end{cases} 
    \end{align}
\end{lemma}

\begin{Proof}[of Lemma \ref{lemma:E|yw|^r}]
    Note that $\begin{bmatrix} \y(t) \\ \mathbf{u}(t) \end{bmatrix}$ can be expressed as
    \begin{align}
        \begin{bmatrix} \y(t) \\ \mathbf{u}(t) \end{bmatrix} = \sum_{k=1}^\infty C_gA_g^{k-1}K_g\e_g(t-k) + \e_g(t) = \sum_{k=0}^\infty \alpha_k(\y,\w)\e_g(t-k)
    \end{align}
    with $\e(t)$ stationary, we apply Lemma \ref{lemma:stationary4thMoment} to get
    \begin{align}
        \bE\left [ \left \|\begin{bmatrix} \y(t) \\ \mathbf{u}(t) \end{bmatrix}\right \|^r \right ] \leq \left ( \sum_{k=0}^\infty \|\alpha_k(\y,\w)\| \right )^r \bE\left [ \|\e_g(t)\|^r \right]
    \end{align}
Let us denote $\|\Sigma_{gen}\|_{\ell_1}=\sum_{k=0}^\infty \|\alpha_k(\y,\w)\|$, the $\ell_1$ norm of the generative system. Furthermore we can apply Lemma \ref{lemma:evenEmoments} and Lemma \ref{lemma:Emoments} to obtain, 
	\begin{align*}
		\bE[\|\e_g(t)\|_2^{r}]\leq G_r(\e_g)=\begin{cases} 2^{\frac{r}{2}}\mu_{\max}(Q_e)^{\frac{r}{2}}(n_u+n_y+\frac{r}{2}-1)!,& r \text{ is even}\\ 2\mu_{\max}(Q_e)^{\frac{r}{2}}\sqrt{(n_u+n_y+r-1)!},& r \text{ is odd} \end{cases} 
	\end{align*}
 with this we have the statement of the lemma. 
\end{Proof}

\begin{lemma} \label{lemma:(a+b)^2r} Let $r\in\mathbb{N}$, and $r\geq0$, then for $a,b\in\reals$ the following holds
    \begin{align}
        (a+b)^{2r} \leq 2^{2r-1}a^{2r}+2^{2r-1}b^{2r}
    \end{align}
\end{lemma}
\begin{Proof}[of Lemma \ref{lemma:(a+b)^2r}]
    \begin{align}
        (a+b)^{2r} = 2^{2r}\frac{1}{2^{2r}}(a+b)^{2r}= 2^{2r} \left (\frac{1}{2}(a+b) \right )^{2r}
    \end{align}
    since $\phi(x)=x^{2r}$ is convex for $r\geq 0$, we have by definition of convexity
    \begin{align}
        \left (\frac{1}{2}(a+b) \right )^{2r}=\phi\left (\frac{a+b}{2} \right )\leq \frac{1}{2} \phi(a) + \frac{1}{2}\phi(b)
    \end{align}
    thus we obtain the statement of the lemma
    \begin{align}
        (a+b)^{2r}\leq \frac{2^{2r}}{2} (a^{2r}+b^{2r}) =  2^{2r-1}(a^{2r}+b^{2r})
    \end{align}
\end{Proof}

\begin{lemma} \label{lemma:moments_|V-Lhat|} Let $r\in\mathbb{N} $, then with notation as above, the following holds
    \begin{align}
        \bE[\|V_N(f)-\hat{\mathcal{L}}_N(f)\|^r] \leq \frac{(n_u+n_y+r-1)!}{\sqrt{N}} \left ( 4\bar{G}_{gen}  \bar{G}_f(f)  \right )^r
    \end{align}
    with 
    \begin{align}
        \bar{G}_f(f)&=\left (1+\|\hat{D}\|+ \frac{\hat{M}\|\hat{B}\|\|\hat{C}\|}{1-\hat{\gamma}} \right ) \frac{\hat{M}\|\hat{C}\| \|\hat{B}\|}{(1-\hat{\gamma})^\frac{3}{2}} \\
        \bar{G}_{gen}&=\|\Sigma_{gen}\|_{\ell_1}^{2} \mu_{\max}(Q_e)
    \end{align}
\end{lemma}

\begin{Proof}
        with $\z_\infty(t)=\y(t)-\hat{\y}_f(t)$, and $\z_f(t)=\y(t)-\hat{\y}_f(t|0)$, we start by applying triangle inequalities
    \begin{multline}
        \bE[\|V_N(f)-\hat{\mathcal{L}}_N(f)\|^r] = \bE \left [\left | \frac{1}{N}\sum_{t=0}^{N-1} \|\z_\infty(t)\|^2-\|\z_f(t)\|^2   \right |^r \right ] \leq \bE \left [ \left (  \frac{1}{N}\sum_{t=0}^{N-1} \left | \|\z_\infty(t)\|^2-\|\z_f(t)\|^2   \right | \right )^r \right ]
    \end{multline}
    \begin{align}
        \bE[\|V_N(f)-\hat{\mathcal{L}}_N(f)\|^r] \leq \frac{1}{N^r} \sum_{t_1=0}^{N-1} \dots \sum_{t_r=0}^{N-1} \bE \left [  \prod_{j=1}^r \left | \|\z_\infty(t_j)\|^2-\|\z_f(t_j)\|^2 \right |\right ]
    \end{align}
    Now using the fact that $|a^2-b^2|=|(a-b)(a+b)| = |a-b|(a+b)$, since $a,b\geq 0$, we get
    \begin{align}
        \bE[\|V_N(f)-\hat{\mathcal{L}}_N(f)\|^r] \leq \frac{1}{N^r} \sum_{t_1=0}^{N-1} \dots \sum_{t_r=0}^{N-1} \bE \left [  \prod_{j=1}^r \left | \|\z_\infty(t_j)\|-\|\z_f(t_j)\| \right |\left ( \|\z_\infty(t_j)\|+\|\z_f(t_j)\| \right ) \right ]
    \end{align}
    We apply Cauchy-Schwarz, i.e. $\bE[XY]\leq |\bE[XY]| \leq \sqrt{\bE[X^2]} \sqrt{\bE[Y^2]}$, with $X=\prod_{j=1}^r \left | \|\z_\infty(t_j)\|-\|\z_f(t_j)\| \right |$, and $Y=\prod_{j=1}^r \left ( \|\z_\infty(t_j)\|+\|\z_f(t_j)\| \right )$, 
    \begin{align}
        \bE[\|V_N(f)-\hat{\mathcal{L}}_N(f)\|^r] \leq \frac{1}{N^r} \sum_{t_1=0}^{N-1} \dots \sum_{t_r=0}^{N-1} \sqrt{\bE \left [  \prod_{j=1}^r \left | \|\z_\infty(t_j)\|-\|\z_f(t_j)\| \right |^2 \right ] } \sqrt{\bE \left [ \prod_{j=1}^r \left ( \|\z_\infty(t_j)\|+\|\z_f(t_j)\| \right )^2 \right ]} \label{eq:ASDFH}
    \end{align}
    For now let's focus on $\bE \left [  \prod_{j=1}^r \left | \|\z_\infty(t_j)\|-\|\z_f(t_j)\| \right |^2 \right ]$, by applying reverse triangle inequality we obtain
    \begin{align}
        \bE \left [  \prod_{j=1}^r \left | \|\z_\infty(t)\|-\|\z_f(t)\| \right |^2 \right ] \leq \bE \left [  \prod_{j=1}^r  \|\z_\infty(t)-\z_f(t)\|^2 \right ]
    \end{align}
    now we apply the inequality of arithmetic-geometric means
    \begin{align}
        \bE \left [  \prod_{j=1}^r  \|\z_\infty(t)-\z_f(t)\|^2 \right ] \leq \frac{1}{r}\sum_{j=1}^r \bE[\|\z_\infty(t)-\z_f(t) \|^{2r}]
    \end{align}
    by applying Lemma \ref{lemma:E|zinf-zf|^r}, we obtain the first term
    \begin{align}
        \bE \left [  \prod_{j=1}^r \left | \|\z_\infty(t_j)\|-\|\z_f(t_j)\| \right |^2 \right ]  \leq  \left (\frac{\hat{M}\|\hat{C}\| \|\hat{B}\|}{1-\hat{\gamma}} \right )^{2r} \bE \left [ \left \| \begin{bmatrix} \y(t) \\ \mathbf{u}(t) \end{bmatrix} \right \|^{2r} \right ] \frac{1}{r} \sum_{j=1}^r \hat{\gamma}^{2rt_j} \label{eq:aslkjgdsaf}
    \end{align}
    Now for the second term $\bE \left [ \prod_{j=1}^r \left ( \|\z_\infty(t_j)\|+\|\z_f(t_j)\| \right )^2 \right ]$, we apply the inequality of arithmetic-geometric means
    \begin{align}
        \bE \left [ \prod_{j=1}^r \left ( \|\z_\infty(t_j)\|+\|\z_f(t_j)\| \right )^2 \right ] \leq \frac{1}{r} \sum_{j=1}^r \bE\left [ \left ( \|\z_\infty(t_j)\|+\|\z_f(t_j)\| \right )^{2r}\right ]
    \end{align}
    By Lemma \ref{lemma:(a+b)^2r}, we obtain
    \begin{align}
        \frac{1}{r} \sum_{j=1}^r \bE\left [ \left ( \|\z_\infty(t_j)\|+\|\z_f(t_j)\| \right )^{2r}\right ] \leq \frac{2^{2r-1}}{r} \sum_{j=1}^r  \left ( \bE\left [ \|\z_\infty(t_j)\|^{2r} \right] + \bE\left [ \|\z_f(t_j)\|^{2r} \right ] \right )
    \end{align}
    By Lemma \ref{lemma:E|z_inf|^r} and Lemma \ref{lemma:E|z_f|^r}, we obtain
    \begin{align}
         \frac{2^{2r-1}}{r} \sum_{j=1}^r  \left ( \bE\left [ \|\z_\infty(t_j)\|^{2r} \right] + \bE\left [ \|\z_f(t_j)\|^{2r} \right ] \right ) &\leq \frac{2^{2r}}{r} \sum_{j=1}^r \left (1+\|\hat{D}\|+ \frac{\hat{M}\|\hat{B}\|\|\hat{C}\|}{1-\hat{\gamma}} \right )^{2r} \bE\left [ \left \|\begin{bmatrix} \y(t) \\ \mathbf{u}(t) \end{bmatrix}\right \|^{2r} \right ]\\
         &=2^{2r} \left (1+\|\hat{D}\|+ \frac{\hat{M}\|\hat{B}\|\|\hat{C}\|}{1-\hat{\gamma}} \right )^{2r} \bE\left [ \left \|\begin{bmatrix} \y(t) \\ \mathbf{u}(t) \end{bmatrix}\right \|^{2r} \right ] \label{eq:ASDGDSFHG}
    \end{align}
    Now taking \eqref{eq:ASDGDSFHG} and \eqref{eq:aslkjgdsaf} back to \eqref{eq:ASDFH}, we have

    \begin{multline}
        \bE[\|V_N(f)-\hat{\mathcal{L}}_N(f)\|^r] \leq \frac{1}{N^r} \sum_{t_1=0}^{N-1} \dots \sum_{t_r=0}^{N-1} \sqrt{\bE \left [  \prod_{j=1}^r \left | \|\z_\infty(t_j)\|-\|\z_f(t_j)\| \right |^2 \right ] } \sqrt{\bE \left [ \prod_{j=1}^r \left ( \|\z_\infty(t_j)\|+\|\z_f(t_j)\| \right )^2 \right ]} \\
        \leq \frac{1}{N^r} \sum_{t_1=0}^{N-1} \dots \sum_{t_r=0}^{N-1} \sqrt{\left (\frac{\hat{M}\|\hat{C}\| \|\hat{B}\|}{1-\hat{\gamma}} \right )^{2r} \bE \left [ \left \| \begin{bmatrix} \y(t) \\ \mathbf{u}(t) \end{bmatrix} \right \|^{2r} \right ] \frac{1}{r} \sum_{j=1}^r \hat{\gamma}^{2rt_j}} \\
        \cdot \sqrt{2^{2r} \left (1+\|\hat{D}\|+ \frac{\hat{M}\|\hat{B}\|\|\hat{C}\|}{1-\hat{\gamma}} \right )^{2r} \bE\left [ \left \|\begin{bmatrix} \y(t) \\ \mathbf{u}(t) \end{bmatrix}\right \|^{2r} \right ] }
    \end{multline}
    \begin{multline}
        \bE[\|V_N(f)-\hat{\mathcal{L}}_N(f)\|^r] \leq 2^{r} \left (1+\|\hat{D}\|+ \frac{\hat{M}\|\hat{B}\|\|\hat{C}\|}{1-\hat{\gamma}} \right )^{r} \left (\frac{\hat{M}\|\hat{C}\| \|\hat{B}\|}{1-\hat{\gamma}} \right )^{r} \bE\left [ \left \|\begin{bmatrix} \y(t) \\ \mathbf{u}(t) \end{bmatrix}\right \|^{2r} \right ] \\
        \cdot \frac{1}{N^r} \sum_{t_1=0}^{N-1} \dots \sum_{t_r=0}^{N-1}\sqrt{\frac{1}{r} \sum_{j=1}^r \hat{\gamma}^{2rt_j}}
    \end{multline}
    Note that we can write
    \begin{align}
        \frac{1}{N^r} \sum_{t_1=0}^{N-1} \dots \sum_{t_r=0}^{N-1}\sqrt{\frac{1}{r} \sum_{j=1}^r \hat{\gamma}^{2rt_j}} = \frac{1}{N^r} \sum_{t_1=0}^{N-1} \dots \sum_{t_r=0}^{N-1} \phi(\frac{1}{r} \sum_{j=1}^r \hat{\gamma}^{2rt_j})
    \end{align}
    thus we can apply Jensen's inequality for concave function $\phi(x)=\sqrt{x}$, i.e. $\phi \left (\frac{1}{\|S\|}\sum_{i\in S} x_i\right ) \geq \frac{1}{\|S\|} \sum_{i\in S} \phi (x_i)$, thus we obtain
    \begin{align}
        \frac{1}{N^r} \sum_{t_1=0}^{N-1} \dots \sum_{t_r=0}^{N-1}\sqrt{\frac{1}{r} \sum_{j=1}^r \hat{\gamma}^{2rt_j}} \leq \sqrt{\frac{1}{N^r} \sum_{t_1=0}^{N-1} \dots \sum_{t_r=0}^{N-1}\frac{1}{r} \sum_{j=1}^r \hat{\gamma}^{2rt_j}}
    \end{align}
    Now by commuting the sums we get
    \begin{align}
        \sqrt{\frac{1}{N^r} \sum_{t_1=0}^{N-1} \dots \sum_{t_r=0}^{N-1}\frac{1}{r} \sum_{j=1}^r \hat{\gamma}^{2rt_j}} = \sqrt{\frac{1}{r} \sum_{j=1}^r \frac{1}{N^r} \sum_{t_1=0}^{N-1} \dots \sum_{t_r=0}^{N-1} \hat{\gamma}^{2rt_j}}
    \end{align}
    now notice that $\hat{\gamma}^{2rt_j}$ only depend on one sum, for which we can use the sum of geometric series, after which the same term will be repeated $N^{r-1}$ times, therefore
    \begin{align}
        \sqrt{\frac{1}{r} \sum_{j=1}^r \frac{1}{N^r} \sum_{t_1=0}^{N-1} \dots \sum_{t_r=0}^{N-1} \hat{\gamma}^{2rt_j}} = \sqrt{\frac{1}{r}\sum_{j=1}^r \frac{N^{r-1}}{N^r} \frac{1-\hat{\gamma}^{2rN}}{1-\hat{\gamma}^{2r}} } =  \frac{1}{\sqrt{N}} \sqrt{\frac{1-\hat{\gamma}^{2rN}}{1-\hat{\gamma}^{2r}}}
    \end{align}
    since $\hat{\gamma}^{2rN}\geq0$, and $(1-\hat{\gamma})^{\frac{r}{2}}\leq (1-\hat{\gamma}^{2r})^\frac{1}{2}$, since
    \begin{align}
        (1-\hat{\gamma})^{\frac{r}{2}}&\leq \left ( (1-\hat{\gamma}^{r})(1+\hat{\gamma}^{r}) \right )^\frac{1}{2}\\
        1&\leq (1+\hat{\gamma}^{r})
    \end{align}
    we obtain
    \begin{align}
        \bE[\|V_N(f)-\hat{\mathcal{L}}_N(f)\|^r] \leq \frac{2^{r}}{\sqrt{N}} \left (1+\|\hat{D}\|+ \frac{\hat{M}\|\hat{B}\|\|\hat{C}\|}{1-\hat{\gamma}} \right )^{r} \left (\frac{\hat{M}\|\hat{C}\| \|\hat{B}\|}{(1-\hat{\gamma})^\frac{3}{2}} \right )^{r} \bE\left [ \left \|\begin{bmatrix} \y(t) \\ \mathbf{u}(t) \end{bmatrix}\right \|^{2r} \right ] \label{eq:moments_|V-Lhat|_pre_yw}
    \end{align}
    We can apply Lemma \ref{lemma:E|yw|^r}, to get
    \begin{align}
        \bE\left [ \left \|\begin{bmatrix} \y(t) \\ \mathbf{u}(t) \end{bmatrix}\right \|^{2r} \right ] \leq \|\Sigma_{gen}\|_{\ell_1}^{2r} G_{2r}(\e_g)
    \end{align}
    since $2r$ is always even, then 
    \begin{align}
        G_{2r}(\e_g) = 2^r\mu_{\max}(Q_e)^{r}(n_u+n_y+r-1)!
    \end{align}
    and with this we obtain the statement of the lemma
    \begin{multline}
        \bE[\|V_N(f)-\hat{\mathcal{L}}_N(f)\|^r] \leq \frac{2^{2r}}{\sqrt{N}} \left (1+\|\hat{D}\|+ \frac{\hat{M}\|\hat{B}\|\|\hat{C}\|}{1-\hat{\gamma}} \right )^{r} \left (\frac{\hat{M}\|\hat{C}\| \|\hat{B}\|}{(1-\hat{\gamma})^\frac{3}{2}} \right )^{r} \\
        \cdot \|\Sigma_{gen}\|_{\ell_1}^{2r} \mu_{\max}(Q_e)^{r}(n_u+n_y+r-1)!  
    \end{multline}
    with some algebraic manipulation we get
    \begin{multline}
        \bE[\|V_N(f)-\hat{\mathcal{L}}_N(f)\|^r] \leq \frac{(n_u+n_y+r-1)!}{\sqrt{N}} \left ( 4\left (1+\|\hat{D}\|+ \frac{\hat{M}\|\hat{B}\|\|\hat{C}\|}{1-\hat{\gamma}} \right ) \frac{\hat{M}\|\hat{C}\| \|\hat{B}\|}{(1-\hat{\gamma})^\frac{3}{2}}   \|\Sigma_{gen}\|_{\ell_1}^{2} \mu_{\max}(Q_e) \right )^r
    \end{multline}
\end{Proof}

\begin{lemma} \label{lemma:mgf(Vn-hatL)} With notation as above for $0<\lambda< \frac{1}{4 n_w \bar{G}_{gen}  \bar{G}_f(f)}$ following holds
    \begin{align}
        \bE[e^{\lambda|V_N(f)-\hat{\mathcal{L}}_N(f)|}]\leq 1+ \frac{(n_y+n_u)!}{\sqrt{N}} \frac{4 \lambda \bar{G}_{gen}  \bar{G}_f(f)}{1-4 \lambda (n_y+n_u)\bar{G}_{gen}  \bar{G}_f(f)}
    \end{align}
\end{lemma}
\begin{Proof}[of Lemma \ref{lemma:mgf(Vn-hatL)}]
    with $X=\lambda | V_N(f) - \hat{\mathcal{L}}_N(f)|$
    \begin{align}
        \bE[e^{\lambda(V_N(f)-\hat{\mathcal{L}}_N(f))}] = 1+\sum_{r=1}^\infty \frac{\lambda^r}{r!} \bE[|V_N(f)-\hat{\mathcal{L}}_N(f)|^r] \leq 1+ \sum_{r=1}^\infty \frac{\lambda^r}{r!} \frac{(n_u+n_y+r-1)!}{\sqrt{N}} \left ( 4\bar{G}_{gen}  \bar{G}_f(f)  \right )^r
    \end{align}
    Furthermore, with $n_w=n_u+n_y$
    \begin{align*}
        \frac{(n_w+r-1)!}{r!}=n_w!\frac{n_w+1}{2}\frac{n_w+2}{3}\dots \frac{n_w+r-1}{r}
    \end{align*}
    and as $\frac{n_w+r-1}{r}\leq n_w$, for all $r\geq 1$, then
    \begin{align*}
        \frac{(n_w+r-1)!}{r!}\leq n_w!\left (n_w \right )^{r-1} = n_w!\frac{\left (n_w \right )^{r}}{n_w}		= \frac{n_w!}{n_w}\left (n_w \right )^r=(n_w-1)!(n_w)^r.
    \end{align*}
    this allows us to write
    \begin{align}
         \bE[e^{\lambda(V_N(f)-\hat{\mathcal{L}}_N(f))}] \leq 1+ \frac{(n_w-1)!}{\sqrt{N}} \sum_{r=1}^\infty \left (4 \lambda n_w \bar{G}_{gen}  \bar{G}_f(f)  \right )^r
    \end{align}
    the infinite sum is absolutely convergent if $$4 \lambda n_w \bar{G}_{gen}  \bar{G}_f(f)<1$$
    that means that
    \begin{align}
        0<\lambda< \frac{1}{4 n_w \bar{G}_{gen}  \bar{G}_f(f)}
    \end{align}
    under this condition we can write
    \begin{align}
         \bE[e^{\lambda(V_N(f)-\hat{\mathcal{L}}_N(f))}] \leq 1+ \frac{(n_w-1)!}{\sqrt{N}} \frac{4 \lambda n_w \bar{G}_{gen}  \bar{G}_f(f)}{1-4 \lambda n_w\bar{G}_{gen}  \bar{G}_f(f)}=1+ \frac{n_w!}{\sqrt{N}} \frac{4 \lambda \bar{G}_{gen}  \bar{G}_f(f)}{1-4 \lambda n_w\bar{G}_{gen}  \bar{G}_f(f)}
    \end{align}
\end{Proof}

\begin{lemma}\label{lemma:L-Vr} Let $\y_\nu(t),\hat{\y}_{f,\nu}(t),\hat{\y}_{f,\nu}(t|s)\in \mathbb{R}^1$ denote the $\nu$'th component of $\y(t),\hat{\y}_{f}(t),\hat{\y}_{f}(t|s)$ respectively,
\begin{align}
   \mathcal{L}_\nu(f) &\triangleq \bE[(\hat{\y}_{f,\nu}(t)-\y_\nu(t))^2]=\lim_{s \rightarrow -\infty} \bE[(\hat{\y}_{f,\nu}(t|s)-\y_\nu(t))^2]\\
   V_{N,\nu}(f) &\triangleq \frac{1}{N}\sum_{t=0}^{N-1} (\hat{\y}_{f,\nu}(t)-\y_\nu(t))^2
\end{align}
and let $\sigma(r)$, be such that the following holds.
	\begin{align}
		\sigma(r)&\geq \sup_{t,k,l}\bE[\|\e(t,k,l)\|_2^r]\\
		\e(t,k,j)&=\begin{cases} Q_e-\e_g(t-k)\e_g^T(t-j),& k=j\\ -\e_g(t-k)\e_g^T(t-j),& k\neq j \end{cases}
	\end{align}
	Then the raw moments are bounded 
	\begin{align} 
		\bE[(\mathcal{L}_\nu(f)-&V_{N,\nu}(f))^r]\leq\frac{1}{N}\sigma(r)4(r-1)G_{e}(f)^{2r}
	\end{align}
\end{lemma}
\begin{Proof}[Proof of Lemma \ref{lemma:L-Vr}]
	
	The prediction error can be expressed as
	\begin{align*}
		(\y_\nu(t)-\hat{\y}_{f,\nu}(t))=\sum_{k=0}^\infty\alpha_k\e_g(t-k)
	\end{align*}
        
	with
	\begin{align*}
		\alpha_k=\alpha_k(\nu)=\begin{cases} D_{e_\nu},& k=0\\ C_{e_\nu}A_e^{k-1}K_e,& k>0 \end{cases}
	\end{align*}
        where $D_{e_\nu}=\mathbf{1}_\nu D_e$, and $C_{e_\nu}=\mathbf{1}_\nu C_e$ denote the $\nu$'th row of matrices $D_e,C_e$ respectively.
	Then generalised loss $\mathcal{L}_\nu(f)$ for component $\nu$ is expressed as
	\begin{align*}
		\mathcal{L}_\nu(f)&=\bE[(\y_\nu(t)-\hat{\y}_{f,\nu}(t))^2]\\
		&=\bE\left[\text{trace}\left ( \left(\sum_{k=0}^\infty\alpha_k\e_g(t-k)\right)\left(\sum_{k=0}^\infty\alpha_k\e_g(t-k)\right)^T \right )\right]\\
		&=\sum_{k=0}^\infty\alpha_kQ_e\alpha_k^T
	\end{align*}
	and infinite horizon prediction loss is 
	\begin{align*}
		V_{N,\nu}(f)&=\frac{1}{N}\sum_{t=0}^{N-1}(\y_\nu(t)-\hat{\y}_{f,\nu}(t))^2\\
		\mathcal{L}_\nu(f)-V_{N,\nu}(f)&=\frac{1}{N}\sum_{t=0}^{N-1}\left (\sum_{k=0}^\infty\alpha_kQ_e\alpha_k^T  - \sum_{k=0}^\infty\sum_{j=0}^{\infty}\alpha_k\e_g(t-k)\e_g(t-j)\alpha_k^T \right )\\
		&=\frac{1}{N}\sum_{t=0}^{N-1}\sum_{k=0}^\infty\sum_{j=0}^{\infty}\alpha_k\e(t,k,j)\alpha_j^T \\
		\e(t,k,j)&=\begin{cases} \text{trace}(Q_e)-\e_g(t-k)\e_g^T(t-j),& k=j\\ -\e_g(t-k)\e_g^T(t-j),& k\neq j \end{cases}
	\end{align*}
	For ease of notation let us define
	\begin{align*}
		\z(t,k,j)=\alpha_k\e(t,k,j)\alpha_j^T
	\end{align*}
	then
	\begin{align*}
		&\bE[(\mathcal{L}_\nu(f)-V_{N,\nu}(f))^r]\\
		%&=\frac{1}{N^r} \bE\left[\lef\sum_{t_1=0}^{N-1}\dots \sum_{t_r=0}^{N-1} %\sum_{k_1,j_1=0}^\infty \dots \sum_{k_r,j_r=0}^\infty %\prod_{l=1}^{r}z(t_l,k_l,j_l)\right| \le \\
		& = \frac{1}{N^r} \sum_{t_1=0}^{N-1}\dots \sum_{t_r=0}^{N-1} \sum_{k_1,j_1=0}^\infty \dots \sum_{k_r,j_r=0}^\infty \bE\left [\prod_{l=1}^{r}z(t_l,k_l,j_l)\right]
	\end{align*}
	Note that, with i.i.d. innovation noise $\e_g(t)$, if
	\begin{align*}
		&t_r-k_r\notin \{ t_i-k_i,t_i-j_i\}_{i=1}^{r-1}\\
		&\quad \land t_r-j_r\notin \{ t_i-k_i,t_i-j_i\}_{i=1}^{r-1}
	\end{align*}
	or similarly
	\begin{equation}
	\label{pf:b8:eq1}
		\{t_r-k_r,t_r-j_r\} \cap \{ t_i-k_i,t_i-j_i\}_{i=1}^{r-1} = \emptyset
	\end{equation}
	then $\z(t_r,k_r,j_r)$ is independent of $\z(t_i,k_i,j_i)$. Moreover,
	notice that $E(\z(t_r,k_r,j_r)]=0$. 
	Hence, if \eqref{pf:b8:eq1}, it holds that
	\begin{equation}
	\label{pf:b8:eq2}
		%&\bE[(\mathcal{L}(f)-V_N(f))^r]=\frac{1}{N^r}\sum_{t_1=0}^{N-1}\dots %\sum_{t_r=0}^{N-1}\\
		%&\sum_{k_1,j_1=0}^\infty \dots \sum_{k_r,j_r=0}^\infty 
		\bE\left [\prod_{l=1}^{r}z(t_l,k_l,j_l)\right] = \bE\left [\prod_{l=1}^{r-1}\z(t_l,k_l,j_l)\right ]\underset{=0}{\underbrace{\bE[\z(t_r,k_r,j_r)]}}=0.
	\end{equation}
	Let us denote
	\begin{align*}
		\mathcal{Z}=\{ t_i-k_i+k_r,t_i-j_i+k_r,t_i-k_i+j_r,t_i-j_i+j_r\}_{i=1}^{r-1}.
	\end{align*}
	Then using \eqref{pf:b8:eq2} for those $\{t_l,k_l,j_l\}_{l=1}^{r}$ which satisfy
	\eqref{pf:b8:eq1}, it follows that
	\begin{equation}
	\label{pf:b8:eq22}
		\bE[(\mathcal{L}_\nu(f)-V_{N,\nu}(f))^r]=\frac{1}{N^r}\sum_{t_1=0}^{N-1}\dots \sum_{t_{r-1}=0}^{N-1}\sum_{k_1,j_1=0}^\infty \dots \sum_{k_r,j_r=0}^\infty \sum_{t_{r}\in\mathcal{Z}}\bE\left [\prod_{l=1}^{r}z(t_l,k_l,j_l)\right ].
	\end{equation}
	Note that 
	\begin{align*}
		\bE\left [\prod_{l=1}^{r}z(t_l,k_l,j_l)\right ]&\leq \left |\bE\left [\prod_{l=1}^{r}z(t_l,k_l,j_l)\right ] \right |\leq \bE\left [\prod_{l=1}^{r}|z(t_l,k_l,j_l)|\right ].
	\end{align*}
	Let us focus on $|\z(t_i,k_i,j_i)|$:
	\begin{align*}
		|\z(t_l,k_l,j_l)| &\leq \|\alpha_{k_l}\|_2\|\alpha_{j_l}\|_2\|\e(t_l,k_l,j_l)\|_2\\
		\bE\left [\prod_{l=1}^{r}|\z(t_l,k_l,j_l)|\right ]&\leq \prod_{l=1}^r\|\alpha_{k_l}\|_2\|\alpha_{j_l}\|_2\bE\left [\prod_{l=1}^r\|\e(t_l,k_l,j_l)\|_2 \right ]
	\end{align*}
	Then using Arithmetic Mean-Geometric Mean Inequality, \cite{steele2004cauchy} we have
	\begin{align}
		\bE\left [\prod_{l=1}^r\|\e(t_l,k_l,j_l)\| \right ] \leq \frac{1}{r}\sum_{l=1}^r \bE[\|\e(t_l,k_l,j_l)\|_2^r] \label{pf:b8:eq3}
	\end{align}
	% 	\textcolor{blue}{For now, let us, assume $\bE[\|e(t_l,k_l,j_l)\|_2^r]<\sigma(r)$, i.e. }
	Now, let $\sigma(r)$, be such that the following holds.
	\begin{align}
		\sigma(r)\geq \sup_{t,k,l}\bE[\|\e(t,k,l)\|_2^r]
	\end{align}
	Then, 
	%\begin{align*}
	\(	\frac{1}{r}\sum_{l=1}^r \bE[\|\e(t_l,k_l,j_l)\|_2^r] \leq \sigma(r) \)
	%\end{align*}
	 and then from \eqref{pf:b8:eq3} it follows that 
	\begin{align}
		\bE\left [\prod_{l=1}^r|\e(t_l,k_l,j_l)| \right ] \leq \sigma(r)
	\end{align}
	Combining this with \eqref{pf:b8:eq22}, it follows that
	\begin{align}
		&\bE[(\mathcal{L}_\nu(f)-V_{N,\nu}(f))^r]\leq\frac{1}{N^r}\sum_{t_1=0}^{N-1}\dots \sum_{t_{r-1}=0}^{N-1}\sum_{k_1,j_1=0}^\infty \dots \sum_{k_r,j_r=0}^\infty \sum_{t_{r}\in\mathcal{Z}}\sigma(r)\prod_{l=1}^r\|\alpha_{k_l}\|_2\|\alpha_{j_l}\|_2
		\label{pf:b8:eq4}
	\end{align}
	% \begin{align*}
	%     \sum_{t_{r}\in\mathcal{Z}}\sigma(r)\prod_{l=1}^r\|\alpha_{k_l}\|_2\|\alpha_{j_l}\|_2
	% \end{align*}
	and the quantity $\sigma(r)\prod_{l=1}^r\|\alpha_{k_l}\|_2\|\alpha_{j_l}\|_2$ does not depend on $t_r$. Moreover 
	\begin{align*}
		\sum_{t_{r}\in\mathcal{Z}}\sigma(r)\prod_{l=1}^r\|\alpha_{k_l}\|_2\|\alpha_{j_l}\|_2\leq
		\sigma(r)\prod_{l=1}^r\|\alpha_{k_l}\|_2\|\alpha_{j_l}\|_2 |\mathcal{Z}|,
	\end{align*}
	where $|\mathcal{Z}|$ is the cardinality of the set $\mathcal{Z}$. Note $|\mathcal{Z}|\leq 4(r-1)$, therefore
	\begin{align*}
		\sum_{t_{r}\in\mathcal{Z}}&\sigma(r)\prod_{l=1}^r\|\alpha_{k_l}\|_2\|\alpha_{j_l}\|_2\leq \sigma(r)\prod_{l=1}^r\|\alpha_{k_l}\|_2\|\alpha_{j_l}\|_2 4(r-1),
	\end{align*}
	Combining the latter inequality with \eqref{pf:b8:eq4}, it follows that
	\begin{align}
		\bE[(\mathcal{L}_\nu(f)-V_{N,\nu}(f))^r]&\leq \frac{1}{N^r}\sum_{t_1=0}^{N-1}\dots \sum_{t_{r-1}=0}^{N-1}\sigma(r)4(r-1)\sum_{k_1,j_1=0}^\infty \dots \sum_{k_r,j_r=0}^\infty  \prod_{l=1}^r\|\alpha_{k_l}\|_2\|\alpha_{j_l}\|_2 
	\end{align}
	Now notice
	\begin{align*}
		G_{e,\nu}(f)^{2r}=\left ( \sum_{k=0}^\infty \|\alpha_k\|_2 \right )^{2r}=\left ( \sum_{k,j=0}^\infty \|\alpha_k\|_2\|\alpha_j\|_2 \right )^r \\
		=\sum_{k_1,j_1=0}^\infty \dots \sum_{k_r,j_r=0}^\infty  \prod_{l=1}^r\|\alpha_{k_l}\|_2\|\alpha_{j_l}\|_2 
	\end{align*}
    therefore we obtain
	\begin{align*}
		\bE[(\mathcal{L}_\nu(f)-V_{N,\nu}(f))^r] &\leq \frac{1}{N^r}\sum_{t_1=0}^{N-1}\dots \sum_{t_{r-1}=0}^{N-1}\sigma(r)4(r-1)G_{e,\nu}(f)^{2r} \\
		&\leq\frac{1}{N^r}N^{r-1}\sigma(r)4(r-1)G_{e,\nu}(f)^{2r}\\
		&\leq\frac{1}{N}\sigma(r)4(r-1)G_{e,\nu}(f)^{2r}\\
	\end{align*}
    and since
    \begin{align*}
		\|\alpha_k(\nu)\|=\begin{cases} \|\mathbf{1}_\nu D_e\|\leq \|D_e\|,& k=0\\ \|\mathbf{1}_\nu C_eA_e^{k-1}K_e\|\leq \|C_eA_e^{k-1}K_e\|,& k>0 \end{cases}
	\end{align*}
    then
    \begin{align}
        G_{e,\nu}\leq G_e = \|D_e\| + \sum_{k=1}^\infty \|C_eA_e^{k-1}K_e\|
    \end{align}
    and since $2r>1$ we obtain the statement of the lemma
    \begin{align}
        \bE[(\mathcal{L}_\nu(f)-V_{N,\nu}(f))^r] &\leq \frac{1}{N}\sigma(r)4(r-1)G_e(f)^{2r}
    \end{align}
    
\end{Proof}

\begin{lemma}\label{lemma:L-Vrfull} with notation as above the following holds
    \begin{align}
        \bE[(\mathcal{L}(f)-V_{N}(f))^r] &\leq \frac{n_y^r}{N}\sigma(r)4(r-1)G_e(f)^{2r}
    \end{align}
\end{lemma}

\begin{Proof}[of Lemma \ref{lemma:L-Vrfull}]
    By definition
    \begin{align}
        \mathcal{L}(f)&=\bE[ (\y(t)-\hat{\y}_f(t))^T(\y(t)-\hat{\y}_f(t)) ] = \sum_{\nu=1}^{n_y} \bE[(\y_\nu(t)-\hat{\y}_{f,\nu}(t))^2]=\sum_{\nu=1}^{n_y}\mathcal{L}_\nu(f)\\
        V_N(f)&=\frac{1}{N} \sum_{t=0}^{N-1} (\y(t)-\hat{\y}_f(t))^T(\y(t)-\hat{\y}_f(t))  = \sum_{\nu=1}^{n_y} \frac{1}{N} \sum_{t=0}^{N-1} (\y_\nu(t)-\hat{\y}_{f,\nu}(t))^2=\sum_{\nu=1}^{n_y}V_{N,\nu}(f)\\
    \end{align}
    then
    \begin{align}
        \bE[(\mathcal{L}(f)-V_N(f))^r]=\bE\left [ \left (\sum_{\nu=1}^{n_y}\mathcal{L}_\nu(f)-V_{N,\nu}(f) \right )^r \right ]=\sum_{\nu_1}^{n_y}\dots \sum_{\nu_r}^{n_y} \bE\left [ \prod_{i=1}^r (\mathcal{L}_{\nu_i}(f)-V_{N,\nu_i}(f)) \right ]
    \end{align}
    Then using Arithmetic Mean-Geometric Mean Inequality, \cite{steele2004cauchy}, we get $\prod_{i=1}^r (\mathcal{L}_{\nu_i}(f)-V_{N,\nu_i}(f))\leq \frac{1}{r}\sum_{i=1}^r (\mathcal{L}_{\nu_i}(f)-V_{N,\nu_i}(f))^r$, and thus
    \begin{align}
        \bE[(\mathcal{L}(f)-V_N(f))^r]\leq \sum_{\nu_1=1}^{n_y}\dots \sum_{\nu_r=1}^{n_y} \frac{1}{r}\sum_{i=1}^r \bE\left [(\mathcal{L}_{\nu_i}(f)-V_{N,\nu_i}(f))^r \right ]
    \end{align}
    From Lemma \ref{lemma:L-Vr}, we have $\bE[(\mathcal{L}_\nu(f)-V_{N,\nu}(f))^r] \leq \frac{1}{N}\sigma(r)4(r-1)G_e(f)^{2r}$, thus
    \begin{align}
        \bE[(\mathcal{L}(f)-V_N(f))^r] &\leq \sum_{\nu_1=1}^{n_y}\dots \sum_{\nu_r=1}^{n_y} \frac{1}{r}\sum_{i=1}^r \frac{1}{N}\sigma(r)4(r-1)G_e(f)^{2r}\\
        &= \frac{n_y^{r}}{N}\sigma(r)4(r-1)G_e(f)^{2r}
    \end{align}
\end{Proof}

\begin{lemma}\label{lemma:sigmar} let $m=n_u+n_y$, then for $r \ge 2$, the quantity 
	\begin{align*}
		\sigma(r)=\max \left \{(\mu_{\max}(Q_e)^r4(m+r-1)!), (\mu_{\max}(Q_e)^{r}3^r(m+r-1)!)  \right \} = \mu_{\max}(Q_e)^{r}3^r(m+r-1)!
	\end{align*}
	%or simply 
	%\begin{align*}
%		\sigma(r)=\mu_{\max}(Q_e)^{r}3^r(m+r-1)!,\quad r\geq 2,
%	\end{align*} 
satisfies
	\begin{align*}
		\sigma(r)\geq \sup_{t,k,l}\bE[\|\e(t,k,l)\|_2^r]
	\end{align*}
\end{lemma}

\begin{Proof}[Proof of Lemma \ref{lemma:sigmar}]
	Recall that
	\begin{align*}
		\e(t,k,j)=\begin{cases} Q_e-\e_g(t-k)\e_g^T(t-j),& k=j\\ -\e_g(t-k)\e_g^T(t-j),& k\neq j \end{cases}
	\end{align*}
	First let us take the case when $k\neq j$. Then
	\begin{align*}
		\bE[\|\e(t,k,l)\|_2^r]=\bE[\|-\e_g(t-k)\e_g^T(t-j)\|_2^r]
	\end{align*}
	Again as $\e_g(t)$ is i.i.d. we have 
	\begin{align*}
		\bE[\|\e(t,k,l)\|_2^r]\leq \bE[\|\e_g(t-k)\|_2^r]\bE[\|\e_g(t-j)\|_2^r]
	\end{align*}
	and due to stationarity of $\e_g(t)$, we have $\bE[\|\e_g(t-k)\|_2^r]=\bE[\|\e_g(t-j)\|_2^r]$, therefore
	\begin{align*}
		\bE[\|\e(t,k,l)\|_2^r]\leq \bE[\|\e_g(t)\|_2^r]^2
	\end{align*}
	and again due to stationarity of $\e_g(t)$, the moments do not depend on $t$, and using Lemma \ref{lemma:Emoments} we obtain
	\begin{align*}
		\sigma(r)\geq \mu_{\max}(Q_e)^r4((m+r-1)!) \geq \bE[\|\e(t,k,l)\|_2^r]^2
	\end{align*}
	% 	From Lemma \ref{lemma:etoz} 
	% 	\begin{align*}
	% 		\bE[\|e(t)\|_2^r]^2\leq (\mu_{\max}(Q_e)^{\frac{r}{2}}\bE[\|z\|_2^r])^2 
	% 	\end{align*}
	% 	and from Lemma \ref{lemma:zMomentsSq} we obtain
	% 	\begin{align*}
	% 		\bE[\|\e_g(t)\|_2^r]^2\leq \mu_{\max}(Q_e)^r4((m+r-1)!)
	% 	\end{align*}
	Now let us take the case when $k=j$. Then
	\begin{align*}
		\bE[\|\e(t,k,l)\|_2^r]&=\bE[\|Q_e-\e_g(t-k)\e_g^T(t-k)\|_2^r]\\
		&\leq \bE[(\|Q_e\|_2+\|\e_g(t)\|_2^2)^r]\\
		&= \bE \left [ \sum_{j=0}^r \begin{pmatrix} r\\ j \end{pmatrix} \|Q_e\|_2^{r-j}\|\e_g(t)\|_2^{2j} \right ]\\
		&=\sum_{j=0}^r \begin{pmatrix} r\\ j \end{pmatrix} \|Q_e\|_2^{r-j}\bE\|\e_g(t)\|_2^{2j}]
	\end{align*}
	As $Q_e$ is a positive definite matrix,$\|Q_e\|_2 = \mu_{max}(Q_e)$, and hence
	\begin{align*}
		\bE[\|\e(t,k,l)\|_2^r]\leq \sum_{j=0}^r \begin{pmatrix} r\\ j \end{pmatrix} \mu_{\max}(Q_e)^{r-j}\bE\|\e_g(t)\|_2^{2j}]
	\end{align*}
	using Lemma \ref{lemma:evenEmoments} we obtain
	\begin{align*}
		\bE[\|\e(t,k,l)\|_2^r]&\leq \sum_{j=0}^r \begin{pmatrix} r\\ j \end{pmatrix} \mu_{\max}(Q_e)^{r-j}\mu_{\max}(Q_e)^{j}2^j(m+j-1)!\\
		&\leq \mu_{\max}(Q_e)^{r} \sum_{j=0}^r \begin{pmatrix} r\\ j \end{pmatrix}2^j(m+j-1)!.
	\end{align*}
	Since for $j\leq r$,  $(m+j-1)!\leq (m+r-1)!$, hence
	\begin{align*}
		&\bE\|\e(t,k,l)\|_2^{2r}]\leq \mu_{\max}(Q_e)^{r}(m+r-1)! \sum_{j=0}^r \begin{pmatrix} r\\ j \end{pmatrix}2^j
	\end{align*}
	Notice $3^r=(1+2)^r=\sum_{j=0}^r \begin{pmatrix} r\\ j \end{pmatrix}2^j$, hence
	\begin{align*}
		\bE\|\e_g(t,k,l)\|_2^{2r}]\leq \mu_{\max}(Q_e)^{r}3^r(m+r-1)!
	\end{align*}
	Hence,
	\begin{align*}
		\sigma(r)=\max \left \{\mu_{\max}(Q_e)^r4(m+r-1)!, \right . \\
		\left .\mu_{\max}(Q_e)^{r}3^r(m+r-1)!  \right \}.
	\end{align*}
	As we are interested in moments higher or equal to two, i.e. $r\geq 2$, then
	\begin{align*}
		\sigma(r)=\mu_{\max}(Q_e)^{r}3^r(m+r-1)!.
	\end{align*}
\end{Proof}
\begin{lemma}\label{lem:mgf} For $\lambda\leq \left ( 3(m+1) n_y \mu_{\max}(Q_e)G_e(f)^{2} \right )^{-1}$, the moment generating function is bounded
        \begin{align}
            		\bE\left [e^{\lambda(\mathcal{L}(f)-V_N(f))} \right ]\leq 1+\frac{2}{N}\frac{(m+1)! \left (3\lambda n_y\mu_{\max}(Q_e)G_e(f)^{2}\right )^2}{(1-3(m+1)\lambda n_y\mu_{\max}(Q_e)G_e(f)^{2})}
        \end{align}
\end{lemma}
\begin{Proof}[Proof of Lemma \ref{lem:mgf}] \label{proof:mgf} %\ref{thm:mgf}
	We can bound the moment generating function via series expansion. First note that $\bE[\mathcal{L}(f)-V_N(f)]=0$, and hence 
	\begin{align*}
		\bE\left [e^{\lambda(\mathcal{L}(f)-V_N(f))} \right ]=1+\lambda\bE[\mathcal{L}(f)-V_N(f)]+\sum_{r=2}^\infty \frac{\lambda^r}{r!}E[(\mathcal{L}(f)-V_N(f))^r].
	\end{align*}
	 Then using Lemma \ref{lemma:L-Vrfull} we get
	\begin{align}
		\bE\left [e^{\lambda(\mathcal{L}(f)-V_N(f))} \right ]\leq 1+\sum_{r=2}^\infty \frac{\lambda^r}{r!}\frac{n_y^r}{N}\sigma(r)4(r-1)G_e(f)^{2r} %\label{eqA:proofofMGF}
	\end{align}
	% 	With \eqref{eqA:proofofMGF} we reach the statement of the theorem, however we still need to show 
	% 	\begin{align*}
	% 		\sigma(r)\geq \sup_{t,k,l}\bE[\|\e(t,k,l)\|_2^r]
	% 	\end{align*}
	Now using Lemma \ref{lemma:sigmar} we obtain
	\begin{align*}
		&\bE\left [e^{\lambda(\mathcal{L}(f)-V_N(f))} \right ]\leq 1+\frac{1}{N}\sum_{r=2}^\infty \frac{(m+r-1)!}{r!}4(r-1)\left (3n_y\lambda\mu_{\max}(Q_e)G_e(f)^{2}\right )^r\\
	\end{align*}
	Notice that $4(r-1)\leq 2^r$, for $r\in\mathbb{N}$. Furthermore
	\begin{align*}
		\frac{(m+r-1)!}{r!}=m!\frac{m+1}{2}\frac{m+2}{3}\dots \frac{m+r-1}{r}
	\end{align*}
	and as $\frac{m+r-1}{r}\leq \frac{m+1}{2}$, for all $r\geq 2$, then
	\begin{align*}
		\frac{(m+r-1)!}{r!}\leq m!\left (\frac{m+1}{2} \right )^{r-1} = m!\frac{\left (\frac{m+1}{2} \right )^{r}}{\frac{m+1}{2}}		= 2\frac{m!}{m+1}\left (\frac{m+1}{2} \right )^r.
	\end{align*}
	%now we can obtain a bound for which we can define the absolute convergence criteria
	 Hence, we can derive the following inequality:
	\begin{align*}
		&\bE\left [e^{\lambda(\mathcal{L}(f)-V_N(f))} \right ]\leq 1+\frac{2}{N}\frac{m!}{m+1} \sum_{r=2}^\infty \left (3(m+1)\lambda n_y\mu_{\max}(Q_e)G_e(f)^{2}\right )^r.
	\end{align*}
	Notice that if $$| 3(m+1)\lambda n_y\mu_{\max}(Q_e)G_e(f)^{2} | < 1,$$ then the 
	infinite sum 
	$\sum_{r=2}^\infty \left (3(m+1)\lambda n_y\mu_{\max}(Q_e)G_e(f)^{2}\right )^r$
	is absolutely convergent, and 
	\[
	  \begin{split}
	   & \sum_{r=2}^\infty \left (3(m+1)\lambda n_y\mu_{\max}(Q_e)G_e(f)^{2}\right )^r 
	   = \frac{ \left (3(m+1)\lambda n_y\mu_{\max}(Q_e)G_e(f)^{2}\right )^2}{1-3(m+1)\lambda n_y\mu_{\max}(Q_e)G_e(f)^{2}}
	  \end{split}  
	\]
	 To sum up, if
	 %which in turn gives conditions on $\lambda$.
	\begin{align*}
		\lambda\leq \left ( 3(m+1)n_y \mu_{\max}(Q_e)G_e(f)^{2} \right )^{-1}.
	\end{align*}
	then 
	\begin{align*}
		\bE\left [e^{\lambda(\mathcal{L}(f)-V_N(f))} \right ]&\leq 1+\frac{2}{N}\frac{m!}{m+1} \frac{ \left (3(m+1)\lambda n_y\mu_{\max}(Q_e)G_e(f)^{2}\right )^2}{1-3(m+1)\lambda n_y\mu_{\max}(Q_e)G_e(f)^{2}}\\
		&\leq 1+\frac{2}{N}\frac{(m+1)! \left (3\lambda n_y\mu_{\max}(Q_e)G_e(f)^{2}\right )^2}{(1-3(m+1)\lambda n_y\mu_{\max}(Q_e)G_e(f)^{2})}.
	\end{align*}
	
\end{Proof}

\begin{lemma}\label{lemma:PAC-BayesianKL_General} For measurable functions $X(f),Y(f)$ on $\mathcal{F}$, 
    With probability at least $1-\delta$, the following holds
    \begin{align}
         \forall\rho:\quad &E_{f\sim \hat{\rho}} X (f) \le \  E_{f\sim \hat{\rho}} Y(f) +\dfrac{1}{\lambda}\!\left[KL(\hat{\rho} \|\pi) + \ln\dfrac{1}{\delta}	+ \Psi_{\pi}(\lambda,N) \right ],
    \end{align}
      with
    \begin{equation}
        \Psi_{\pi}(\lambda,N)=	\ln E_{f\sim\pi} \bE[e^{\lambda(X(f)-Y(f))}]
    \end{equation}
\end{lemma}

\begin{Proof}[ of Lemma \ref{lemma:PAC-BayesianKL_General}] 
	Let us apply the Donsker \& Varadhan variational 
	formula to the function $\lambda(X(f)-Y(f))$
   it then follows that
   \begin{align}
		 \sup_{\hat{\rho}} (\lambda E_{f\sim \hat{\rho}} X(f) - \lambda E_{f\sim \hat{\rho}} Y(f) - KL(\hat{\rho} \|\pi)) = \ln E_{f\sim\pi} e^{\lambda(X(f)-Y(f))}, \label{T:pacAlt:eq0.1}
	\end{align}
In particular, 
  \begin{align}
		 e^{\sup_{\hat{\rho}} (\lambda E_{f\sim \hat{\rho}} X(f) - \lambda E_{f\sim \hat{\rho}} Y(f) - KL(\hat{\rho} \|\pi))} = e^{\ln E_{f\sim\pi} e^{\lambda(X(f)- Y(f))}}= E_{f\sim\pi} e^{\lambda(X(f)-Y(f))} \label{T:pacAlt:eq0.2}
	\end{align}
  and hence
  \begin{align}
		 \bE[e^{\sup_{\hat{\rho}} (\lambda E_{f\sim \hat{\rho}} X(f) - \lambda E_{f\sim \hat{\rho}} Y(f) - KL(\hat{\rho} \|\pi))}]=\bE [E_{f\sim\pi} e^{\lambda(X(f)- Y(f))}] = \\ \nonumber
     E_{f\sim\pi} \bE[e^{\lambda(X(f)-Y(f))}]=e^{\Psi_{\pi}(\lambda,N)}
   \label{T:pacAlt:eq0.3}
	\end{align}
  with
    \begin{equation}
        \Psi_{\pi}(\lambda,N)=	\ln E_{f\sim\pi} \bE[e^{\lambda(X(f)-Y(f))}]
    \end{equation}
 Hence,
 \begin{align}
		 \bE[e^{\sup_{\hat{\rho}} (\lambda E_{f\sim \hat{\rho}} X(f) - \lambda E_{f\sim \hat{\rho}} Y(f) - KL(\hat{\rho} \|\pi)}]e^{-\Psi_{\pi}(\lambda,N)} = 1
   \label{T:pacAlt:eq0.4}
	\end{align}
Since 
 \begin{align}
		 \bE[e^{\sup_{\hat{\rho}} (\lambda E_{f\sim \hat{\rho}} X(f) - \lambda E_{f\sim \hat{\rho}} Y(f) - KL(\hat{\rho} \|\pi)}]e^{-\Psi_{\pi}(\lambda,N)} =  \nonumber \\
   \bE[e^{\sup_{\hat{\rho}} (\lambda E_{f\sim \hat{\rho}} X(f) - \lambda E_{f\sim \hat{\rho}} Y(f) - KL(\hat{\rho} \|\pi)-\Psi_{\pi}(\lambda,N)}]
   \label{T:pacAlt:eq0.5}
	\end{align}
 it follows that 
 \begin{align}
   \bE[e^{\sup_{\hat{\rho}} (\lambda E_{f\sim \hat{\rho}} X(f) - \lambda E_{f\sim \hat{\rho}} Y(f) - KL(\hat{\rho} \|\pi))-\Psi_{\pi}(\lambda,N)}] = 1
   \label{T:pacAlt:eq0.6}
	\end{align}
 By Chernoff's bound applied to
 the random variable
 $\mathcal{X}=\sup_{\hat{\rho}} (\lambda E_{f\sim \hat{\rho}} (f) - \lambda E_{f\sim \hat{\rho}} Y(f) - KL(\hat{\rho} \|\pi))-\Psi_{\pi}(\lambda,N)$ it then follows that for any $a > 0$
 \begin{align*}
  \bP(\mathcal{X} \ge a) \le  \frac{E[e^{\mathcal{X}}]}{e^{a}} \le e^{-a}
\end{align*}
By choosing $a=\ln \frac{1}{\delta}$, it follows that
\begin{align*}
  \bP(\mathcal{X} \ge \ln \frac{1}{\delta}) \le \delta
\end{align*}
and hence, 
\begin{align*}
  \bP(\mathcal{X} \le  \ln \frac{1}{\delta}) \ge 1-\delta
\end{align*}
By substituting the definition  of $\mathcal{X}$ and regrouping 
the terms, it then  follows that
\begin{align*}
  \bP( \sup_{\hat{\rho}} (\lambda E_{f\sim \hat{\rho}} X (f) - \lambda E_{f\sim \hat{\rho}} Y(f) - KL(\hat{\rho} \|\pi)) \le  \ln \frac{1}{\delta}+\Psi_{\pi}(\lambda,N)) \ge 1-\delta
\end{align*}
Note that 
\begin{align*}
\{ \omega  \mid \sup_{\hat{\rho}} (\lambda E_{f\sim \hat{\rho}} X (f) - \lambda E_{f\sim \hat{\rho}} Y(f)(\omega) - KL(\hat{\rho} \|\pi)) \le  \ln \frac{1}{\delta}+\Psi_{\pi}(\lambda,N) \}= \\
\{
   \omega \mid \forall \hat{\rho}: 
   E_{f\sim \hat{\rho}} X(f) \le
     E_{f\sim \hat{\rho}} Y(f)(\omega) +
     \dfrac{1}{\lambda}\!\left[KL(\hat{\rho} \|\pi) + \ln\dfrac{1}{\delta}	+ \Psi_{\pi}(\lambda,N) \right
\}
\end{align*}
and hence
it then follows that
	with probability at least $1-\delta$, the following 
	holds
	\begin{align}
		 \forall\rho:\quad &E_{f\sim \hat{\rho}} X (f) \le \  E_{f\sim \hat{\rho}} Y(f) +\dfrac{1}{\lambda}\!\left[KL(\hat{\rho} \|\pi) + \ln\dfrac{1}{\delta}	+ \Psi_{\pi}(\lambda,N) \right ],
	\end{align}
 \end{Proof}

\begin{Corollary}\label{cor:PACKL_L-V} By Lemma \ref{lemma:PAC-BayesianKL_General}, and Lemma \ref{lem:mgf}, for $0<\lambda\leq \inf_{f\in\mathcal{F}} \left ( 3(m+1) n_y \mu_{\max}(Q_e)G_e(f)^{2} \right )^{-1}$, with $\mathcal{M}_\pi$, denoting the set of all absolutely continuous probability densities w.r.t. $\pi$, then with probability at least $1-\delta$, the following holds
	\begin{align}
		 \forall\rho\in\mathcal{M}_\pi:\quad &E_{f\sim \hat{\rho}} \mathcal{L} (f) \le  E_{f\sim \hat{\rho}} V_N(f) +\dfrac{1}{\lambda}\!\left[KL(\hat{\rho} \|\pi) + \ln\dfrac{1}{\delta}	+ \widehat{\Psi}_{\pi,1}(\lambda,N) \right ],
	\end{align}
 with 
 \begin{align}
     \widehat{\Psi}_{\pi,1}(\lambda,N) \triangleq \ln E_{f\sim\pi} \left ( 1+\frac{2}{N}\frac{(m+1)! \left (3\lambda n_y\mu_{\max}(Q_e)G_e(f)^{2}\right )^2}{(1-3(m+1)\lambda n_y\mu_{\max}(Q_e)G_e(f)^{2})} \right )
 \end{align}
\end{Corollary}

\begin{Corollary}\label{cor:PACKL_V-hatL}
    By Lemma \ref{lemma:PAC-BayesianKL_General}, and Lemma \ref{lemma:mgf(Vn-hatL)}, for $0<\lambda \leq \inf_{f\in\mathcal{F}} \left ( 4 n_w \bar{G}_{gen}  \bar{G}_f(f)\right )^{-1}$, with $\mathcal{M}_\pi$, denoting the set of all absolutely continuous probability densities w.r.t. $\pi$, then with probability at least $1-\delta$, the following holds
    	\begin{align}
    		 \forall\rho\in\mathcal{M}_\pi:\quad &E_{f\sim \hat{\rho}} V_N (f) \le  E_{f\sim \hat{\rho}} \hat{\mathcal{L}}_N(f) +\dfrac{1}{\lambda}\!\left[KL(\hat{\rho} \|\pi) + \ln\dfrac{1}{\delta}	+ \widehat{\Psi}_{\pi,2}(\lambda,N) \right ],
    	\end{align}
     with 
    \begin{align}
     \widehat{\Psi}_{\pi,2}(\lambda,N) \triangleq \ln E_{f\sim\pi } \left ( 1+ \frac{(n_y+n_u)!}{\sqrt{N}} \frac{4 \lambda \bar{G}_{gen}  \bar{G}_f(f)}{1-4 \lambda (n_y+n_u)\bar{G}_{gen}  \bar{G}_f(f)} \right )
    \end{align}

\end{Corollary}

\begin{lemma} \label{lemma:unbounded_full_KL}
     For
         \begin{align}
             0<\tilde{\lambda} \leq \frac{1}{2}\Big ( \sup_{f\in\mathcal{F}} \max\{ 3(m+1) n_y \mu_{\max}(Q_e)G_e(f)^{2} , 4 n_w \bar{G}_{gen}  \bar{G}_f(f) \} \Big )^{-1}
         \end{align}with probability at least $1-2\delta$, the following holds
         \begin{align}
             \forall\rho\in\mathcal{M}_\pi:\quad &E_{f\sim \hat{\rho}} \mathcal{L} (f) \le  E_{f\sim \hat{\rho}} \hat{\mathcal{L}}_N(f) +\dfrac{1}{\tilde{\lambda}}\!\left[KL(\hat{\rho} \|\pi) + \ln\dfrac{1}{\delta}	+ \frac{\widehat{\Psi}_{\pi,2}(2\tilde{\lambda},N)+ \widehat{\Psi}_{\pi,1}(2\tilde{\lambda},N)}{2} \right ]  
         \end{align}
         with 
         \begin{align}
             \widehat{\Psi}_{\pi,1}(2\tilde{\lambda},N) = \Psi_{\pi,1}(\tilde{\lambda},N) = \ln E_{f\sim\pi} \left ( 1+\frac{2}{N}\frac{(m+1)! \left (6\tilde{\lambda} n_y\mu_{\max}(Q_e)G_e(f)^{2}\right )^2}{(1-6(m+1)\tilde{\lambda} n_y\mu_{\max}(Q_e)G_e(f)^{2})} \right )\\
             \widehat{\Psi}_{\pi,2}(2\tilde{\lambda},N) = \Psi_{\pi,2}(\tilde{\lambda},N) = \ln E_{f\sim\pi } \left ( 1+ \frac{(n_y+n_u)!}{\sqrt{N}} \frac{8 \tilde{\lambda} \bar{G}_{gen}  \bar{G}_f(f)}{1-8 \tilde{\lambda} (n_y+n_u)\bar{G}_{gen}  \bar{G}_f(f)} \right )
         \end{align} 
     
     \begin{Proof} \label{proof:thm:unbounded}
         we have 
         \begin{align}
             P(\omega\in S_1)\geq 1-\delta \\
             P(\omega\in S_2)\geq 1-\delta
         \end{align}
         with 
         \begin{align}
             S_1\triangleq\{\omega\in\Omega|\forall\rho\in\mathcal{M}_\pi:\quad &E_{f\sim \hat{\rho}} \mathcal{L} (f) \le  E_{f\sim \hat{\rho}} V_N(f) +\dfrac{1}{\lambda}\!\left[KL(\hat{\rho} \|\pi) + \ln\dfrac{1}{\delta}	+ \widehat{\Psi}_{\pi,1}(\lambda,N) \right ] \}\\
             S_2\triangleq \{\omega\in\Omega|\forall\rho\in\mathcal{M}_\pi:\quad &E_{f\sim \hat{\rho}} V_N (f) \le  E_{f\sim \hat{\rho}} \hat{\mathcal{L}}_N(f) +\dfrac{1}{\lambda}\!\left[KL(\hat{\rho} \|\pi) + \ln\dfrac{1}{\delta}	+ \widehat{\Psi}_{\pi,2}(\lambda,N) \right ] \}
         \end{align}
         with $\bar{A}$ denoting the complementary set of $A$, i.e. $\bar{A}=\Omega\setminus A$
         \begin{align}
             P(\omega\in \bar{S}_1)<\delta\\
             P(\omega\in \bar{S}_2)<\delta\\
         \end{align}
         Thus by union bound we get
         \begin{align}
             P\left (\omega\in (\bar{S}_1 \cup \bar{S}_2)\right )<2\delta
         \end{align}
         and thus
          \begin{align}
             P\left (\omega\in (S_1 \cap S_2)\right )\geq 1-2\delta
         \end{align}
         with this we can write: with probability at least $1-2\delta$, the following holds
         \begin{align}
             \forall\rho\in\mathcal{M}_\pi:\quad &E_{f\sim \hat{\rho}} \mathcal{L} (f) \le  E_{f\sim \hat{\rho}} \hat{\mathcal{L}}_N(f) +\dfrac{2}{\lambda}\!\left[KL(\hat{\rho} \|\pi) + \ln\dfrac{1}{\delta}	+ \frac{\widehat{\Psi}_{\pi,2}(\lambda,N)+ \widehat{\Psi}_{\pi,1}(\lambda,N)}{2} \right ]  
         \end{align}
         In order to bring this to a more common way of writing PAC-Bayesian bounds, let us define $\tilde{\lambda}=0.5\lambda \leftrightarrow \lambda = 2\tilde{\lambda}$, thus we can write, for
         \begin{align}
             0<\tilde{\lambda} \leq \frac{1}{2}\Big ( \sup_{f\in\mathcal{F}} \max\{ 3(m+1) n_y \mu_{\max}(Q_e)G_e(f)^{2} , 4 n_w \bar{G}_{gen}  \bar{G}_f(f) \} \Big )^{-1}
         \end{align}with probability at least $1-2\delta$, the following holds
         \begin{align}
             \forall\rho\in\mathcal{M}_\pi:\quad &E_{f\sim \hat{\rho}} \mathcal{L} (f) \le  E_{f\sim \hat{\rho}} \hat{\mathcal{L}}_N(f) +\dfrac{1}{\tilde{\lambda}}\!\left[KL(\hat{\rho} \|\pi) + \ln\dfrac{1}{\delta}	+ \frac{\widehat{\Psi}_{\pi,2}(2\tilde{\lambda},N)+ \widehat{\Psi}_{\pi,1}(2\tilde{\lambda},N)}{2} \right ]  
         \end{align}
         with 
         \begin{align}
             \widehat{\Psi}_{\pi,1}(2\tilde{\lambda},N) = \Psi_{\pi,1}(\tilde{\lambda},N) = \ln E_{f\sim\pi} \left ( 1+\frac{2}{N}\frac{(m+1)! \left (6\tilde{\lambda} n_y\mu_{\max}(Q_e)G_e(f)^{2}\right )^2}{(1-6(m+1)\tilde{\lambda} n_y\mu_{\max}(Q_e)G_e(f)^{2})} \right )\\
             \widehat{\Psi}_{\pi,2}(2\tilde{\lambda},N) = \Psi_{\pi,2}(\tilde{\lambda},N) = \ln E_{f\sim\pi } \left ( 1+ \frac{(n_y+n_u)!}{\sqrt{N}} \frac{8 \tilde{\lambda} \bar{G}_{gen}  \bar{G}_f(f)}{1-8 \tilde{\lambda} (n_y+n_u)\bar{G}_{gen}  \bar{G}_f(f)} \right )
         \end{align}
     \end{Proof}
\end{lemma}

\subsection{Bounded noise}
In this section we state the lemmas and proofs associated with bounded innovation noise case.
\begin{lemma}\label{lemma:bounded_e_moments} Let $\e_g(t)\in \mathcal{E}\subset \mathbb{R}^{n_y+n_y}$, be a zero mean, independant, and bounded stochastic process, s.t. $|\e_{g,i}(t)|\leq c_e$, $\forall i\in \{1,\dots,nu+ny\}$, i.e $\e_{g,i}(t)$ is the $i$'th component of $\e_g(t)$
\begin{align}
    \bE[\|\e_g(t)\|^r]\leq  \left ( c_e \sqrt{n_y+n_u} \right )^r
\end{align}

\begin{Proof}
    \begin{align}
            \bE[\|\e_g(t)\|^r]=\bE\left [\left ( \sqrt{\sum_{i=1}^{nu+ny} \e_{g,i}^2(t)} \right)^r \right ] \leq \left ( \sqrt{\sum_{i=1}^{nu+ny} c_e^2} \right)^r = \left ( \sqrt{ (n_u+n_y) c_e^2} \right)^r = \left ( c_e \sqrt{n_y+n_u} \right )^r
    \end{align}
\end{Proof}
\end{lemma}

\begin{lemma}\label{lemma:bounded_sigmar} Let $\e_g(t)\in \mathcal{E}\subset \mathbb{R}^{n_y+n_y}$, be a zero mean, independant, and bounded stochastic process, s.t. $|\e_{g,i}(t)|\leq c_e$, $\forall i\in \{1,\dots,nu+ny\}$, i.e $\e_{g,i}(t)$ is the $i$'th component of $\e_g(t)$
    \begin{align}
        \sigma(r)= \left ( 2c_e^2 (n_y+n_u) \right )^r \geq \sup_{t,k,l} \bE[\|e(t,k,l)\|^r_2]\\
        e(t,k,l) = \bE[\e_g(t-k)\e_g^T(t-l)]-\e_g(t-k)\e_g^T(t-l)
    \end{align}
\begin{Proof}
 %    For the case of $k=l$, we have
 %    \begin{align}
 %        \bE\left [ \| \bE[\e_g(t-k)\e_g^T(t-l)]-\e_g(t-k)\e_g^T(t-l) \|^r \right ]
 %    \end{align}
 %        	Recall that
	% \begin{align*}
	% 	\e(t,k,j)=\begin{cases} Q_e-\e_g(t-k)\e_g^T(t-j),& k=j\\ -\e_g(t-k)\e_g^T(t-j),& k\neq j \end{cases}
	% \end{align*}
	First let us take the case when $k\neq j$. Then, due to independance of $\e_g(t)$, we have $\bE[\e_g(t-k)\e_g(t-j)]=0$, and thus
	\begin{align*}
		\bE[\|\e(t,k,l)\|_2^r]=\bE[\|-\e_g(t-k)\e_g^T(t-j)\|_2^r]
	\end{align*}
	Again as $\e_g(t)$ is i.i.d. we have 
	\begin{align*}
		\bE[\|\e(t,k,l)\|_2^r]\leq \bE[\left ( \|\e_g(t-k)\|_2\|\e_g^T(t-j)\|_2\right )^r] \leq  \bE[\|\e_g(t-k)\|_2^r]\bE[\|\e_g(t-j)\|_2^r]
	\end{align*}
	and due to stationarity of $\e_g(t)$, we have $\bE[\|\e_g(t-k)\|_2^r]=\bE[\|\e_g(t-j)\|_2^r]$, therefore
	\begin{align*}
		\bE[\|\e(t,k,l)\|_2^r]\leq \bE[\|\e_g(t)\|_2^r]^2
	\end{align*}
	and again due to stationarity of $\e_g(t)$, the moments do not depend on $t$, and using Lemma \ref{lemma:bounded_e_moments} we obtain
	\begin{align*}
	\forall k\neq j, \;	\bE[\|\e(t,k,l)\|_2^r] \leq  \left ( c_e^2 (n_y+n_u) \right )^r
	\end{align*}
	Now let us take the case when $k=j$. Then 
 \begin{align}
     \bE\left [ \| \bE[\e_g(t-k)\e_g^T(t-l)]-\e_g(t-k)\e_g^T(t-l) \|^r \right ] \leq \bE\left [ \left (  \| \bE[\e_g(t-k)\e_g^T(t-l)]\|+ \| \e_g(t-k)\e_g^T(t-l) \| \right )^r \right ]
 \end{align}
 By convexity $(a+b)^r = 2^r\frac{1}{2^r}(a+b)^r = 2^r\left (\frac{1}{2}(a+b) \right )^r \leq 2^{r-1}(a^r+b^r)$, we obtain
 \begin{align}
     \bE[\|\e(t,k,l)\|_2^r] \leq 2^{r-1} \left ( \bE\left [\| \bE[\e_g(t-k)\e_g^T(t-l)]\|^r \right ]+ \bE \left [ \| \e_g(t-k)\e_g^T(t-l) \|^r \right ] \right )\\
     =2^{r-1} \left ( \| \bE[\e_g(t-k)\e_g^T(t-l)]\|^r + \bE \left [ \| \e_g(t-k)\e_g^T(t-l) \|^r \right ] \right )\\
     \leq 2^{r-1} \left (  \bE[\|\e_g(t-k)\e_g^T(t-l)\|^r] + \bE \left [ \| \e_g(t-k)\e_g^T(t-l) \|^r \right ] \right )
     \leq 2^r \bE \left [ \| \e_g(t)\|^{2r} \right ]
 \end{align}
  Again by using Lemma \ref{lemma:bounded_e_moments}, we obtain
  \begin{align}
      \forall k=j \; \bE[\|\e(t,k,l)\|_2^r] \leq  \left ( 2c_e^2 (n_y+n_u) \right )^r
  \end{align}
  Thus we obtain the statement of the lemma 
    \begin{align}
        \forall t,k,j\; \bE[\|\e(t,k,l)\|_2^r] \leq \max \{ \left ( c_e^2 (n_y+n_u) \right )^r, \left ( 2c_e^2 (n_y+n_u) \right )^r \} = \left ( 2c_e^2 (n_y+n_u) \right )^r
    \end{align}
\end{Proof}

\end{lemma}

\begin{lemma}\label{lemma:bounded_mgf(L-V)} With notation as above, with $|\e_{g,i}|\leq c_e$, the following holds
    \begin{align}
        \bE[e^{\lambda(\mathcal{L}(f)-V_N(f))}]\leq 1+\frac{1}{N}e^{\lambda 4c_e^2n_y(n_y+n_u)G_e(f)^2}
    \end{align}

    \begin{Proof} By power series, and $\bE[\mathcal{L}(f)-V_N(f)]=0$, we have 
        \begin{align}
            \bE[e^{\lambda(\mathcal{L}(f)-V_N(f))}] = 1 + \sum_{r=2}^\infty \frac{\lambda^r}{r!} \bE[(\mathcal{L}(f)-V_N(f))^r]
        \end{align}
        Now by Lemma \ref{lemma:L-Vrfull}, and Lemma \ref{lemma:bounded_sigmar}, and $4(r-1)\leq 2^r$ we have
        \begin{align}
            \bE[(\mathcal{L}(f)-V_N(f))^r] \leq \frac{1}{N} (4c_e^2n_y(n_y+n_u)G_e(f)^2)^r 
        \end{align}
        Thus, 
        \begin{align}
            \bE[e^{\lambda(\mathcal{L}(f)-V_N(f))}] \leq 1 + \frac{1}{N} \sum_{r=2}^\infty \frac{1}{r!}  (\lambda 4c_e^2n_y(n_y+n_u)G_e(f)^2)^r 
        \end{align}
        now since $\lambda 4c_e^2n_y(n_y+n_u)G_e(f)^2\geq 0$, then
        \begin{align}
             1 + \frac{1}{N} \sum_{r=2}^\infty \frac{1}{r!}  (\lambda 4c_e^2n_y(n_y+n_u)G_e(f)^2)^r \\
             \leq 1 + \frac{1}{N} \sum_{r=0}^\infty \frac{1}{r!}  (\lambda 4c_e^2n_y(n_y+n_u)G_e(f)^2)^r\\
             = 1+\frac{1}{N}e^{\lambda 4c_e^2n_y(n_y+n_u)G_e(f)^2}
        \end{align}
    \end{Proof}
\end{lemma}

\begin{lemma}\label{lemma:bounded_mgf(V-hatL)}
With notation as above, with $|\e_{g,i}|\leq c_e$, the following holds
    \begin{align}
        \bE[e^{\lambda(V_N(f)-\hat{\mathcal{L}}(f))}]\leq 1+\frac{1}{\sqrt{N}}e^{2\lambda G_f(f) \|\Sigma_{gen}\|_{\ell_1}^2 c_e^2 (n_y+n_u)}
    \end{align}
    with 
    \begin{align}
        G_f(f)\triangleq \left (1+\|\hat{D}\|+ \frac{\hat{M}\|\hat{B}\|\|\hat{C}\|}{1-\hat{\gamma}} \right ) \left (\frac{\hat{M}\|\hat{C}\| \|\hat{B}\|}{(1-\hat{\gamma})^\frac{3}{2}} \right )
    \end{align}

    \begin{Proof}
        By power series, we have 
        \begin{align}
            \bE[e^{\lambda(V_N(f)-\hat{\mathcal{L}}(f))}]\leq \bE[e^{\lambda|V_N(f)-\hat{\mathcal{L}}(f)|}] = 1 + \sum_{r=1}^\infty \frac{\lambda^r}{r!} \bE[|V_N(f)-\hat{\mathcal{L}}(f)|^r]
        \end{align}
        For the terms $\bE[|V_N(f)-\hat{\mathcal{L}}_N(f)|^r]$, we reuse the proof of Lemma \ref{lemma:moments_|V-Lhat|}, and continue from \eqref{eq:moments_|V-Lhat|_pre_yw}, i.e.
        \begin{align}
        \bE[\|V_N(f)-\hat{\mathcal{L}}_N(f)\|^r] \leq \frac{2^{r}}{\sqrt{N}} \left (1+\|\hat{D}\|+ \frac{\hat{M}\|\hat{B}\|\|\hat{C}\|}{1-\hat{\gamma}} \right )^{r} \left (\frac{\hat{M}\|\hat{C}\| \|\hat{B}\|}{1-\hat{\gamma}} \right )^{r} \bE\left [ \left \|\begin{bmatrix} \y(t) \\ \mathbf{u}(t) \end{bmatrix}\right \|^{2r} \right ]  \sqrt{\frac{1}{1-\hat{\gamma}^{2r}}}
    \end{align}
    Note that 
    \begin{align}
       \left ( 1-\hat{\gamma}\right )^\frac{r}{2} \leq  \left ( 1-\hat{\gamma}^{2r}\right )^{\frac{1}{2}}
    \end{align}
    it is easy to see since for $\hat{\gamma}\in [0,1)$, the following holds 
    \begin{align}
        \left ( 1-\hat{\gamma}\right )^r &\leq  1-\hat{\gamma}^{2r} = (1-\hat{\gamma}^r)(1+\hat{\gamma}^r)\\
        1 &\leq 1+\hat{\gamma}^r
    \end{align}
    This allows us to simplify the expression to
    \begin{align}
        \bE[\|V_N(f)-\hat{\mathcal{L}}_N(f)\|^r] \leq \frac{2^{r}}{\sqrt{N}} \left (1+\|\hat{D}\|+ \frac{\hat{M}\|\hat{B}\|\|\hat{C}\|}{1-\hat{\gamma}} \right )^{r} \left (\frac{\hat{M}\|\hat{C}\| \|\hat{B}\|}{1-\hat{\gamma}} \right )^{r} \bE\left [ \left \|\begin{bmatrix} \y(t) \\ \mathbf{u}(t) \end{bmatrix}\right \|^{2r} \right ] \left ( \frac{1}{\sqrt{1-\hat{\gamma}}} \right )^r 
    \end{align}
    Now, from Lemma \ref{lemma:stationary4thMoment}, we get
    \begin{align}
        \bE\left [ \left \|\begin{bmatrix} \y(t) \\ \mathbf{u}(t) \end{bmatrix}\right \|^{2r} \right ] \leq \|\Sigma_{gen}\|_{\ell_1}^{2r} \bE[\|\e_g(t)\|^{2r}]
    \end{align}
    by lemma \ref{lemma:bounded_e_moments}, we get
        \begin{align}
        \bE\left [ \left \|\begin{bmatrix} \y(t) \\ \mathbf{u}(t) \end{bmatrix}\right \|^{2r} \right ] \leq \left ( \|\Sigma_{gen}\|_{\ell_1}^2 c_e^2 (n_y+n_u) \right )^r
    \end{align}
    Thus, with $G_f(f)\triangleq  \dfrac{1}{\sqrt{1-\hat{\gamma}}}\left (1+\|\hat{D}\|+ \frac{\hat{M}\|\hat{B}\|\|\hat{C}\|}{1-\hat{\gamma}} \right ) \left (\frac{\hat{M}\|\hat{C}\| \|\hat{B}\|}{1-\hat{\gamma}} \right )$
    \begin{align}
        \bE[\|V_N(f)-\hat{\mathcal{L}}_N(f)\|^r] \leq  \frac{1}{\sqrt{N}}   \left (2G_f(f) \|\Sigma_{gen}\|_{\ell_1}^2 c_e^2 (n_y+n_u) \right )^r 
    \end{align}
    Thus
    \begin{align}
        \bE[e^{\lambda|V_N(f)-\hat{\mathcal{L}}(f)|}] &\leq 1 + \frac{1}{\sqrt{N}} \sum_{r=1}^\infty \frac{1}{r!} \left (2\lambda G_f(f) \|\Sigma_{gen}\|_{\ell_1}^2 c_e^2 (n_y+n_u) \right )^r \\
        &\leq 1+\frac{1}{\sqrt{N}}e^{2\lambda G_f(f) \|\Sigma_{gen}\|_{\ell_1}^2 c_e^2 (n_y+n_u)}
        \end{align}
    and therefore the statement of the lemma holds.
    \end{Proof}
    
\end{lemma}

\begin{Corollary} \label{cor:thm:bounded}
    By lemma \ref{lemma:PAC-BayesianKL_General}, lemmas \ref{lemma:bounded_mgf(L-V)},\ref{lemma:bounded_mgf(V-hatL)}, and by applying a union bound, we obtain, for $\lambda>0$, $\delta\in[0,1)$, the set of absolutely continuous probability density functions $\mathcal{M}_\pi$ w.r.t. $\pi$, the following holds with probability at least $1-2\delta$
    \begin{align}
        \forall \rho\in\mathcal{M}_\pi:\quad E_{f\sim\rho} \mathcal{L}(f) \leq E_{f\sim\rho }\hat{\mathcal{L}}_N(f) + \frac{1}{\lambda}\left [\KL(\rho||\pi)+\ln\left (\frac{1}{\delta}\right ) + \widehat{\Psi}_{c_e,\pi}(\lambda,N)\right ]
    \end{align}
    with 
    \begin{align}
        \widehat{\Psi}_{c_e,\pi}(\lambda,N) \triangleq \frac{1}{2} \left ( \widehat{\Psi}_{c_e,\pi,1}(\lambda,N)+\widehat{\Psi}_{c_e,\pi,2}(\lambda,N)\right )\\
        \widehat{\Psi}_{c_e,\pi,1}(\lambda,N) \triangleq \ln E_{f\sim\pi } \left ( 1+\frac{1}{N}e^{\lambda 4c_e^2n_y(n_y+n_u)G_e(f)^2} \right )\\
        \widehat{\Psi}_{c_e,\pi,2}(\lambda,N) \triangleq \ln E_{f\sim\pi } \left (1+\frac{1}{\sqrt{N}}e^{2\lambda G_f(f) \|\Sigma_{gen}\|_{\ell_1}^2 c_e^2 (n_y+n_u)} \right )
    \end{align}
\end{Corollary}

\subsection{Bounded innovation noise case: Alternative formulation}

\begin{lemma}\label{lem:repeatedCauchy}
    for a sequence of random variables $x_j\in\mathbb{R}$, and $j\in\{1,\dots,r\}$
    \begin{align}
        \bE\left [\prod_{j=1}^r x_j \right ] \leq \left ( \prod_{j=1}^{r-1} \bE \left [ x_j^{(2^j)} \right ]^{(2^{-j})} \right ) \bE \left [ x_r^{(2^{r-1})} \right ]^{2^{-(r-1)}}
    \end{align}

\end{lemma}
\begin{Proof}[of Lemma \ref{lem:repeatedCauchy}]
\renewcommand{\bE}[1]{\mathbf{E}\left [ #1 \right ]}
We first apply Cauchy-Schwarz inequality $\bE{\prod_{j=1}^r x_j}\leq |\bE{\prod_{j=1}^r x_j}| =|\bE{\left ( x_1 \right ) \left ( \prod_{j=2}^r x_j \right )}| \leq \sqrt{\bE{x_1^2}}\sqrt{\bE{\prod_{j=2}^r x_j^2}}$, and obtain
\begin{align}
    \bE{\prod_{j=1}^r x_j} \leq \bE{x_j^2 }^{2^{-1}}\bE{\prod_{j=2}^r x_j^2 }^{2^{-1}}
\end{align}
Then we apply Cauchy-Schwarz again
\begin{align}
    \bE{\prod_{j=1}^r x_j} \leq \bE{x_1^2 }^{2^{-1}}\bE{x_2^{(2^2)} }^{2^{-2}}\bE{\prod_{j=3}^r x_j^{(2^2)} }^{2^{-2}} = \prod_{j=1}^2 \bE{x_j^{(2^j)}}^{(2^{-j})} \bE{\prod_{j=2+1}^r x_j^{(2^2)} }^{2^{-2}}
\end{align}
We repeat this process until we have
\begin{align}
    \bE{\prod_{j=1}^r x_j} \leq \prod_{j=1}^{r-2} \bE{x_j^{(2^j)}}^{(2^{-j})} \bE{x_{r-1}^{(2^{r-2})} x_r^{(2^{r-2})} }^{2^{-(r-2)}}
\end{align}
Then we apply the final Cauchy-Schwarz inequality and obtain the statement of the lemma
\begin{align}
    \bE{\prod_{j=1}^r x_j} &\leq \prod_{j=1}^{r-2} \bE{x_j^{(2^j)}}^{(2^{-j})} \bE{x_{r-1}^{(2^{r-1})} }^{2^{-(r-1)}}\bE{x_r^{(2^{r-1})} }^{2^{-(r-1)}}\\
    &=\prod_{j=1}^{r-1} \bE{x_j^{(2^j)}}^{(2^{-j})} \bE{x_r^{(2^{r-1})} }^{2^{-(r-1)}}
\end{align}

\end{Proof}

\begin{lemma} \label{lemma:newE|V-Lhat|^r}
Let $m=n_y+n_u$. If $|e_g(t)| < c_e$, then 
        \begin{align}
         \bE[\|V_N(f)-\hat{\mathcal{L}}_N(f)\|^r] \leq  \bar{G}_{f,1}(f) \|\Sigma_{gen}\|_{\ell_1} (c_e\sqrt{m}) \left (\frac{2\|\Sigma_{gen}\|_{\ell_1}(c_e\sqrt{m})}{N} \bar{G}_{f,2}(f) \right )^r
    \end{align}
    where 
    $\bar{G}_{f,1}(f)\triangleq \left (  \frac{\hat{M} \| \hat{C} \| \|\hat{B}\|}{1-\hat{\gamma}} \right ) $, and $\bar{G}_{f,2}(f)\triangleq \left (1+\|\hat{D}\|+ \frac{\hat{M}\|\hat{B}\|\|\hat{C}\|}{1-\hat{\gamma}} \right ) \frac{1}{1-\hat{\gamma}}$
    %\begin{align}   
    %     \left (  \frac{4\bar{G}_f(f)\|\Sigma_{gen}\|_{\ell_1}^{3} (c_e \sqrt{m}) }{N} \right )^{r} ,
    %\end{align}
    %    with 
    %\begin{align}
        %\bar{G}_f(f)&\triangleq \frac{1}{1-\hat{\gamma}}\left (  \frac{\hat{M} \| \hat{C} \| \|\hat{B}\|}{1-\hat{\gamma}} \right )\left (1+\|\hat{D}\|+ \frac{\hat{M}\|\hat{B}\|\|\hat{C}\|}{1-%\hat{\gamma}} \right ),\\
        \( \|\Sigma_{gen}\|_{\ell_1}\triangleq\|I\|+\sum_{k=1}^\infty \|C_gA_g^{k-1}K_g\| \).
    %\end{align}
\end{lemma}

\begin{Proof}[of Lemma \ref{lemma:newE|V-Lhat|^r}]
        with $\z_\infty(t)=\y(t)-\hat{\y}_f(t)$, and $\z_f(t)=\y(t)-\hat{\y}_f(t|0)$, we start by applying triangle inequalities
    \begin{multline}
        \bE[\|V_N(f)-\hat{\mathcal{L}}_N(f)\|^r] = \bE \left [\left | \frac{1}{N}\sum_{t=0}^{N-1} \|\z_\infty(t)\|^2-\|\z_f(t)\|^2   \right |^r \right ] \leq \bE \left [ \left (  \frac{1}{N}\sum_{t=0}^{N-1} \left | \|\z_\infty(t)\|^2-\|\z_f(t)\|^2   \right | \right )^r \right ]
    \end{multline}
    \begin{align}
        \bE[\|V_N(f)-\hat{\mathcal{L}}_N(f)\|^r] \leq \frac{1}{N^r} \sum_{t_1=0}^{N-1} \dots \sum_{t_r=0}^{N-1} \bE \left [  \prod_{j=1}^r \left | \|\z_\infty(t_j)\|^2-\|\z_f(t_j)\|^2 \right |\right ]
    \end{align}
    Now using the fact that $|a^2-b^2|=|(a-b)(a+b)| = |a-b|(a+b)$, since $a,b\geq 0$, we get
    \begin{align}
        \bE[\|V_N(f)-\hat{\mathcal{L}}_N(f)\|^r] \leq \frac{1}{N^r} \sum_{t_1=0}^{N-1} \dots \sum_{t_r=0}^{N-1} \bE \left [  \prod_{j=1}^r \left | \|\z_\infty(t_j)\|-\|\z_f(t_j)\| \right |\left ( \|\z_\infty(t_j)\|+\|\z_f(t_j)\| \right ) \right ]
    \end{align}
    We apply Cauchy-Schwarz, i.e. $\bE[XY]\leq |\bE[XY]| \leq \sqrt{\bE[X^2]} \sqrt{\bE[Y^2]}$, with $X=\prod_{j=1}^r \left | \|\z_\infty(t_j)\|-\|\z_f(t_j)\| \right |$, and $Y=\prod_{j=1}^r \left ( \|\z_\infty(t_j)\|+\|\z_f(t_j)\| \right )$, 
    \begin{align}
        \bE[\|V_N(f)-\hat{\mathcal{L}}_N(f)\|^r] \leq \frac{1}{N^r} \sum_{t_1=0}^{N-1} \dots \sum_{t_r=0}^{N-1} \sqrt{\bE \left [  \prod_{j=1}^r \left | \|\z_\infty(t_j)\|-\|\z_f(t_j)\| \right |^2 \right ] } \sqrt{\bE \left [ \prod_{j=1}^r \left ( \|\z_\infty(t_j)\|+\|\z_f(t_j)\| \right )^2 \right ]} \label{eq:ASDFH}
    \end{align}
    For now let's focus on $\bE \left [  \prod_{j=1}^r \left | \|\z_\infty(t_j)\|-\|\z_f(t_j)\| \right |^2 \right ]$, by applying reverse triangle inequality we obtain
    \begin{align}
        \bE \left [  \prod_{j=1}^r \left | \|\z_\infty(t_j)\|-\|\z_f(t_j)\| \right |^2 \right ] \leq \bE \left [  \prod_{j=1}^r  \|\z_\infty(t_j)-\z_f(t_j)\|^2 \right ]
    \end{align}
    For the ease of notation for the next step, let us define $x_j\triangleq \|\z_\infty(t_j)-\z_f(t_j)\|^2$, then the quantity of interest is
    \begin{align}
        \bE \left [  \prod_{j=1}^r  x_j \right ]
    \end{align}
    For the above quantity we can apply Lemma \ref{lem:repeatedCauchy}, which states
    \begin{align}
        \bE\left [\prod_{j=1}^r x_j \right ] \leq \prod_{j=1}^{r-1} \bE \left [ x_j^{(2^j)} \right ]^{(2^{-j})} \bE \left [ x_r^{(2^{r-1})} \right ]^{2^{-(r-1)}}
    \end{align}
    From Lemma \ref{lemma:E|zinf-zf|^r}, we also know that
    \begin{align}
        \bE[\|\z_\infty(t)-\z_f(t)\|^r] \leq \hat{\gamma}^{rt}\left (  \frac{\hat{M} \| \hat{C} \| \|\hat{B}\|}{1-\hat{\gamma}} \right )^r \bE \left [ \left \|\begin{bmatrix} \y(t) \\ \mathbf{u}(t) \end{bmatrix} \right \|^r \right ] 
    \end{align}
    Thus combining Lemma \ref{lem:repeatedCauchy} and Lemma \ref{lemma:E|zinf-zf|^r}, we get 
    \begin{multline}
        \bE \left [  \prod_{j=1}^r  \|\z_\infty(t_j)-\z_f(t_j)\|^2 \right ] \leq \prod_{j=1}^{r-1} \hat{\gamma}^{2t_j} \left (  \frac{\hat{M} \| \hat{C} \| \|\hat{B}\|}{1-\hat{\gamma}} \right )^2 \bE \left [ \left \|\begin{bmatrix} \y(t) \\ \mathbf{u}(t) \end{bmatrix} \right \|^{(2^{j+1})} \right ]^\frac{1}{2^j} \\
        \times \hat{\gamma}^{2t_r} \left (  \frac{\hat{M} \| \hat{C} \| \|\hat{B}\|}{1-\hat{\gamma}} \right )^2 \bE \left [ \left \|\begin{bmatrix} \y(t) \\ \mathbf{u}(t) \end{bmatrix} \right \|^{(2^{r})} \right ]^\frac{1}{2^{r-1}}
    \end{multline}
    with Lemma \ref{lemma:E|yw|^r}, and Lemma \ref{lemma:bounded_e_moments}, we have 
    \begin{align}
        \bE\left [ \left \|\begin{bmatrix} \y(t) \\ \mathbf{u}(t) \end{bmatrix}\right \|^{r} \right ] \leq \|\Sigma_{gen}\|_{\ell_1}^{r} (c_e\sqrt{m})^r,
    \end{align}
    thus we get 
    \begin{multline}
        \bE \left [  \prod_{j=1}^r  \|\z_\infty(t_j)-\z_f(t_j)\|^2 \right ] \leq \prod_{j=1}^{r-1} \hat{\gamma}^{2t_j} \left (  \frac{\hat{M} \| \hat{C} \| \|\hat{B}\|}{1-\hat{\gamma}} \right )^2  \left ( \|\Sigma_{gen}\|_{\ell_1}^{2^{j+1}} (c_e\sqrt{m})^{2^{j+1}} \right )^{2^{-j}} \\
        \cdot \hat{\gamma}^{2t_r} \left (  \frac{\hat{M} \| \hat{C} \| \|\hat{B}\|}{1-\hat{\gamma}} \right )^2  \left ( \|\Sigma_{gen}\|_{\ell_1}^{2r} (c_e\sqrt{m})^{2r} \right )^{2^{-(r-1)}},
    \end{multline}
    With some algebraic simplification we obtain the first term
    \begin{align}
        \bE \left [  \prod_{j=1}^r  \|\z_\infty(t_j)-\z_f(t_j)\|^2 \right ] \leq \left (  \frac{\hat{M} \| \hat{C} \| \|\hat{B}\|}{1-\hat{\gamma}} \right )^2 \|\Sigma_{gen}\|_{\ell_1}^{2} (c_e\sqrt{m})^{2}   \prod_{j=1}^{r} \hat{\gamma}^{2t_j},
    \end{align}

    Now for the second term $\bE \left [ \prod_{j=1}^r \left ( \|\z_\infty(t_j)\|+\|\z_f(t_j)\| \right )^2 \right ]$, we apply the inequality of arithmetic-geometric means
    \begin{align}
        \bE \left [ \prod_{j=1}^r \left ( \|\z_\infty(t_j)\|+\|\z_f(t_j)\| \right )^2 \right ] \leq \frac{1}{r} \sum_{j=1}^r \bE\left [ \left ( \|\z_\infty(t_j)\|+\|\z_f(t_j)\| \right )^{2r}\right ]
    \end{align}
    By Lemma \ref{lemma:(a+b)^2r}, we obtain
    \begin{align}
        \frac{1}{r} \sum_{j=1}^r \bE\left [ \left ( \|\z_\infty(t_j)\|+\|\z_f(t_j)\| \right )^{2r}\right ] \leq \frac{2^{2r-1}}{r} \sum_{j=1}^r  \left ( \bE\left [ \|\z_\infty(t_j)\|^{2r} \right] + \bE\left [ \|\z_f(t_j)\|^{2r} \right ] \right )
    \end{align}

    By Lemma \ref{lemma:E|z_inf|^r} and Lemma \ref{lemma:E|z_f|^r}, we obtain
    \begin{align}
         \frac{2^{2r-1}}{r} \sum_{j=1}^r  \left ( \bE\left [ \|\z_\infty(t_j)\|^{2r} \right] + \bE\left [ \|\z_f(t_j)\|^{2r} \right ] \right ) &\leq \frac{2^{2r}}{r} \sum_{j=1}^r \left (1+\|\hat{D}\|+ \frac{\hat{M}\|\hat{B}\|\|\hat{C}\|}{1-\hat{\gamma}} \right )^{2r} \bE\left [ \left \|\begin{bmatrix} \y(t) \\ \mathbf{u}(t) \end{bmatrix}\right \|^{2r} \right ]\\
         &=2^{2r} \left (1+\|\hat{D}\|+ \frac{\hat{M}\|\hat{B}\|\|\hat{C}\|}{1-\hat{\gamma}} \right )^{2r} \bE\left [ \left \|\begin{bmatrix} \y(t) \\ \mathbf{u}(t) \end{bmatrix}\right \|^{2r} \right ] 
    \end{align}
        with Lemma \ref{lemma:E|yw|^r}, and Lemma \ref{lemma:bounded_e_moments}, we have 
    \begin{align}
        \bE\left [ \left \|\begin{bmatrix} \y(t) \\ \mathbf{u}(t) \end{bmatrix}\right \|^{r} \right ] \leq \|\Sigma_{gen}\|_{\ell_1}^{r} (c_e\sqrt{m})^r,
    \end{align}
    we get
    \begin{align}
        \bE \left [ \prod_{j=1}^r \left ( \|\z_\infty(t_j)\|+\|\z_f(t_j)\| \right )^2 \right ] \leq \left ( 2\|\Sigma_{gen}\|_{\ell_1}(c_e\sqrt{m})\left (1+\|\hat{D}\|+ \frac{\hat{M}\|\hat{B}\|\|\hat{C}\|}{1-\hat{\gamma}} \right )  \right )^{2r}   \label{eq:ASDGDSFHG}
    \end{align}

    Now taking \eqref{eq:ASDGDSFHG} and \eqref{eq:aslkjgdsaf} back to \eqref{eq:ASDFH}, we have

    \begin{multline}
        \bE[\|V_N(f)-\hat{\mathcal{L}}_N(f)\|^r] \leq \frac{1}{N^r} \sum_{t_1=0}^{N-1} \dots \sum_{t_r=0}^{N-1} \sqrt{\bE \left [  \prod_{j=1}^r \left | \|\z_\infty(t_j)\|-\|\z_f(t_j)\| \right |^2 \right ] } \sqrt{\bE \left [ \prod_{j=1}^r \left ( \|\z_\infty(t_j)\|+\|\z_f(t_j)\| \right )^2 \right ]} \\
        \leq \frac{1}{N^r} \sum_{t_1=0}^{N-1} \dots \sum_{t_r=0}^{N-1} \sqrt{\left (  \frac{\hat{M} \| \hat{C} \| \|\hat{B}\|}{1-\hat{\gamma}} \right )^2 \|\Sigma_{gen}\|_{\ell_1}^{2} (c_e\sqrt{m})^{2}   \prod_{j=1}^{r} \hat{\gamma}^{2t_j}} \\
        \cdot \sqrt{\left ( 2\|\Sigma_{gen}\|_{\ell_1}(c_e\sqrt{m})\left (1+\|\hat{D}\|+ \frac{\hat{M}\|\hat{B}\|\|\hat{C}\|}{1-\hat{\gamma}} \right )  \right )^{2r} }
    \end{multline}
    with $G_f(f)\triangleq\left (  \frac{\hat{M} \| \hat{C} \| \|\hat{B}\|}{1-\hat{\gamma}} \right )\left (1+\|\hat{D}\|+ \frac{\hat{M}\|\hat{B}\|\|\hat{C}\|}{1-\hat{\gamma}} \right )$
    \begin{multline}
        \bE[\|V_N(f)-\hat{\mathcal{L}}_N(f)\|^r] \leq \left (  \frac{\hat{M} \| \hat{C} \| \|\hat{B}\|}{1-\hat{\gamma}} \right ) \|\Sigma_{gen}\|_{\ell_1} (c_e\sqrt{m}) \left ( 2\|\Sigma_{gen}\|_{\ell_1}(c_e\sqrt{m})\left (1+\|\hat{D}\|+ \frac{\hat{M}\|\hat{B}\|\|\hat{C}\|}{1-\hat{\gamma}} \right )  \right )^{r} \\
        \cdot \frac{1}{N^r} \sum_{t_1=0}^{N-1} \dots \sum_{t_r=0}^{N-1}\prod_{j=1}^{r} \hat{\gamma}^{t_j}
    \end{multline}
    Note that $\left (\sum_{t=0}^{N-1} \hat{\gamma}^t\right )^r = \sum_{t_1=0}^{N-1}\dots\sum_{t_r=0}^{N-1}\prod_{j=1}^r \hat{\gamma}^{t_j}$, and by applying the sum of the geometric series we obtain 
    \begin{multline}
         \bE[\|V_N(f)-\hat{\mathcal{L}}_N(f)\|^r] \leq   \left (  \frac{\hat{M} \| \hat{C} \| \|\hat{B}\|}{1-\hat{\gamma}} \right ) \|\Sigma_{gen}\|_{\ell_1} (c_e\sqrt{m}) \left ( 2\|\Sigma_{gen}\|_{\ell_1}(c_e\sqrt{m})\left (1+\|\hat{D}\|+ \frac{\hat{M}\|\hat{B}\|\|\hat{C}\|}{1-\hat{\gamma}} \right )  \right )^{r}\\
         \cdot \left ( \frac{1-\hat{\gamma}^N}{N(1-\hat{\gamma})} \right )^r
    \end{multline}
    Note that $1-\hat{\gamma}^N\leq 1$, so with $\bar{G}_{f,1}(f)\triangleq \left (  \frac{\hat{M} \| \hat{C} \| \|\hat{B}\|}{1-\hat{\gamma}} \right ) $, and $\bar{G}_{f,2}(f)\triangleq \left (1+\|\hat{D}\|+ \frac{\hat{M}\|\hat{B}\|\|\hat{C}\|}{1-\hat{\gamma}} \right ) \frac{1}{1-\hat{\gamma}}$ the statement of the lemma follows.    
\end{Proof}
\begin{lemma} \label{lemma:mgf(Vn-hatL:alt)} With notation as above the  following holds
    \begin{align}
        \bE[e^{\lambda|V_N(f)-\hat{\mathcal{L}}_N(f)|}]\leq 1
         + \bar{G}_{f,1}(f) \|\Sigma_{gen}\|_{\ell_1} (c_e\sqrt{m})
        \sum_{r=1}^{\infty} \frac{\left (\lambda \frac{2\|\Sigma_{gen}\|_{\ell_1}(c_e\sqrt{m})}{N} \bar{G}_{f,2}(f) \right )^r}{r!} \\
       =(1-\bar{G}_{f,1}(f) \|\Sigma_{gen}\|_{\ell_1} (c_e\sqrt{m}))+ \bar{G}_{f,1}(f) \|\Sigma_{gen}\|_{\ell_1} (c_e\sqrt{m}) e^{\lambda \frac{2\|\Sigma_{gen}\|_{\ell_1}(c_e\sqrt{m})}{N} \bar{G}_{f,2}(f)}
    \nonumber 
    \end{align}
\end{lemma}
\begin{Proof}[of Lemma \ref{lemma:mgf(Vn-hatL)}]
    with $X=\lambda |V_N(f) - \hat{\mathcal{L}}_N(f)|$
    \begin{align}
        \bE[e^{\lambda(V_N(f)-\hat{\mathcal{L}}_N(f))}] = 1+\sum_{r=1}^\infty \frac{\lambda^r}{r!} \bE[|V_N(f)-\hat{\mathcal{L}}_N(f)|^r] \leq 1+ \sum_{r=1}^\infty \frac{\lambda^r}{r!} \left (\frac{2\|\Sigma_{gen}\|_{\ell_1}(c_e\sqrt{m})}{N} \bar{G}_{f,2}(f) \right )^r    
    \end{align}

\end{Proof}

\begin{lemma}[Alternative bound using \cite{alquier2012pred}]
\label{cor:KL(L-V):alt2}
With probability at least $1-\delta$, the following holds
    \begin{align}
         \forall\rho:\quad &E_{f\sim \hat{\rho}} \mathcal{L} (f) \le \  E_{f\sim \hat{\rho}} V_N(f) +\dfrac{1}{\lambda}\!\left[\KL(\hat{\rho} \|\pi) + \ln\dfrac{1}{\delta}	+ \Psi_{\pi,2}(\lambda,N) \right ],
    \end{align}
      with
    \begin{equation}
        \Psi_{\pi,2}(\lambda,N)=	\ln E_{f\sim\pi} \bE[e^{\lambda(\mathcal{L}(f)-V_N(f))}] \leq  \ln E_{f\sim\pi} \left(
        e^{\frac{\lambda^2}{2N} (G_e(f)+G_{e,1}(f))^2C^2 (4G_e(f)C+1)^2}\right)  
    \end{equation}
   where  $C=c_e\sqrt{n_u+n_y}$
   \[ G_{e,1}(f)=\|D_e\|_2+\sum_{k=1}^{\infty} (k+1) \|C_eA_e^{k-1}K_e\|_2 \]
   In particular, 
   $\lim_{N \rightarrow \infty} \Psi_{\pi,2}(\lambda,N)=0$ for any $\lambda > 0$
   and for $\lambda_N=\sqrt{N}$,
   $\lim_{N \rightarrow} \frac{1}{\lambda_N} \Psi_{\pi,2}(\lambda_N,N)=0$.
   \end{lemma}
% Note that as the spectral radius $0 \le \rho(A_e) < \gamma < 1$,
% then $\|A_e^k\| < \gamma^k M$ for some $M>1$, and hence, 
% \begin{align*}
%     G_{e,1}(f) \le \|D_e\|_2 + M \|C_e\|_2\|K_e\|_2 
%    \sum_{k=1}^{\infty} (k+1)\gamma^{k-1} = 
%    \le \|D_e\|_2 + M \|C_e\|_2\||K_e\|_2 
%    (\sum_{k=1}^{\infty} \gamma^{k-1} +\sum_{k=1}^{\infty} k \gamma^{k-1}) \\
%    \le \sum_{k=1}^{\infty} (k+1)\gamma^{k-1} = 
%     G_{e,1}(f) \le \|D_e\|_2 + M \|C_e\|_2\| \|K_e\|_2
%     (\frac{1}{1-\gamma}+\frac{1}{(1-\gamma)^2}) < +\infty
% \end{align*}
\begin{Proof}[Proof of Lemma \ref{cor:KL(L-V):alt2}]
 For each $f \in \mathcal{F}$, consider
 $\mathbf{X}_t=\y(t)-\hat{\y}_f(t)$.
 Then $\mathbf{X}_t$ %is stationary and as
 \[ \mathbf{X}_t=\sum_{k=0}^{\infty} \alpha_k \e_g(t-k), \]
 where 
  \begin{align*}
		\alpha_k=\begin{cases} D_{e},& k=0\\ C_{e} A_e^{k-1}K_e,& k>0 \end{cases}
\end{align*}
 By \cite[Proposition 4.2]{alquier2013prediction} 
 $X_t$ is a weakly dependent process in the terminology of
 \cite{alquier2013prediction}, and $\|X_t\| \le G_e(f) C$ and
 the coefficient $\theta_{\infty,N}(1)$ satisfies
 $\theta_{\infty,N}(1) < 2G_{e,1}(f)C$ for all $N \mathbb{N}$. 
 Consider the function $h(x_1,\ldots,x_N)=\frac{1}{(2L+1)} \sum_{i=1}^{N} \|x_i\|_2^2$ defined on
 $\mathcal{X}=[-L,L]^N$, where $L=2G_e(f) C$. 
 Then $h$ is $1-Lipschitz$. 
 Notice that $\lambda V_N(f)=\frac{\lambda}{N} (2L+1)h(\mathbf{X}(0),\ldots,\mathbf{X}(N-1))$. 
 Then 
 \[ \bE[e^{\lambda(\mathcal{L}(f)-V_N(f))}]=\bE[e^{\frac{\lambda}{N} (2L+1) (\bE[h(\mathbf{X}(0),\ldots, \mathbf{X}(N-1))]-h(\mathbf{X}(0),\ldots,\mathbf{X}(N-1))}]
 \]
 and hence by \cite[Theorem 6.6]{alquier2013prediction}
 \[ \bE[e^{\lambda(\mathcal{L}(f)-V_N(f))}] \le e^{\frac{\lambda^2}{N} (2L+1)^2 ( \|\mathbf{X}_0\|_{\infty} + \theta_{\infty,N}(1))^2/2} 
 \]
 where $ \|\mathbf{X}_0\|_{\infty}$  is the smallest real number such
 that $\|\mathbf{X}_0\| \le \|\mathbf{X}_0\|_{\infty}$  with probability $1$.
 By using the definition $L$, and the facts that 
 $\|X_t\| \le G_e(f) C$ and $\theta_{\infty,N}(1) < 2G_{e,1}(f)C$
 the statement of the lemma follows.

 \begin{align}
     \bE[e^{\lambda(\mathcal{L}(f)-V_N(f))}] \le e^{\frac{\lambda^2}{N} (2L+1)^2 ( \|\mathbf{X}_0\|_{\infty} + \theta_{\infty,N}(1))^2/2} \leq e^{\frac{\lambda^2}{N} (4G_e(f)C+1)^2 ( G_e(f) + 2G_{e,1})^2 C^2/2}
 \end{align}
\end{Proof}

\begin{Proof}[of Theorem \ref{thm:bounded_alt}] \label{proof:thm:bounded_alt}

    By applying Lemma \ref{cor:KL(L-V):alt2}, Lemma \ref{lemma:mgf(Vn-hatL:alt)}, and by applying the union bound as in Lemma \ref{lemma:unbounded_full_KL}, we obtain, for $\lambda>0$, $\delta \in (0,1]$, with probability at least $1-2\delta$
    \begin{align}
        \forall \rho\in\mathcal{M}_\pi: \; E_{f\sim\rho} \mathcal{L}(f) \leq \E_{f\sim\rho} \hat{\mathcal{L}}_N(f) + \frac{2}{\lambda}\left [ \KL(\rho|\pi) + \ln \frac{1}{\delta} +\frac{\Psi_1(\lambda,N)+\Psi_2(\lambda,N)}{2} \right ]
    \end{align}
    with
    \begin{align}
        \Psi_1(\lambda,N) &\triangleq \ln E_{f\sim\pi} e^{\frac{\lambda^2}{2N} (4G_e(f)C+1)^2 ( G_e(f) + 2G_{e,1})^2 C^2}\\ 
        \Psi_2(\lambda,N) &\triangleq \ln E_{f\sim\pi} \left ( (1-\bar{G}_{f,1}(f) \|\Sigma_{gen}\|_{\ell_1} (c_e\sqrt{m}))+ \bar{G}_{f,1}(f) \|\Sigma_{gen}\|_{\ell_1} (c_e\sqrt{m}) e^{\lambda \frac{2\|\Sigma_{gen}\|_{\ell_1}(c_e\sqrt{m})}{N} \bar{G}_{f,2}(f)} \right )
    \end{align}
    Now with $\tilde{\lambda}\triangleq 0.5\lambda \leftrightarrow \lambda=2\tilde{\lambda}$, we obtain the statement of the lemma: for $\tilde{\lambda}>0$, $\delta\in(0,1]$, then
    with probability at least $1-2\delta$
    \begin{align}
        \forall \rho\in\mathcal{M}_\pi: \; E_{f\sim\rho} \mathcal{L}(f) \leq \E_{f\sim\rho} \hat{\mathcal{L}}_N(f) + \frac{1}{\tilde{\lambda}}\left [ \KL(\rho|\pi) + \ln \frac{1}{\delta} +\frac{\Psi_1(\tilde{\lambda},N)+\Psi_2(\tilde{\lambda},N)}{2} \right ]
    \end{align}
    with
    \begin{align}
        \Psi_1(\tilde{\lambda},N) &\triangleq \ln E_{f\sim\pi} e^{\frac{\tilde{\lambda}^2}{N} 2(4G_e(f)C+1)^2 ( G_e(f) + 2G_{e,1})^2 C^2}\\ 
        \Psi_2(\tilde{\lambda},N) &\triangleq \ln E_{f\sim\pi} \left ( (1-\bar{G}_{f,1}(f) \|\Sigma_{gen}\|_{\ell_1} (c_e\sqrt{m}))+ \bar{G}_{f,1}(f) \|\Sigma_{gen}\|_{\ell_1} (c_e\sqrt{m}) e^{\frac{\tilde{\lambda}}{N} 8\|\Sigma_{gen}\|_{\ell_1}(c_e\sqrt{m}) \bar{G}_{f,2}(f)} \right )
    \end{align}
    
\end{Proof}

\end{document}